\documentclass[10pt,twocolumn,letterpaper]{article}

\usepackage{pdfpages}
\usepackage{3dv}
\usepackage{times}
\usepackage{epsfig}
\usepackage{graphicx}
\usepackage{amsmath}
\usepackage{amssymb}

\usepackage{caption}
\usepackage{subcaption}

\usepackage{dcolumn}
\usepackage{booktabs}
\usepackage[normalem]{ulem}
\usepackage{rotating}
\usepackage{multirow}
\usepackage{xr}
\usepackage{comment}
\usepackage{ctable}
\usepackage{colortbl}
\usepackage{collcell}
\usepackage{hhline}

\usepackage[utf8]{inputenc}
\usepackage{amsfonts,amsthm}
\usepackage{xcolor}
\usepackage{algorithm}
\usepackage[noend]{algpseudocode}
\usepackage{xspace}
\usepackage{array}
\usepackage{cite}
\usepackage{float}
\usepackage{titling}
\usepackage[pagebackref=true,breaklinks=true,letterpaper=true,colorlinks,bookmarks=false]{hyperref}

\usepackage{amsmath, amssymb, amsfonts, bm, mathtools}

\makeatletter

\newcommand{\minimize}{\@ifstar{\@minimizes}{\@minimize}}
\newcommand{\@minimizes}[1]{\ensuremath{ \operatorname{minimize } } }
\newcommand{\@minimize }[1]{\ensuremath{&\operatorname{minimize} &&}}
\newcommand{\maximize}{\@ifstar{\@maximizes}{\@maximize}}
\newcommand{\@maximizes}[1]{\ensuremath{ \operatorname{maximize } } }
\newcommand{\@maximize }[1]{\ensuremath{&\operatorname{maximize} &&}}
\newcommand{\minimizex}{\@ifstar{\@minimizexs}{\@minimizex}}
\newcommand{\@minimizexs}[1]{\ensuremath{ \underset{#1}{\operatorname{minimize}}\ }}
\newcommand{\@minimizex }[1]{\ensuremath{&\underset{#1}{\operatorname{minimize}}&&}}
\newcommand{\maximizex}{\@ifstar{\@maximizexs}{\@maximizex}}
\newcommand{\@maximizexs}[1]{\ensuremath{ \underset{#1}{\operatorname{maximize}}\ }}
\newcommand{\@maximizex }[1]{\ensuremath{&\underset{#1}{\operatorname{maximize}}&&}}
\newcommand{\argmin}{\@ifstar{\@argmins}{\@argmin}}
\newcommand{\@argmins}[1]{\ensuremath{ \operatorname{argmin } } }
\newcommand{\@argmin }[1]{\ensuremath{&\operatorname{argmin} &&}}
\newcommand{\argmax}{\@ifstar{\@argmaxs}{\@argmax}}
\newcommand{\@argmaxs}[1]{\ensuremath{ \operatorname{argmax } } }
\newcommand{\@argmax }[1]{\ensuremath{&\operatorname{argmax} &&}}
\newcommand{\argminx}{\@ifstar{\@argminxs}{\@argminx}}
\newcommand{\@argminxs}[1]{\ensuremath{ \underset{#1}{\operatorname{argmin}}\ }}
\newcommand{\@argminx }[1]{\ensuremath{&\underset{#1}{\operatorname{argmin}}&&}}
\newcommand{\argmaxx}{\@ifstar{\@argmaxxs}{\@argmaxx}}
\newcommand{\@argmaxxs}[1]{\ensuremath{ \underset{#1}{\operatorname{argmax}}\ }}
\newcommand{\@argmaxx }[1]{\ensuremath{&\underset{#1}{\operatorname{argmax}}&&}}

\makeatother

\newcommand{\cY}{\mathcal{Y}}
\newcommand{\cX}{\mathcal{X}}

\newcommand{\figref}[1]{\Fig~\ref{#1}}
\newcommand{\secref}[1]{Section~\ref{#1}}
\renewcommand{\eqref}[1]{Equation~\ref{#1}}
\newcommand{\tabref}[1]{Table~\ref{#1}}

\makeatletter
\DeclareRobustCommand\onedot{\futurelet\@let@token\@onedot}
\def\@onedot{\ifx\@let@token.\else.\null\fi\xspace}
\def\eg{e.g\onedot} 
\def\ie{i.e\onedot} 
\def\cf{cf\onedot}

\def\wrt{with respect to }
\def\etal{et~al\onedot} 
\def\Fig{Fig\onedot}   
\makeatother

\newcommand{\boldparagraph}[1]{\vspace{0.2cm}\noindent{\bf #1:} }

\newcommand{\inputD}{\mathcal{X}}
\newcommand{\inputLetter}{x}
\newcommand{\inputs}{\boldsymbol{\inputLetter}}
\newcommand{\inputuv}{\inputLetter_{u,v}}

\newcommand{\flags}{\boldsymbol{\flagLetter}}
\newcommand{\flagLetter}{o}
\newcommand{\flaguv}{\flagLetter_{u,v}}

\newcommand{\netfunction}{f}
\newcommand{\netconffunction}{f^\flagLetter}

\newcommand{\rotatedlabel}[1]{\begin{sideways}#1\end{sideways}}

\setupctable{captionskip=5pt}
\def\colorModel{hsb} 
\newcommand\ColCell[1]{
  \pgfmathparse{#1>5?0:1}  
    \ifnum\pgfmathresult=0\relax\color{white}\fi
    \pgfmathsetmacro\compA{0}      
    \pgfmathsetmacro\compB{0} 
    \pgfmathsetmacro\compC{(#1>100)?0:1-(#1/10)}      
  \edef\x{\noexpand\centering\noexpand\cellcolor[\colorModel]{\compA,\compB,\compC}}\x #1
  }
\newcolumntype{E}{>{\collectcell\ColCell}m{0.5cm}<{\endcollectcell}}  

\graphicspath{ {figures/} }

\threedvfinalcopy 

\def\threedvPaperID{70} 
\def\httilde{\mbox{\tt\raisebox{-.5ex}{\symbol{126}}}}

\ifthreedvfinal\fi
\begin{document}
	

\title{Sparsity Invariant CNNs}

\author{Jonas Uhrig$^{\star,1,2}$ \quad
Nick Schneider$^{\star,1,3}$ \quad
Lukas Schneider$^{1,4}$\\
Uwe Franke$^{1}$ \quad
Thomas Brox$^{2}$ \quad
Andreas Geiger$^{4,5}$\\
~\\
$^*$ The first two authors contributed equally to this work\\
$^1$Daimler R\&D Sindelfingen \quad $^2$University of Freiburg\\
$^3$KIT Karlsruhe \quad $^4$ETH Z\"urich \quad $^5$MPI T\"ubingen\\
{\tt\small \{jonas.uhrig,nick.schneider\}@daimler.com}
}
\maketitle



\begin{abstract}
In this paper, we consider convolutional neural networks operating on sparse inputs with an application to depth upsampling from sparse laser scan data. First, we show that traditional convolutional networks perform poorly when applied to sparse data even when the location of missing data is provided to the network.
To overcome this problem, we propose a simple yet effective sparse convolution layer which explicitly considers the location of missing data during the convolution operation.
We demonstrate the benefits of the proposed network architecture in synthetic and real experiments \wrt various baseline approaches.
Compared to dense baselines, the proposed sparse convolution network generalizes well to novel datasets and is invariant to the level of sparsity in the data.
For our evaluation, we derive a novel dataset from the KITTI benchmark, comprising 93k depth annotated RGB images. Our dataset allows for training and evaluating depth upsampling and depth prediction techniques in challenging real-world settings and will be made available upon publication.
\end{abstract}


\section{Introduction}

Over the last few years, convolutional neural networks (CNNs)
have impacted nearly all areas of computer vision.
In most cases, the input to the CNN is an image or video,
represented by a densely populated matrix or tensor. By combining
convolutional layers with non-linearites and pooling layers, CNNs are able
to learn distributed representations, extracting low-level
features in the first layers, followed by successively higher-level
features in subsequent layers. However, when the input to the
network is sparse and irregular (\eg, when only 10\% of the
pixels carry information), it becomes less clear how the
convolution operation should be defined as for each
filter location the number and placement of the inputs varies.

\setlength\tabcolsep{2pt}
\begin{figure}[tb]
\begin{center}
    \begin{tabular}{cc}
        \includegraphics[width=0.23\textwidth]{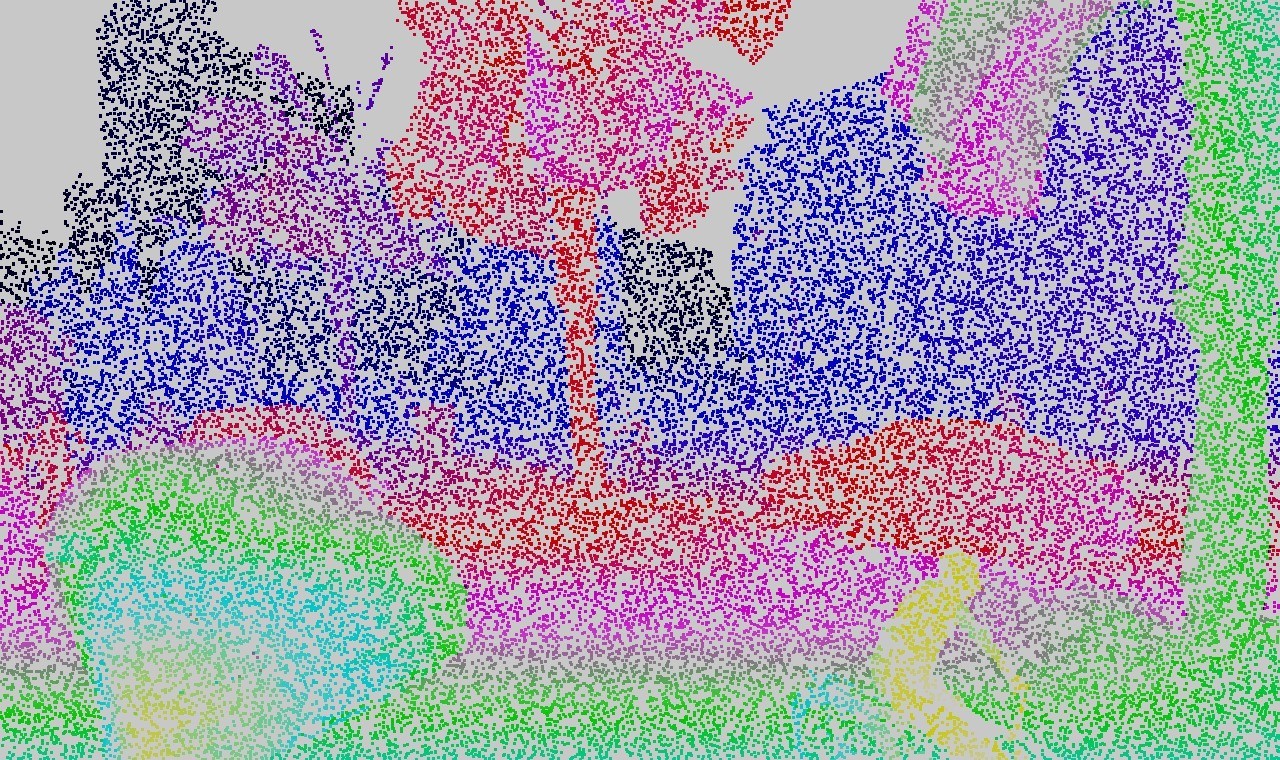} &
        \includegraphics[width=0.23\textwidth]{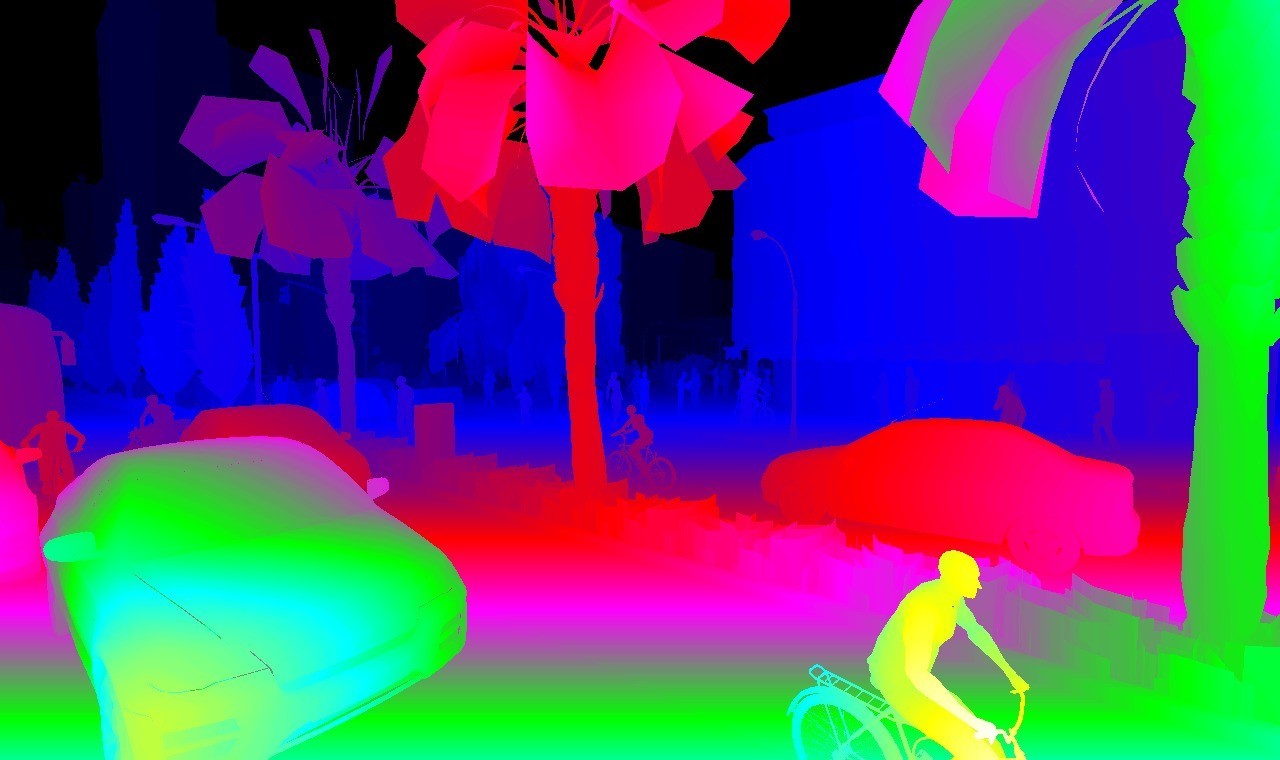}\\
        \footnotesize{(a) Input (visually enhanced)} &
        \footnotesize{(b) Ground truth} \\
        \includegraphics[width=0.23\textwidth]{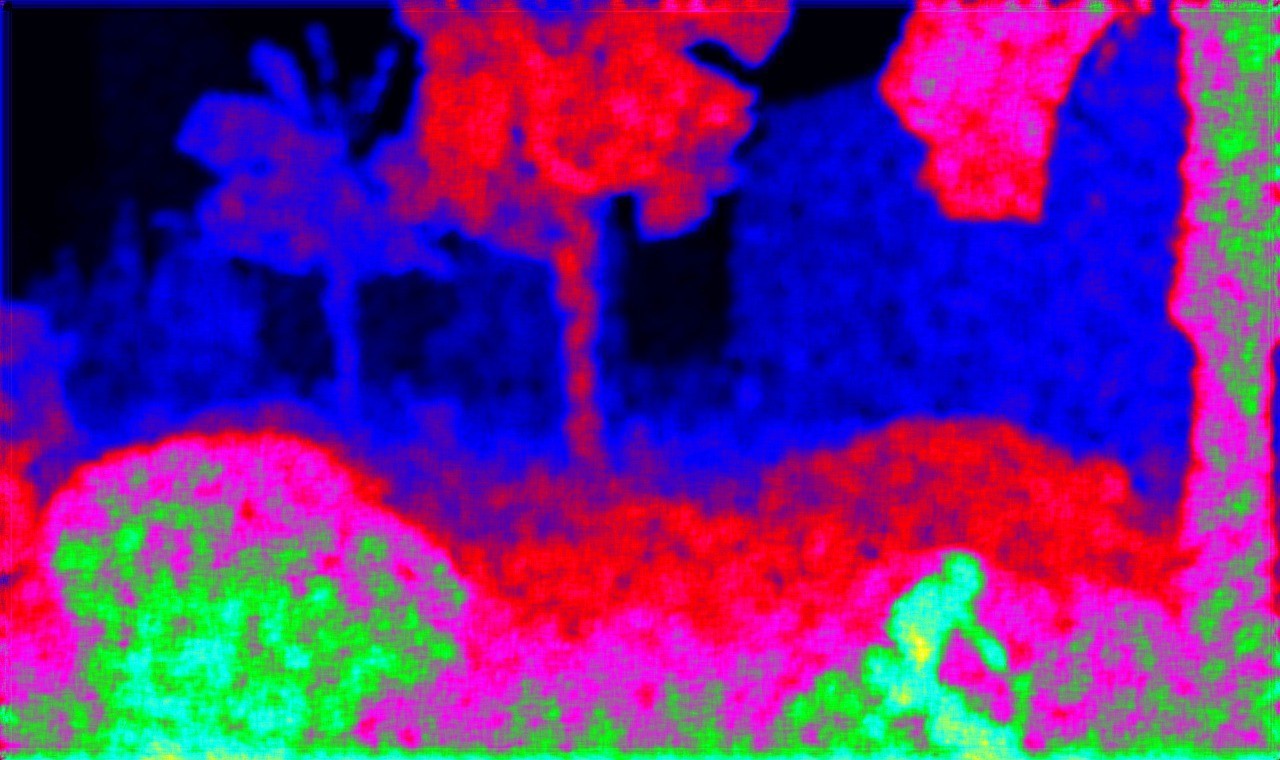} &
        \includegraphics[width=0.23\textwidth]{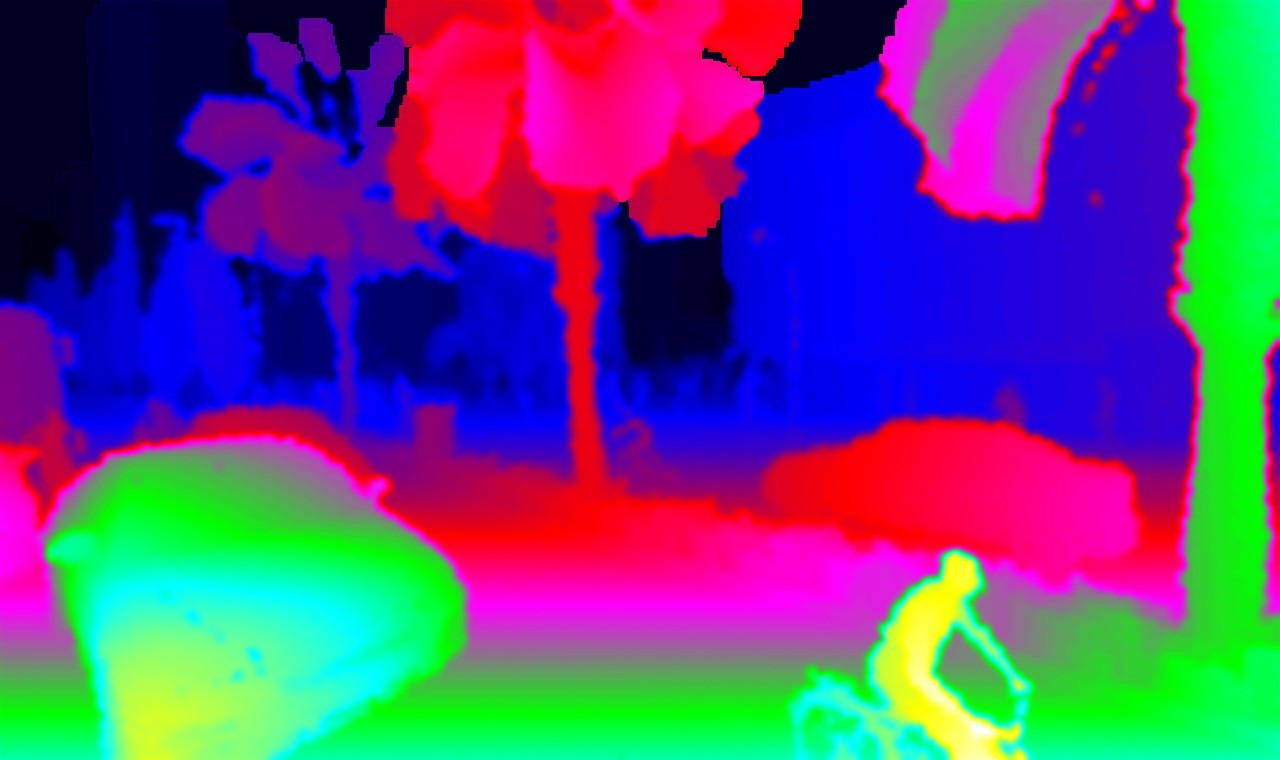} \\
        \footnotesize{(c) Standard ConvNet} &
        \footnotesize{(d) Our Method}\\
    \end{tabular}
    \caption{\textbf{Depth Map Completion.} Using sparse, irregular depth measurements (a) as inputs leads to noisy results when processed with standard CNNs (c). In contrast, our method (d) predicts
    smooth and accurate depth maps by explicitly considering sparsity during convolution.}
	\label{fig:illustration}
\end{center}
\end{figure}

The na\"{i}ve approach to this problem is to assign a
default value to all non-informative sites \cite{Chen2016ARXIVa,Li2016RSS}.
Unfortunately, this approach leads to suboptimal results as
the learned filters must be invariant to all possible patterns
of activation whose number grows exponentially with the filter size.
In this paper, we investigate a simple yet effective solution to this problem
which outperforms the na\"{i}ve approach
and several other baselines. In particular, we introduce
a novel sparse convolutional layer which weighs the elements
of the convolution kernel according to the validity of the input
pixels. Additionally, a second stream stream carries information
about the validity of pixels to subsequent layers of the
network. This enables our approach to handle large levels
of sparsity without significantly compromising accuracy.

Importantly, our representation is invariant to the level of sparsity in the input.
As evidenced by our experiments, training our network at a sparsity
level different from the sparsity level at test time does not
significantly deteriorate the results. This has important
applications, \eg, in the context of robotics where
algorithms must be robust to changes in sensor configuration.

One important area of application for the proposed technique is
enhancement of 3D laser scan data, see \figref{fig:illustration} for an illustration.
While laser scanners provide valuable information about depth and
reflectance, the resulting point clouds are typically very sparse,
in particular when considering mobile scanners like the
Velodyne HDL-64e\footnote{\url{http://velodynelidar.com/hdl-64e.html}}
used in autonomous driving \cite{Geiger2012CVPR}.

Learning models which are able to increase the density of such scans is thus highly desirable.
Unfortunately, processing high-resolution data directly in 3D is challenging
without compromising accuracy \cite{Riegler2017CVPR}.

An alternative, which we follow in this paper, is to project the
laser scan onto a virtual or real 2D image plane resulting in a 2.5D representation.
Besides modeling depth prediction as a 2D regression problem, this
representation has the advantage that additional dense
information (\eg, RGB values from a color camera) can be easily integrated.
However, projected laser scans are typically very sparse and not guaranteed to
align with a regular pixel grid, hence leading to poor results when processed with standard CNNs. In contrast, the proposed method produces compelling results even
when the input is sparse and irregularly distributed.

We evaluate our method in ablation studies
and against several state-of-the-art baselines.
For our evaluation, we leverage the synthetic Synthia dataset \cite{Ros2016CVPR} as well as a newly proposed real-world dataset with 93k depth annotated images derived from the KITTI raw dataset \cite{Geiger2013IJRR}.
Our dataset is the first to provide a significant number of high-quality depth annotations for this scenario.
Besides attaining higher
accuracy in terms of depth and semantics we demonstrate our method's ability
to generalize across varying datasets and levels of sparsity.
Our code and dataset will be released upon publication.
\externaldocument{method}

\section{Related Work}

In this section, we discuss methods which operate on sparse {\it inputs} followed by techniques that consider sparsity {\it within} the CNN. We briefly discuss the state-of-the-art in invariant representation learning and conclude with a review on related depth upsampling techniques.

\boldparagraph{CNNs with Sparse Inputs}
The na\"{i}ve approach to handling sparse inputs is to either zero the invalid values or to create an additional input channel for the network which encodes the validity of each pixel. For detecting objects in laser scans, Chen \etal \cite{Chen2016ARXIVa} and Li \etal \cite{Li2016RSS}
project the 3D point clouds from the laser scanner onto a low resolution image, zero the missing values and run a standard CNN on this input.
For optical flow interpolation and inpainting, Zweig \etal \cite{Zweig2016ARXIV} and Köhler \etal \cite{Koehler2014GCPR} pass an additional binary validity mask to the network.
As evidenced by our experiments, both strategies are suboptimal compared to explicitly considering sparsity inside the convolution layers.

Jampani \etal \cite{Jampani2016CVPR} use bilateral filters as layers inside a CNN and learn the parameters of the corresponding permutohedral convolution kernel. While their layer handles sparse irregular inputs, it requires guidance information to construct an effective permutohedral representation and is computationally expensive for large grids.
Compared to their approach our sparse convolutional networks yield significantly better results for depth upsampling while being as efficient as regular CNNs.

Graham \cite{graham2014spatially,graham2015sparse} and Riegler \etal  \cite{Riegler2017CVPR} consider sparse 3D inputs.
In contrast to our work, their focus is on improving computational efficiency and memory demands by partitioning the space according to the input.
However, regular convolution layers are employed which suffer from the same drawbacks as the na\"{i}ve approach described above.

\boldparagraph{Sparsity in CNNs}
A number of works \cite{Han2015NIPS,Liu2015CVPR,Park2016ARXIV,Wen2016NIPS,Figurnov2016NIPS}
also consider sparsity {\it within} convolutional neural networks.
Liu \etal \cite{Liu2015CVPR} show how to reduce the redundancy in the parameters using a sparse
decomposition.
Their approach eliminates more than
90\%  of  parameters,  with  a  drop  of  accuracy  of  less than  1\%  on  ILSVRC2012.
Wen \etal \cite{Wen2016NIPS} propose to regularize the structures (\ie, filters, channels and layer depth) of deep neural networks to obtain a hardware friendly representation. They report speed-up factors of 3 to 5 \wrt regular CNNs.
While these works focus on improving efficiency of neural networks by exploiting sparsity {\it within} the network, we consider the problem of sparse {\it inputs} and do not tackle efficiency. A combination of the two lines of work will be an interesting direction for future research.

\boldparagraph{Invariant Representations}
Learning models robust to variations of the input is a long standing goal of computer vision. The most commonly used solution to ensure robustness is data augmentation~\cite{Simard2003ICDAR,Krizhevsky2012NIPS,Laptev2016CVPR}. More recently, {\it geometric} invariances (\eg, rotation, perspective transformation) have been incorporated directly into the filters of CNNs~\cite{Cohen2016ICML, Worrall2017CVPR, Zhou2017CVPR,Jaderberg2015NIPS,Henriques2016ARXIV}. In this paper, we consider the problem of learning representations invariant to the {\it level of sparsity} in the input. As evidenced by our experiments, our model performs well even when the sparsity level differs significantly between the training and the test set. This has important implications as it allows for replacing the sensor (\eg, laser scanner) without retraining the network.

\boldparagraph{Depth Upsampling}
We evaluate the effectiveness of our approach for the task of depth upsampling, which is an active area of research with applications in, \eg, stereo vision, optical flow and 3D reconstruction from laser scan data.
While some methods operate directly on the depth input, others require guidance, \eg, from a high resolution image.

Methods for \textit{non-guided depth upsampling} are closely related to those for single image superresolution.
Early approaches have leveraged repetitive structures to identify
similar patches across different scales in 2D \cite{Glasner2009,mac2012patch} and 3D \cite{Hornacek2013CVPR}.
More recently, deep learning based methods for depth \cite{riegler2016atgv} and image
superresolution \cite{yang2010image,dong2014learning,Dong2016PAMI,Kim2016CVPR} have surpassed traditional
upsampling techniques in terms of accuracy and efficiency.
However, all aforementioned methods assume that the data is located on a regular grid and therefore cannot be applied for upsampling sparse and irregularly distributed 3D laser scan data as considered in this paper.

\textit{Guided upsampling}, on the other hand, uses the underlying assumption that the target domain shares commonalities with a high-resolution guidance image, \eg, that image edges align with depth discontinuities.
A popular choice for guided upsampling is guided bilateral filtering \cite{Chan2008,Dolson2010a,Kopf2007,Yang2007a,Liu2013CVPRa}.
More advanced approaches are based on global energy minimization \cite{Diebel2005NIPS,Park2011ICCV,Ferstl2013,barron2016fast,riegler2016deep}, compressive sensing \cite{Hawe2011ICCV},
or incorporate semantics for improved performance
\cite{schneider2016semantically}.
Several approaches also exploit end-to-end models for guided depth upsampling of regular data
\cite{Hui2016ECCV,Song2016ACCV}.
While some of the aforementioned techniques are able to handle sparse inputs, they heavily rely on the guidance signal. In contrast, here we propose a learning based solution to the problem, yielding compelling results even without image guidance.
Unlike existing CNN-based approaches for depth upsampling \cite{Hui2016ECCV,Song2016ACCV}, the proposed convolution layer handles sparse irregular inputs which occur, \eg, in 3D laser scan data.
\externaldocument{experiments}

\section{Method}

Let $f$ denote a mapping from input domain $\cX$ (\eg, intensity, depth) to output domain $\cY$ (\eg, depth, semantics), implemented via a convolutional neural network.
In this paper, we consider
the case, where the inputs $\inputs = \{\inputuv\} \in \inputD$ are only
partially observed. Let $\flags = \{\flaguv\}$ denote corresponding binary variables indicating if an input is observed ($\flaguv=1$) or not ($\flaguv=0$).
The output of a standard convolutional layer in a CNN is computed via
\begin{equation}
\netfunction_{u,v} ( \inputs ) = {
\sum_{i,j=-k}^{k}\inputLetter_{u+i,v+j}\,w_{i,j}}  + b
\label{eq:regular_convolution}
\end{equation}
with kernel size $2k+1$, weight $w$ and bias $b$. If the input comprises multiple features, $\inputuv$ and $w_{i,j}$ represent vectors whose length depends on the number of input channels.

\begin{figure*}[t!]
\begin{subfigure}{0.65\linewidth}
\includegraphics[width=\linewidth]{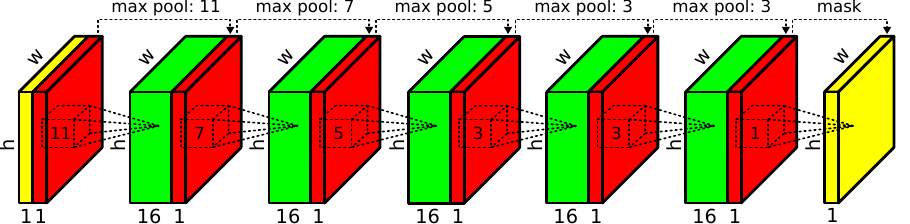}
\caption{{\bf Network Architecture}}
\label{fig:network_architecture}
\end{subfigure}\hspace{0.6cm}
\begin{subfigure}{0.3\linewidth}
\includegraphics[width=\linewidth]{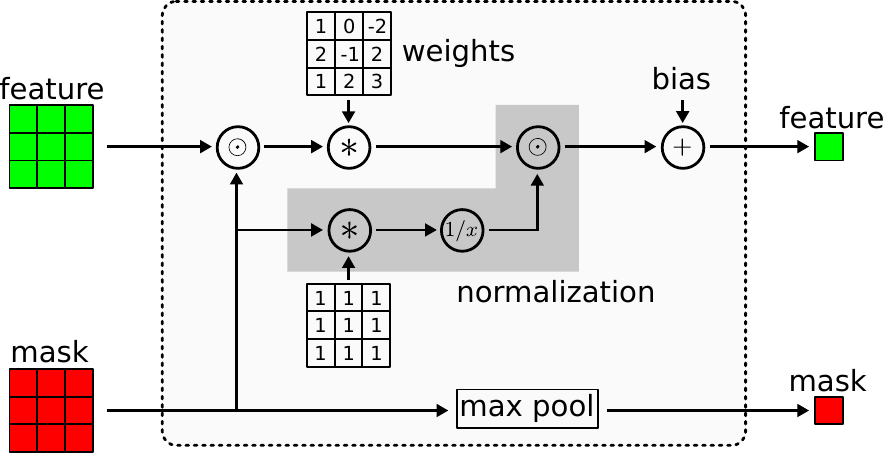}
\caption{{\bf Sparse Convolution}}
\label{fig:sparse_convolution}
\end{subfigure}
\caption{{\bf Sparse Convolutional Network.} (\subref{fig:network_architecture}) The input to our network is a sparse depth map (yellow) and a binary observation mask (red). It passes through several sparse convolution layers (dashed) with decreasing  kernel sizes from $11\times 11$ to $3\times 3$. (\subref{fig:sparse_convolution}) Schematic of our sparse convolution operation. Here, $\odot$ denotes elementwise multiplication, $*$ convolution, $1/x$ inversion and ``max pool'' the max pooling operation. The input feature can be single channel or multi-channel.}
\label{fig:sparse_convolution_network}
\end{figure*}

\subsection{Na\"{i}ve Approach}

There are two na\"{i}ve ways to deal with unobserved inputs.
First, invalid inputs $\inputuv$ can be encoded using a default value, \eg, zero.
The problem with this approach is that the network must learn to distinguish between observed inputs and those being invalid. This is a difficult task as the number of possible binary patterns grows exponentially with the kernel size.
Alternatively, ${\flags}$ can be used as an additional input to the network in the hope
that the network learns the correspondence between the observation mask and the inputs.
Unfortunately, both variants struggle to learn robust representations from sparse inputs (see \secref{sec:experiments}).

\subsection{Sparse Convolutions}

To tackle this problem, we propose a convolution operation which explicitly considers sparsity by evaluating only observed pixels and normalizing the output appropriately:
\begin{equation}
\netfunction_{u,v} ( \inputs , \flags )
 = \frac{
  \sum_{i,j=-k}^{k}\flagLetter_{u+i,v+j}\,\inputLetter_{u+i,v+j}\,w_{i,j}}
 {\sum_{i,j=-k}^{k}\flagLetter_{u+i,v+j}+\epsilon} + b
\label{eq:sparse_convolution}
\end{equation}
Here, a small $\epsilon$ is added
to the denominator to avoid division by zero at filter locations where none of the input pixels $\inputLetter_{u+i,v+j}$ are observed. Note that \eqref{eq:sparse_convolution} evaluates to a (scaled) standard convolution when the input is dense.

The primary motivation behind the proposed sparse convolution operation is to render the filter output invariant to the actual number of observed inputs which varies significantly between filter locations due to the sparse and irregular input.
Note that in contrast to other techniques \cite{riegler2016atgv,Ferstl2013} which artificially upsample the input (\eg, via interpolation), our approach operates directly on the input and doesn't introduce additional distractors.

When propagating information to subsequent layers, it is important to keep track of the visibility state and make it available to the next layers of the network. In particular, we like to mark output locations as ``unobserved'' when none of the filter's inputs has been observed. We thus determine subsequent observation masks in the network
$\netconffunction_{u,v}( \flags )$ via the max pooling operation
\begin{equation}
\netconffunction_{u,v}( \flags ) = \max_{i,j=-k,..,k}\flagLetter_{u+i,v+j}
\end{equation}
which evaluates to $1$ if at least one observed variable is visible to the
filter and $0$ otherwise. In combination with the output of the convolution this serves as input for the next sparse convolution layer.
The complete architecture of our network is illustrated in \figref{fig:sparse_convolution_network}.

\subsubsection{Skip Connections}
\label{sec:weighted_skip}

So far we have only considered the convolution operation.
However, state-of-the-art CNNs comprise many different types of layers implementing
different mathematical operations. Many of those can be easily generalized to consider observation masks. Layers that take the outputs of multiple preceding
layers and combine them to a single output, \eg, by summation, are used
frequently in many different network architectures, \eg, summation in inception
modules \cite{Szegedy2015CVPR} or skip connections in ResNets \cite{He2016CVPR} as well as fully
convolutional networks \cite{FCN2015}. With additional observation indicators,
the summation of input layers for each channel $c$ and location $(u,v)$
can be redefined as a normalized sum over the observed inputs $\inputLetter^{l}$

\begin{equation}
\netfunction^+ ( \inputs , \flags ) = \frac{\sum_{l=1}^n\flagLetter^{l}\inputLetter^{l}}
 {\sum_{l=1}^n\flagLetter^{l}}
\end{equation}
where $n$ denotes the number of input layers.
If all pixels are observed, this expression simplifies to the standard operation $\sum_{l=1}^n\inputLetter^{l}$.

\externaldocument{experiments}

\section{Large-Scale Dataset}

Training and evaluating the proposed depth upsampling approach requires access to a large annotated dataset.
While evaluation on synthetic datasets \cite{Ros2016CVPR,Gaidon2016CVPR,mayer_sceneflownet_2016} is possible, it remains an open question if the level of realism attained by such datasets is sufficient to conclude about an algorithm's performance in challenging real-world situations.

Unfortunately, all existing real-world datasets with sanitized depth ground truth are small in scale. The
Middlebury benchmark \cite{Scharstein2002IJCV,Scharstein2014GCPR} provides depth estimates only for a dozen
images and only in controlled laboratory conditions. While the Make3D dataset \cite{Saxena2009PAMI} considers
more realistic scenarios, only 500 images of small resolution are provided.
Besides, KITTI \cite{Geiger2012CVPR,Menze2015CVPR} provides 400 images of street scenes with associated depth ground
truth. However, none of these datasets is large enough for end-to-end training of high-capacity deep neural networks.

For our evaluation, we therefore created a new large-scale dataset based on the KITTI raw datasets \cite{Geiger2013IJRR}
which comprises 93k frames with semi-dense depth ground truth. While the KITTI raw datasets provide depth information
in the form of raw Velodyne scans, significant manual effort is typically required to remove noise in the laser scans,
artifacts due to occlusions (\eg, due to the different centers of projection of the laser scanner and the camera) or
reflecting/transparent surfaces in the scene \cite{Geiger2012CVPR}. It is therefore highly desirable to automate this task.

In this paper, we propose to remove outliers in the laser scans by comparing the scanned depth to results from a stereo
reconstruction approach using semi-global matching \cite{Hirschmueller2008PAMI}.
While stereo reconstructions typically lead to depth bleeding artifacts at object boundaries, LiDaR sensors create streaking
artifacts along their direction of motion.
To remove both types of outliers, we enforce
consistency between laser scans and stereo reconstruction and remove all LiDaR points exhibiting large relative errors.
For comparing both measurements,
we transform the SGM disparity maps to depth values using KITTI's provided calibration files.
We further follow \cite{Geiger2012CVPR} and accumulate 11 laser scans to increase the density of the generated depth maps.
While the environment is mostly static, some of the KITTI sequences comprise dynamic objects, where a laser scan accumulation causes
many outliers on dynamic objects. Therefore, we use the SGM depth maps only once to clean the accumulated laser scan projection
(instead of cleaning each laser scan separately) in order to remove all outliers in one step: Occlusions, dynamic motion and
measurement artifacts.
We also observed that most errors due to reflecting and transparent surfaces can be removed with this simple technique as
SGM and LiDaR rarely agree in those regions.

\begin{figure*}
\begin{center}
\setlength{\tabcolsep}{1pt}
\begin{tabular}{ccccc}
KITTI 2015 \cite{Menze2015CVPR} & Our Dataset & Raw LiDaR & Acc. LiDaR & SGM \cite{Hirschmueller2008PAMI} \\ \vspace{-0.1cm}
\includegraphics[width=0.194\textwidth, trim={20pt 20pt 20pt 20pt}, clip]{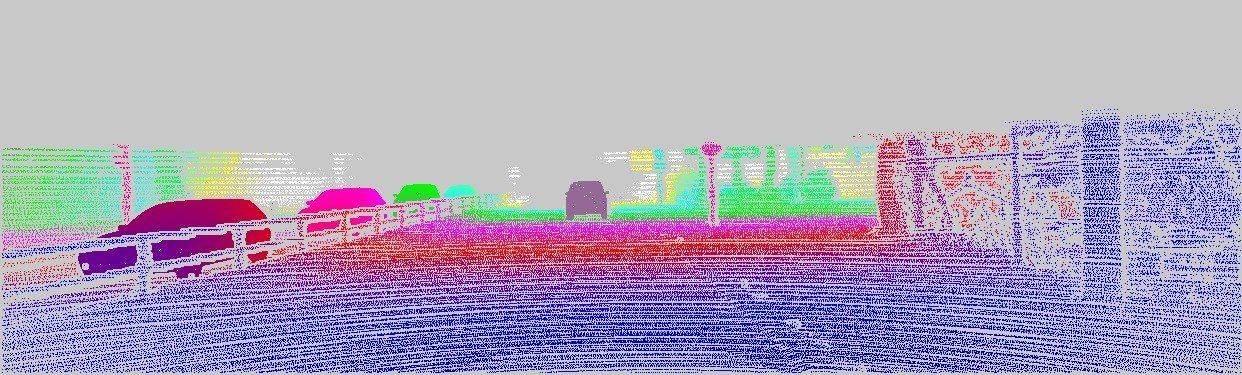}             &
\includegraphics[width=0.194\textwidth, trim={20pt 20pt 20pt 20pt}, clip]{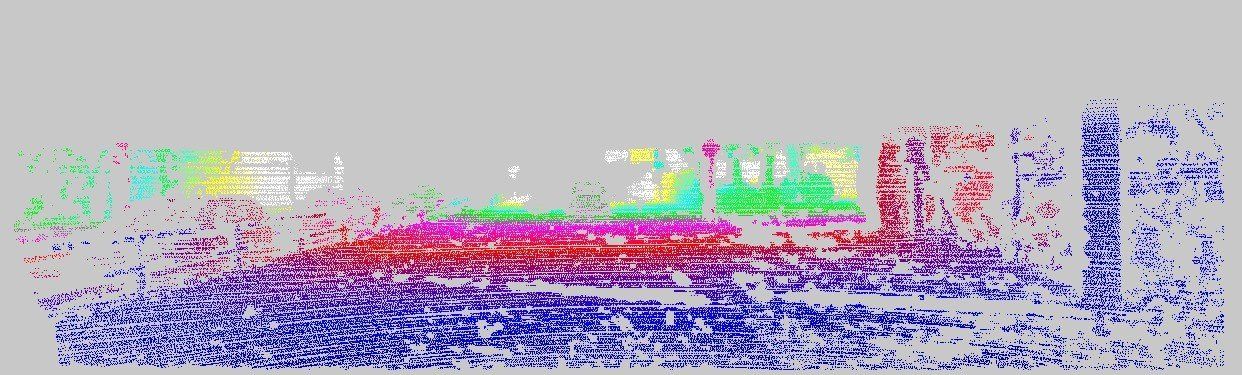}         &
\includegraphics[width=0.194\textwidth, trim={20pt 20pt 20pt 20pt}, clip]{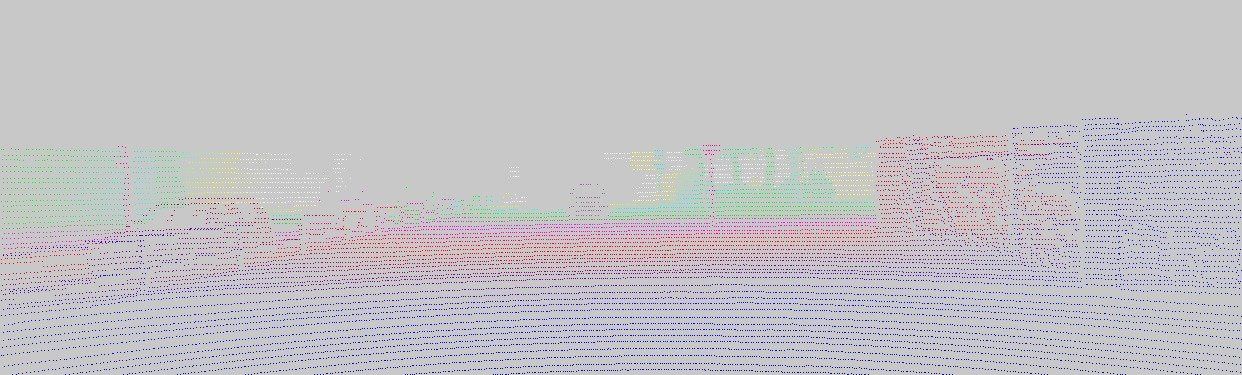}         &
\includegraphics[width=0.194\textwidth, trim={20pt 20pt 20pt 20pt}, clip]{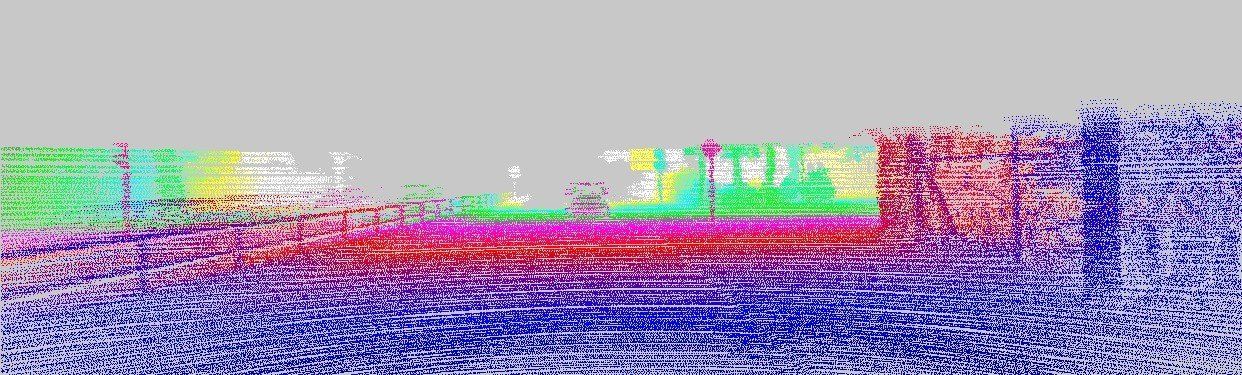}         &
\includegraphics[width=0.194\textwidth, trim={20pt 20pt 20pt 20pt}, clip]{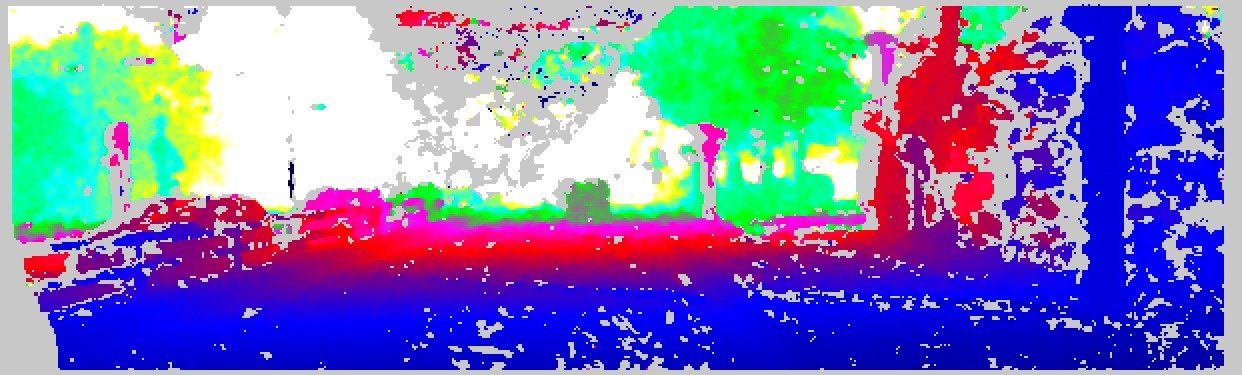} \\ \vspace{-0.1cm}
\includegraphics[width=0.194\textwidth, trim={20pt 20pt 20pt 20pt}, clip]{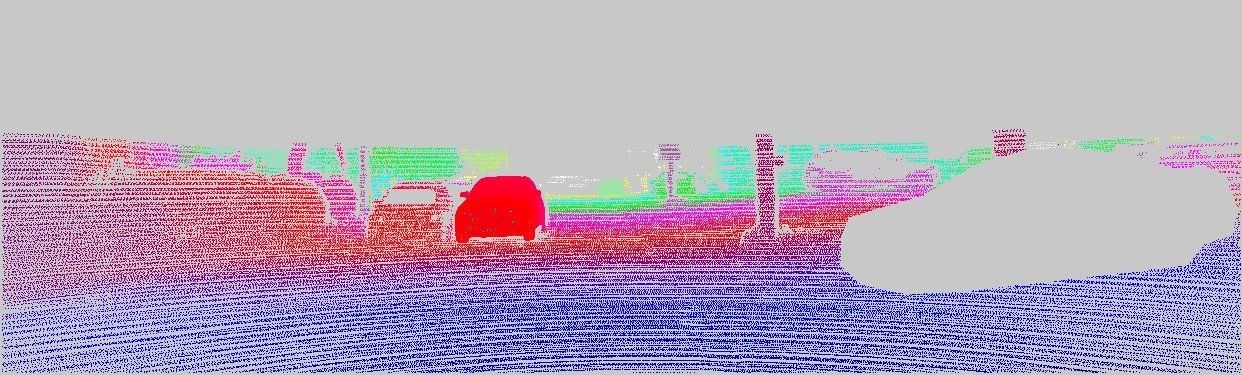}             &
\includegraphics[width=0.194\textwidth, trim={20pt 20pt 20pt 20pt}, clip]{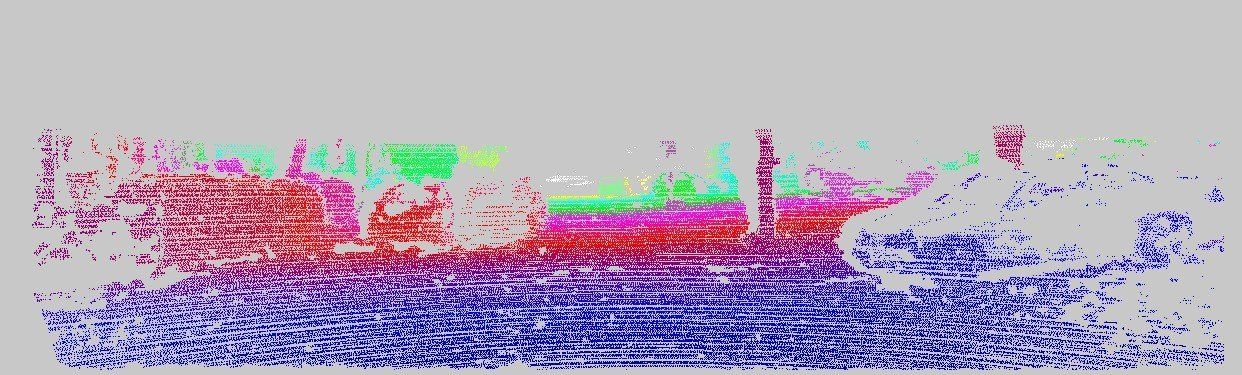}         &
\includegraphics[width=0.194\textwidth, trim={20pt 20pt 20pt 20pt}, clip]{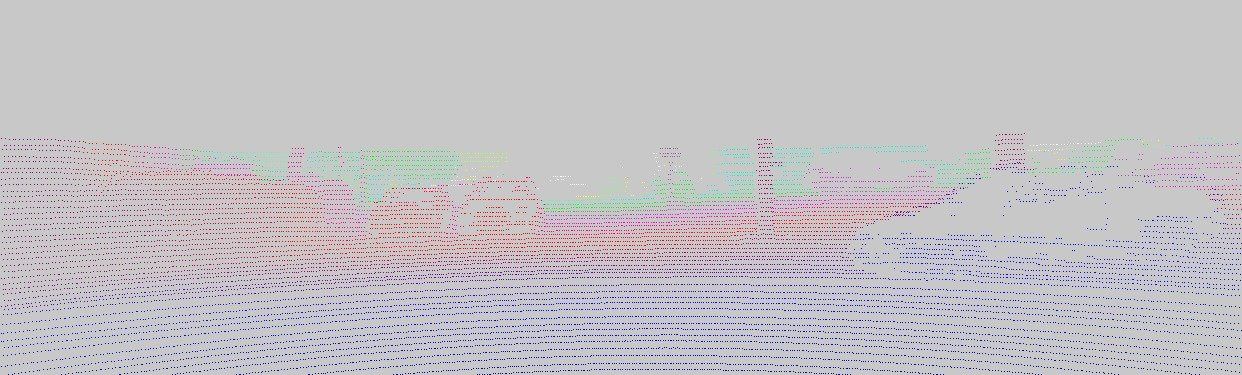}         &
\includegraphics[width=0.194\textwidth, trim={20pt 20pt 20pt 20pt}, clip]{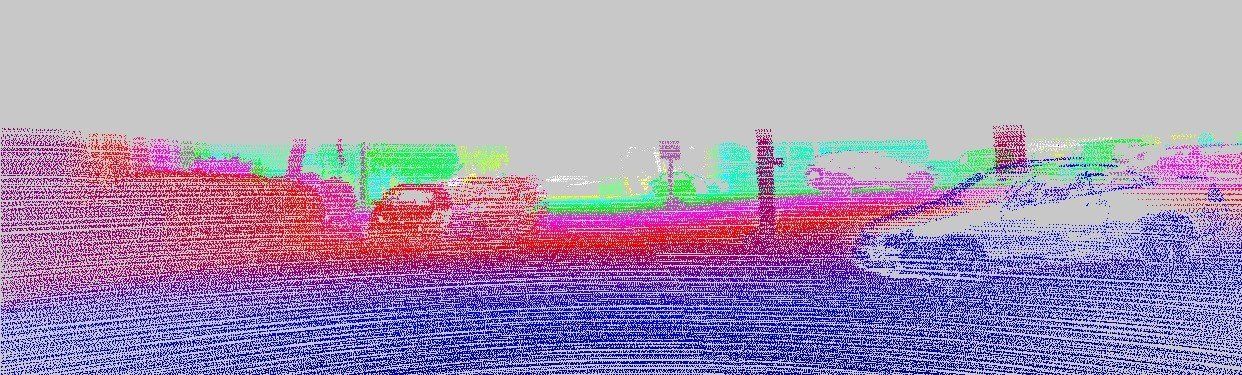}         &
\includegraphics[width=0.194\textwidth, trim={20pt 20pt 20pt 20pt}, clip]{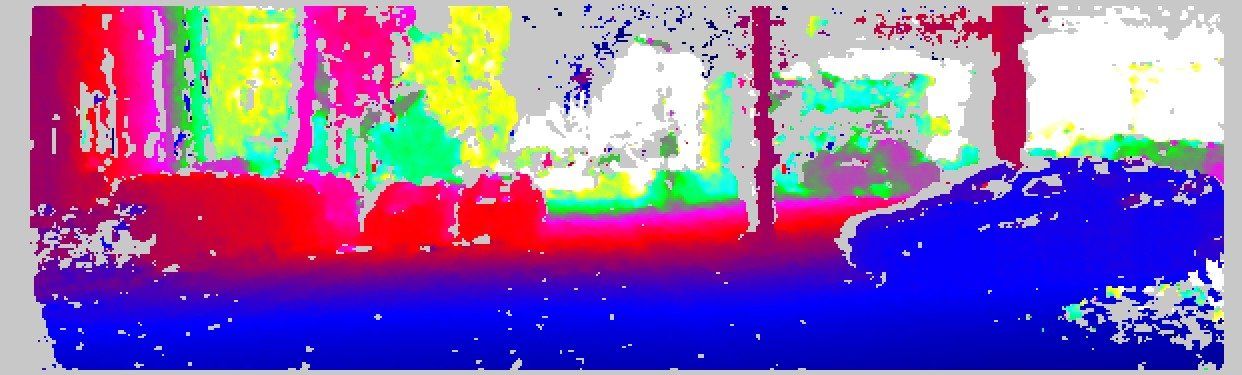}\\ \vspace{-0.1cm}
\includegraphics[width=0.194\textwidth, trim={20pt 20pt 20pt 20pt}, clip]{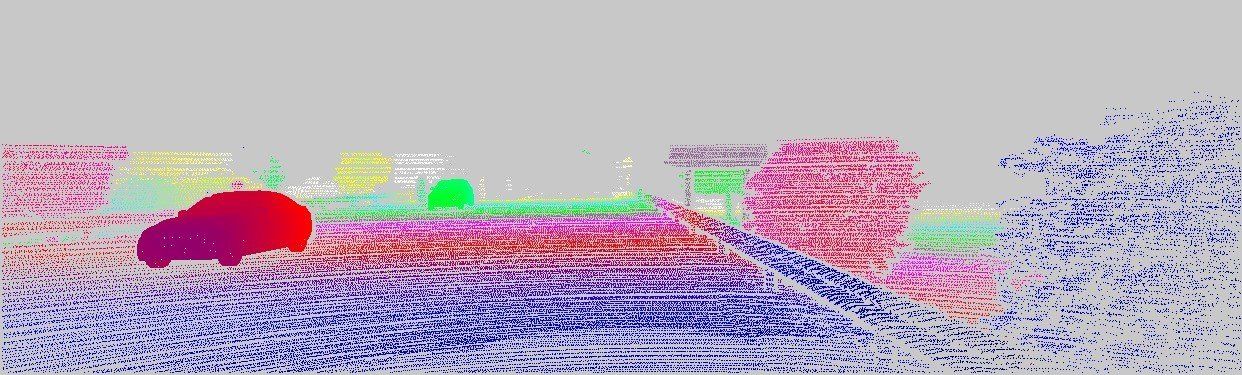}             &
\includegraphics[width=0.194\textwidth, trim={20pt 20pt 20pt 20pt}, clip]{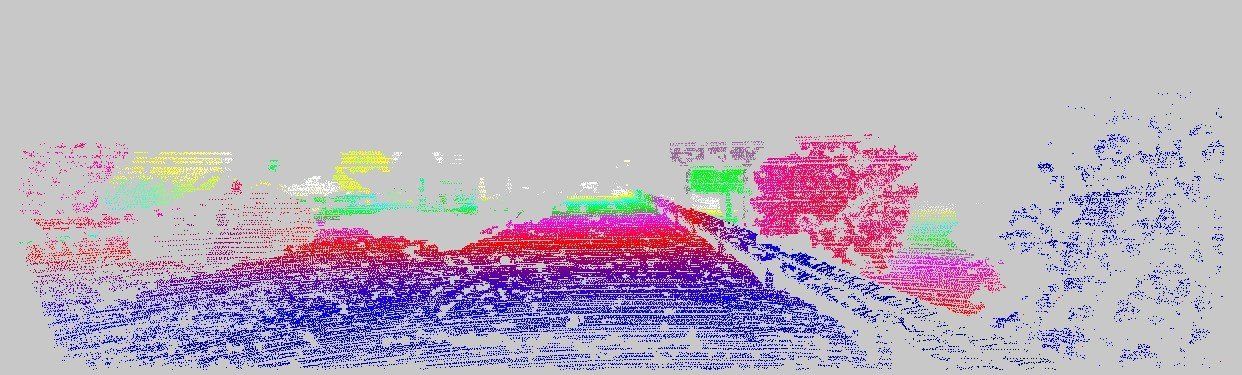}         &
\includegraphics[width=0.194\textwidth, trim={20pt 20pt 20pt 20pt}, clip]{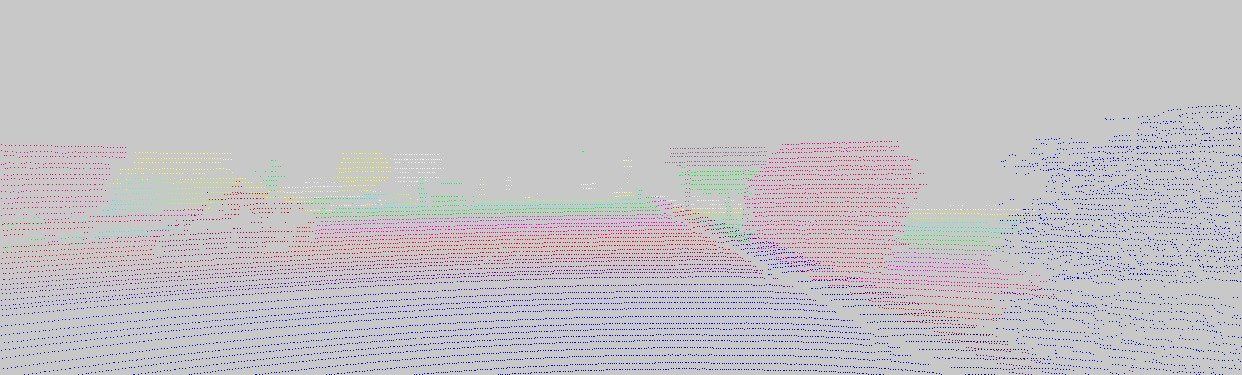}         &
\includegraphics[width=0.194\textwidth, trim={20pt 20pt 20pt 20pt}, clip]{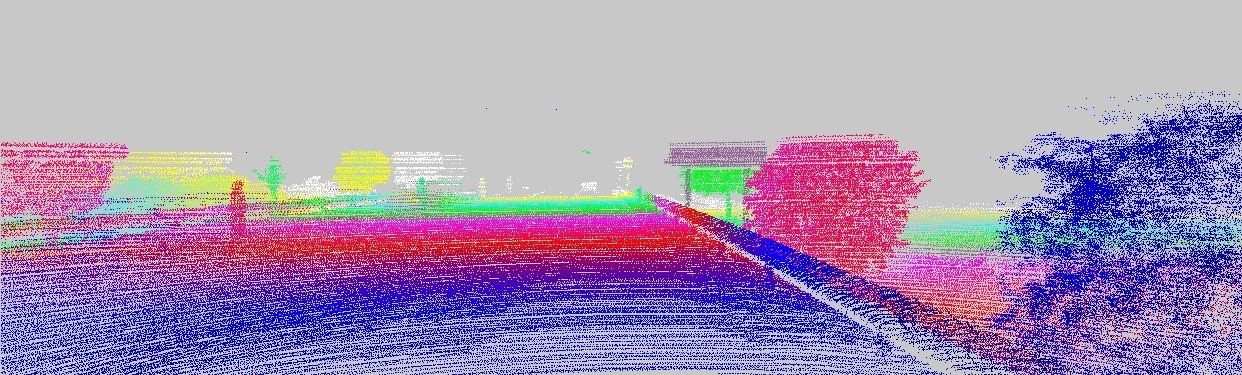}         &
\includegraphics[width=0.194\textwidth, trim={20pt 20pt 20pt 20pt}, clip]{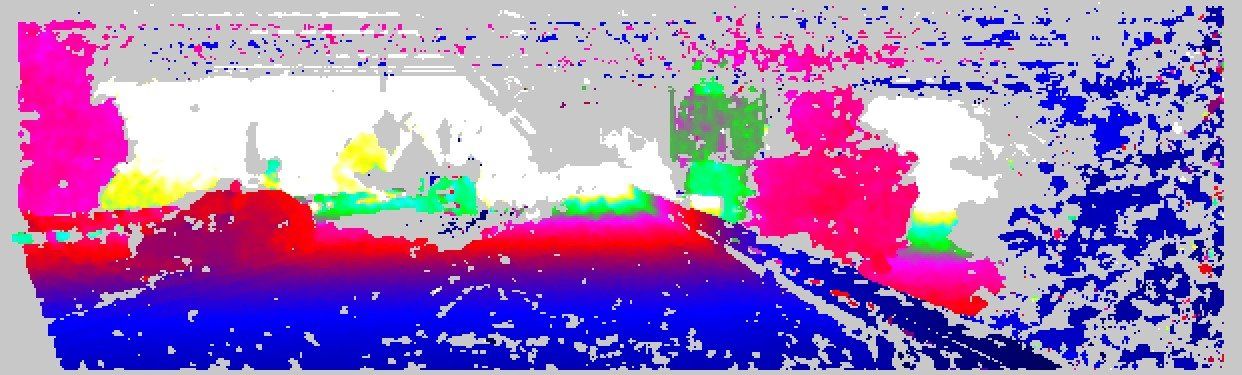} \\ \midrule \vspace{-0.1cm}
\includegraphics[width=0.194\textwidth, trim={20pt 20pt 20pt 20pt}, clip]{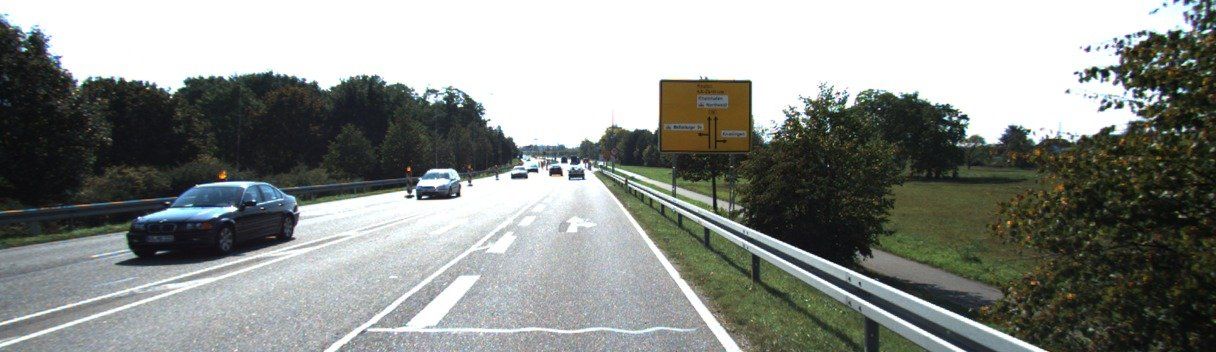}          &
\includegraphics[width=0.194\textwidth, trim={20pt 20pt 20pt 20pt}, clip]{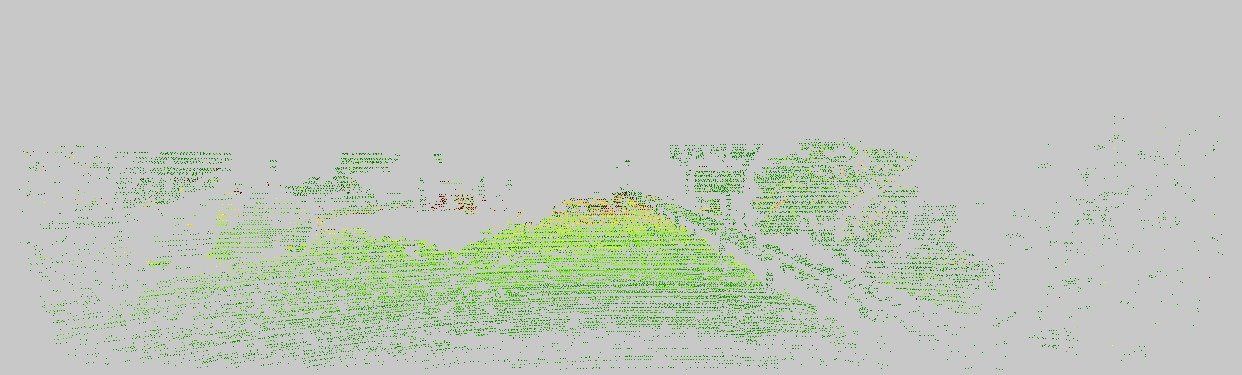} &
\includegraphics[width=0.194\textwidth, trim={20pt 20pt 20pt 20pt}, clip]{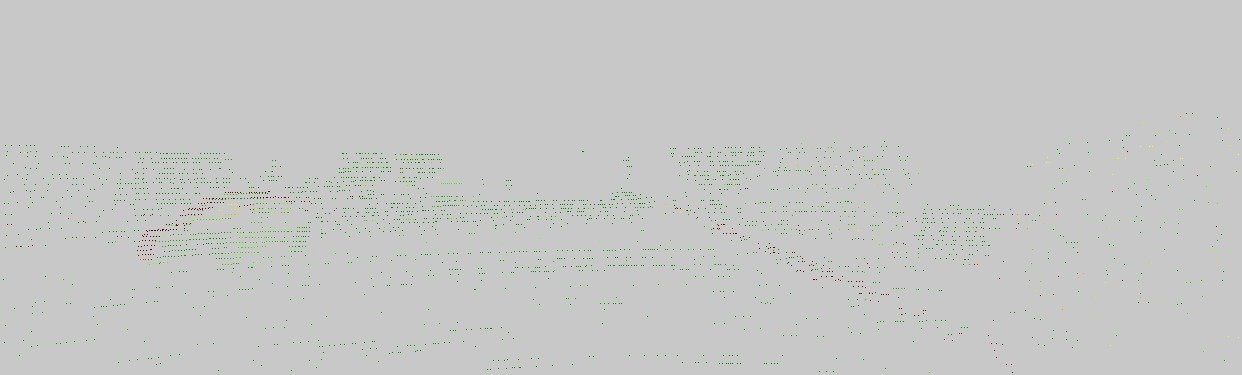}     &
\includegraphics[width=0.194\textwidth, trim={20pt 20pt 20pt 20pt}, clip]{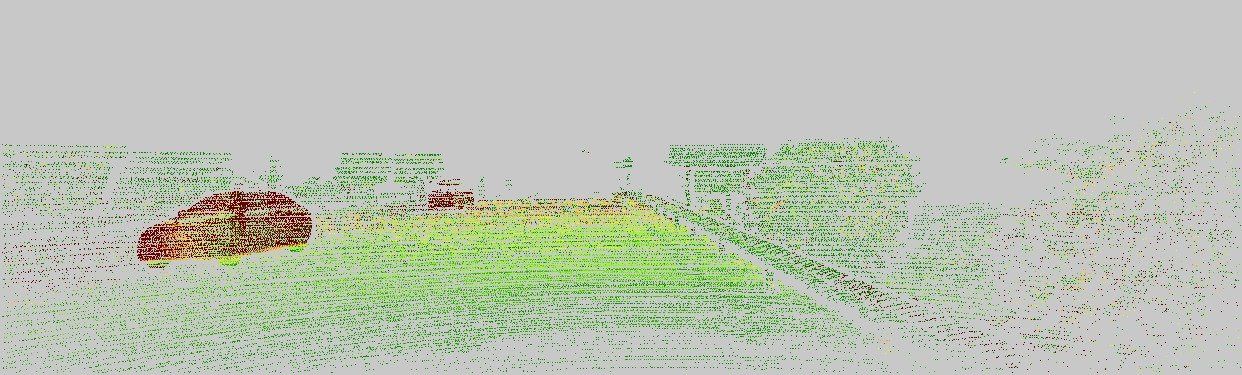}     &
\includegraphics[width=0.194\textwidth, trim={20pt 20pt 20pt 20pt}, clip]{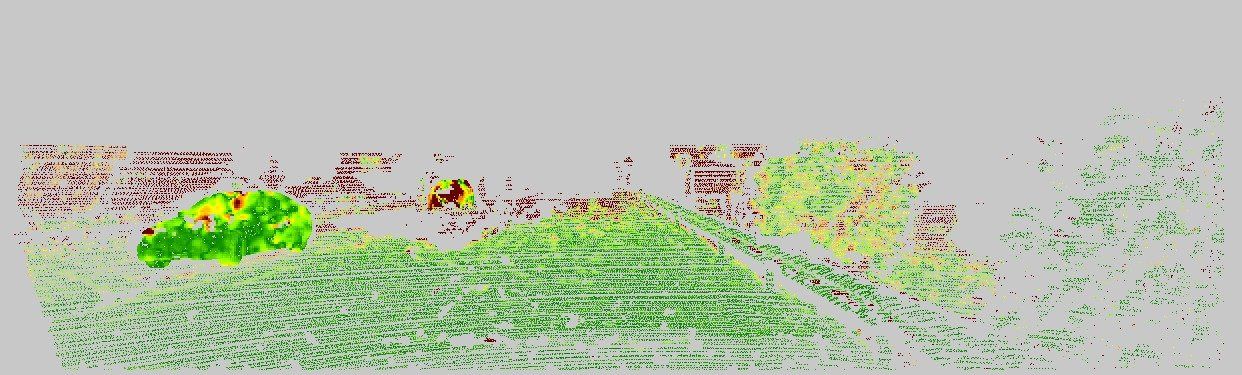}\\
RGB Image & \multicolumn{4}{c}{Error Maps wrt. KITTI 2015 \cite{Menze2015CVPR}}
\end{tabular}
\end{center}
\vspace{-0.3cm}
\caption{%
\textbf{Large-scale Dataset.} Qualitative results of our depth annotated dataset. From left to right we compare:
depth maps of the manually curated KITTI 2015 dataset, our automatically generated data, raw and accumulated LiDaR scans,
and SGM \cite{Hirschmueller2008PAMI} results. Differences to the KITTI 2015 depth maps are shown in the last row from 0 (green) to 2 (red) meters.
}
\label{fig:data_eval}
\end{figure*}

\subsection{Dataset Evaluation}
\label{subsec:dataset_eval}

Before using the proposed dataset for evaluation in \secref{sec:experiments}, we verify its quality.
Towards this goal, we exploit the manually cleaned training set of the KITTI 2015 stereo benchmark as reference data.
We compute several error
measures for our generated depth maps using the provided depth ground truth and compare ourself to the raw
and accumulated LiDaR scans as well as the SGM depth maps in \tabref{tab:data_eval}. The SGM reconstruction
is very dense but also rather inaccurate compared to the raw laser scans.
In terms of mean
absolute error (MAE) our dataset reaches approximately the same accuracy level as the raw LiDaR scans.
However, for the metrics ``root mean squared error (RMSE)'',  ``KITTI outliers'' (disparity error $\geq3$px
and $\geq5\%$), as well as the $\delta$ inlier ratios (maximal mean relative error of $\delta_i = 1.25^i$ for
$i \in \{1,2,3\}$), our dataset outperforms all baseline results.
At the same time, we achieve four times denser depth maps than raw LiDaR scans.
A qualitative comparison is presented in \figref{fig:data_eval}.

After manually separating the foreground and background regions on the benchmark
depth maps, we evaluate the errors present on dynamic objects and background in Table \ref{tab:data_eval_foreground}.
The result indicates that our proposed accumulation and clean-up pipeline is able to remove outliers
in the raw LiDaR scans and at the same time significantly increases the density of the data.
Qualitatively, we find only little errors in our dataset. Most of the remaining errors are located on dynamic objects or
at high distances, \cf \figref{fig:data_eval} (bottom). In comparison, SGM results are inaccurate at large
distances and LiDaR scans result in occlusion errors due to the different placement of the LiDaR sensor and the virtual camera used for projection (we use the image plane of the KITTI reference camera for all our experiments). Note that dynamic objects (\eg, car on the left) lead to significant errors in the accumulated LiDaR scans which are largely reduced with our technique.

\setlength{\tabcolsep}{3.5pt}
\ctable[
    caption = { Evaluation of reference depth maps using the manually
                curated ground truth depth maps of the KITTI 2015 training set \cite{Menze2015CVPR}.
                Note that our dataset is generated fully automatically and achieves highest
                accuracy while providing high density. All metrics are computed in
                disparity space.},
    label   = {tab:data_eval},
    pos     = {tb},
    doinside= \scriptsize,
    width   = 0.45\textwidth,
    mincapwidth = \linewidth
]{Xc|cccccc}{}{
        \FL
                                   & \multirow{2}{*}{Density} & MAE & RMSE & KITTI         & \multicolumn{3}{c}{$\delta_i$ inlier rates}      \\
                                   &                          &        [px]              &       [px]                & outliers      & $\delta_1$     & $\delta_2$     & $\delta_3$     \ML
        SGM                        &         82.4\%           &         1.07         &         2.80          &         4.52  &         97.00  &         98.67  &         99.19  \NN
        Raw LiDaR                  &          4.0\%           & \textbf{0.35}        &         2.62          &         1.62  &         98.64  &         99.00  &         99.27  \NN
        Acc. LiDaR                 &         30.2\%           &         1.66         &         5.80          &         9.07  &         93.16  &         95.88  &         97.41  \NN
        Our Dataset                &         16.1\%           & \textbf{0.35}        & \textbf{0.84}         & \textbf{0.31} & \textbf{99.79} & \textbf{99.92} & \textbf{99.95} \LL
}

\setlength{\tabcolsep}{3pt}
\ctable[
    caption = {Evaluation of \tabref{tab:data_eval} split according to foreground (car) / background (non-car) regions. },
    label   = {tab:data_eval_foreground},
    pos     = {tb},
    doinside= \scriptsize,
    width   = 0.47\textwidth,
    mincapwidth = \linewidth
]{Xcccccc}{}{
        \FL
        \multirow{2}{*}{Depth Map} & MAE      & RMSE      & KITTI                     & \multicolumn{3}{c}{$\delta_i$ inlier rates}                                             \\
                                   &      [px]                     &          [px]                  & outliers                  & $\delta_1$                  & $\delta_2$                  & $\delta_3$                  \ML
        SGM                        &          1.2/1.1          &           3.0/2.8          &          5.9/4.4          &         97.6 /96.9          &         98.2 /98.7          &       98.5/99.3             \NN
        Raw LiDaR                  &          3.7/\textbf{0.2} &          10.0/1.9          &         17.4/0.9          &         84.3 /99.3          &         86.1 /99.6          &       88.6/99.7             \NN
        Acc. LiDaR                 &          7.7/1.1          &          12.0/4.8          &         59.7/4.3          &         55.7 /96.7          &         73.7 /98.0          &       83.0/98.8             \NN
        Our Dataset                & \textbf{0.9}/0.3          &  \textbf{2.2}/\textbf{0.8} & \textbf{3.0}/\textbf{0.2} & \textbf{98.6}/\textbf{99.8} & \textbf{99.0}/\textbf{99.9} & \textbf{99.3}/\textbf{99.9} \LL
}

For our experimental evaluation in the next section, we split our dataset into 85k images for training, 3k images for testing and 4k images for validation.
For all splits we
ensure a similar distribution over KITTI scene categories (city, road, residential and campus) while
keeping the sequence IDs unique for each split to avoid overfitting to nearby frames.
On acceptance of this paper, we will setup an online evaluation server for evaluating the performance
of depth upsampling and depth estimation algorithms.

In the following section, we leverage this dataset
for our depth upsampling experiments using the proposed sparse CNN, which is the main focus of this paper.

\externaldocument{method}

\section{Experiments}
\label{sec:experiments}

\subsection{Depth Upsampling}

\setlength\tabcolsep{3pt}
\begin{figure*}[ht]
	\centering
	\footnotesize
	\noindent\begin{tabular}{ccc}
		\includegraphics[width=0.32\textwidth]{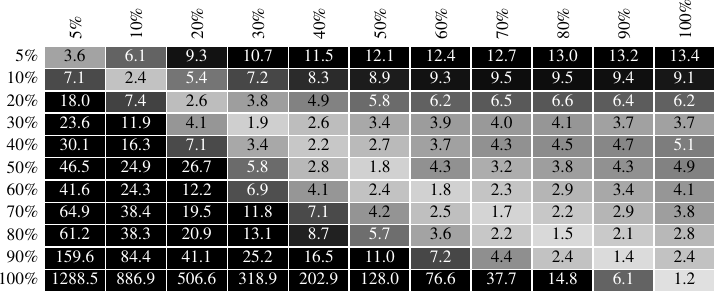}~&
		\includegraphics[width=0.32\textwidth]{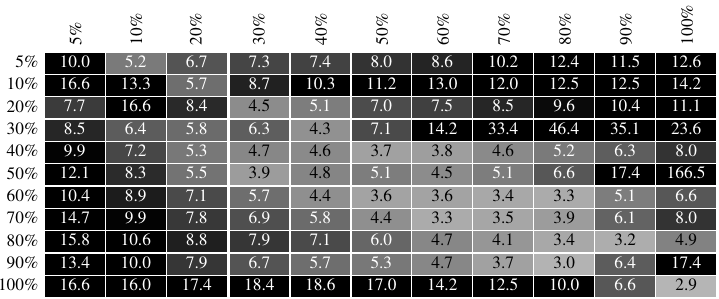}~&
		\includegraphics[width=0.32\textwidth]{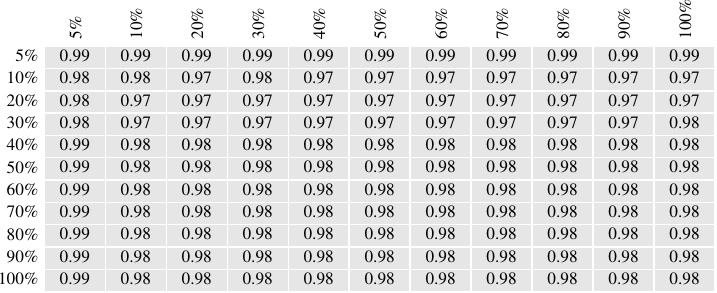} \\
		(a) ConvNet & (b) ConvNet + mask & (c) SparseConvNet
	\end{tabular}
	\arrayrulecolor{black}
	\vspace{-0.2cm}
	\caption{Comparison of three different networks on the Synthia dataset \cite{Ros2016CVPR} while varying the sparsity level of the training split (left) {\it and} the sparsity of the test split (top). From left-to-right: ConvNet,
		ConvNet with concatenated validity mask and the proposed SparseConvNet. All numbers represent mean average errors (MAE).}
	\label{fig:sparsity_eval}
\end{figure*}
\setlength\tabcolsep{6pt}

\setlength\tabcolsep{0pt}
\newcolumntype{Y}{>{\centering\arraybackslash}X}
\begin{figure*}[ht]
	\begin{center}
		\footnotesize
		\centering
		\begin{tabularx}{\textwidth}{YYYYY}
			\includegraphics[width=0.195\textwidth]{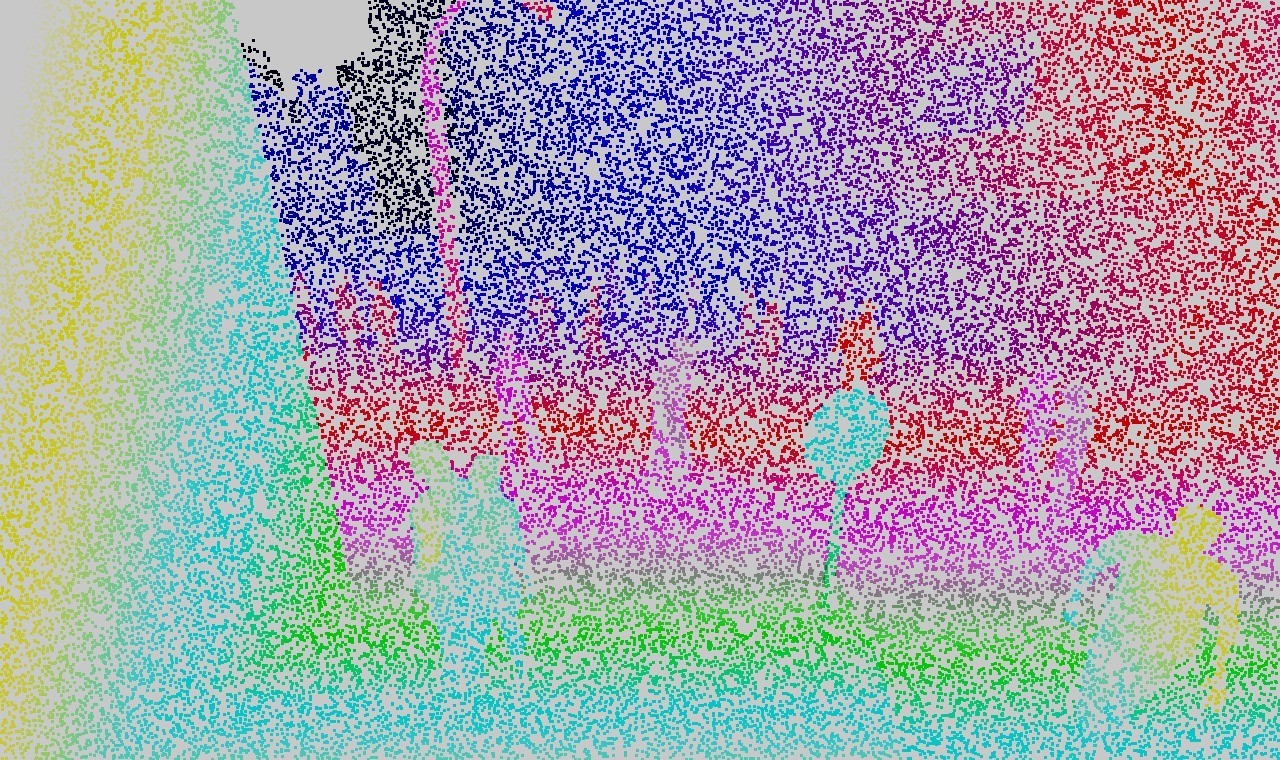} &
			\includegraphics[width=0.195\textwidth]{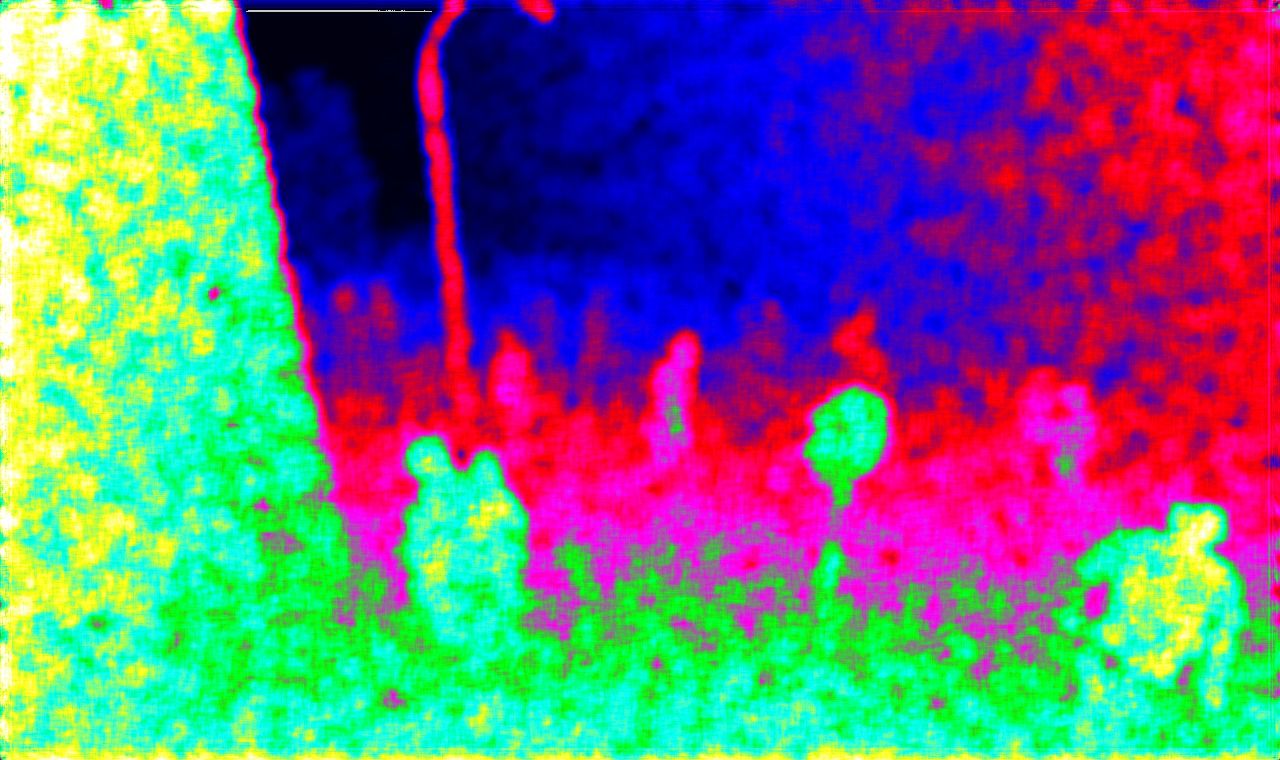} &
			\includegraphics[width=0.195\textwidth]{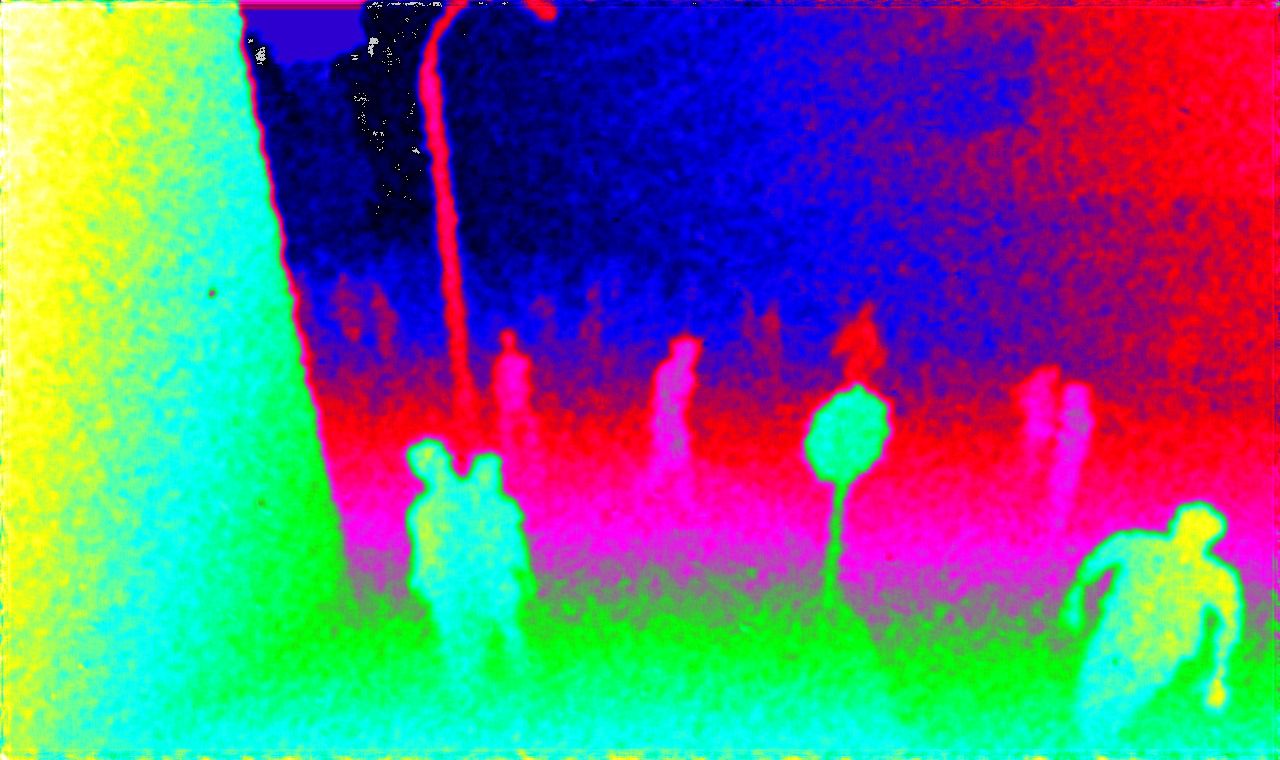} &
			\includegraphics[width=0.195\textwidth]{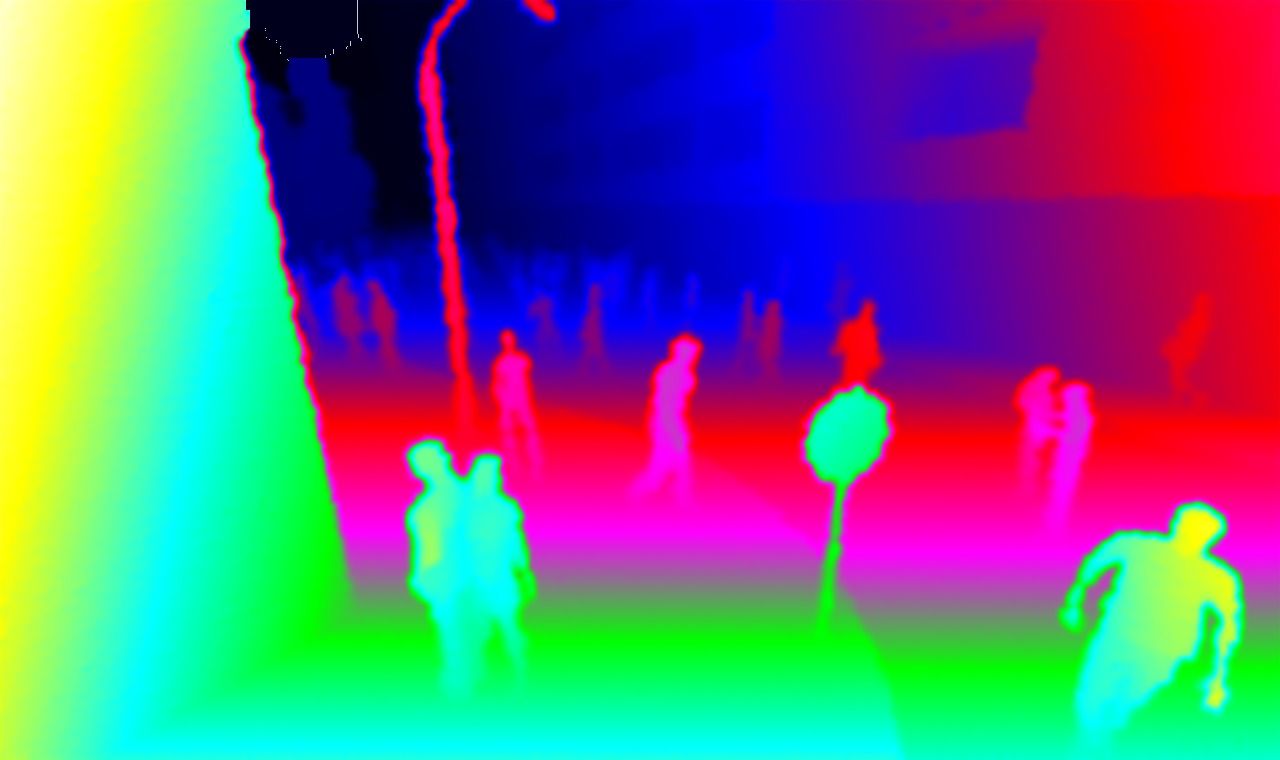} &
			\includegraphics[width=0.195\textwidth]{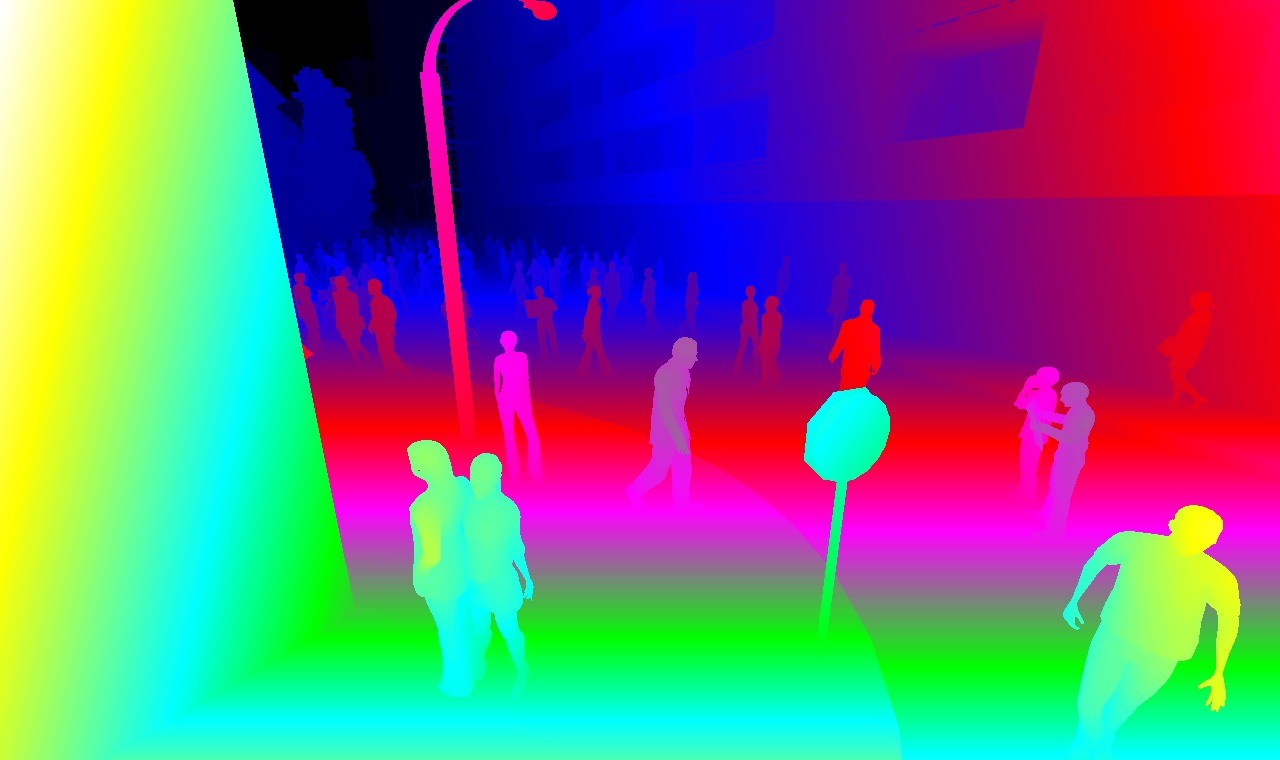}
			\\
			\includegraphics[width=0.195\textwidth]{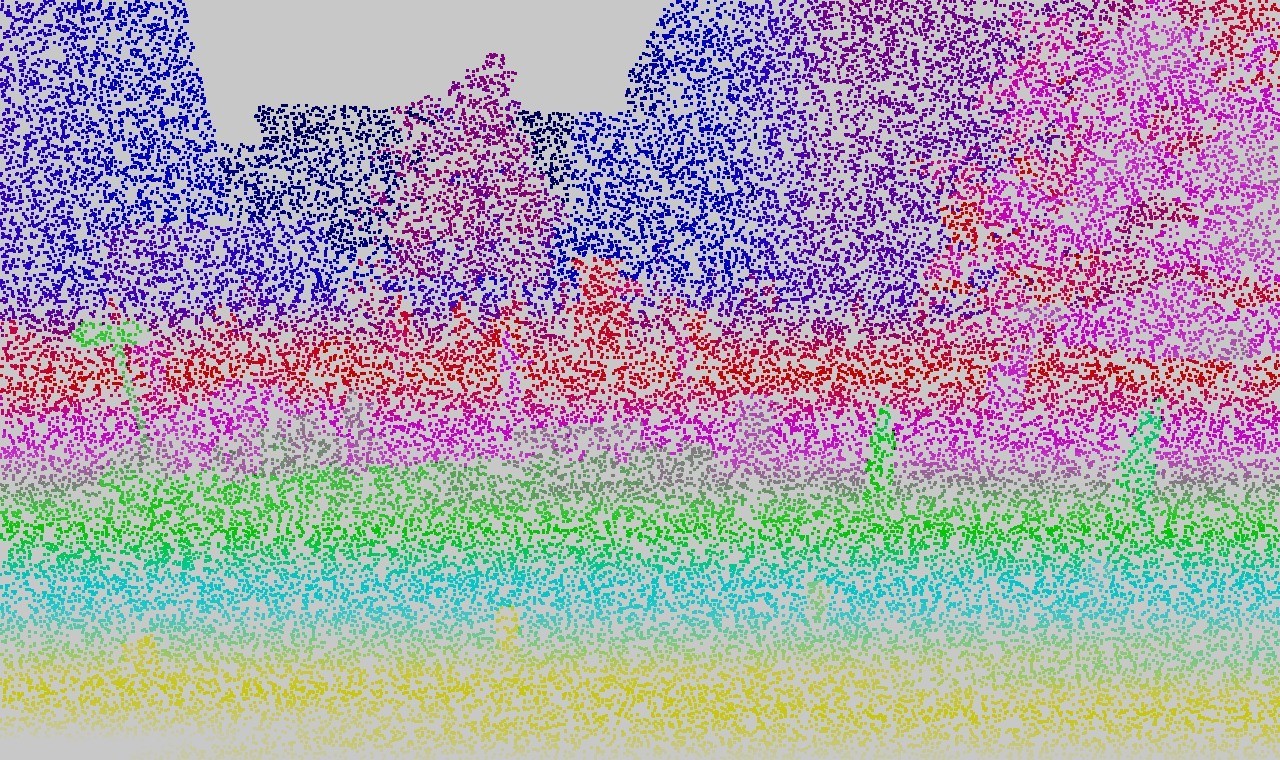} &
			\includegraphics[width=0.195\textwidth]{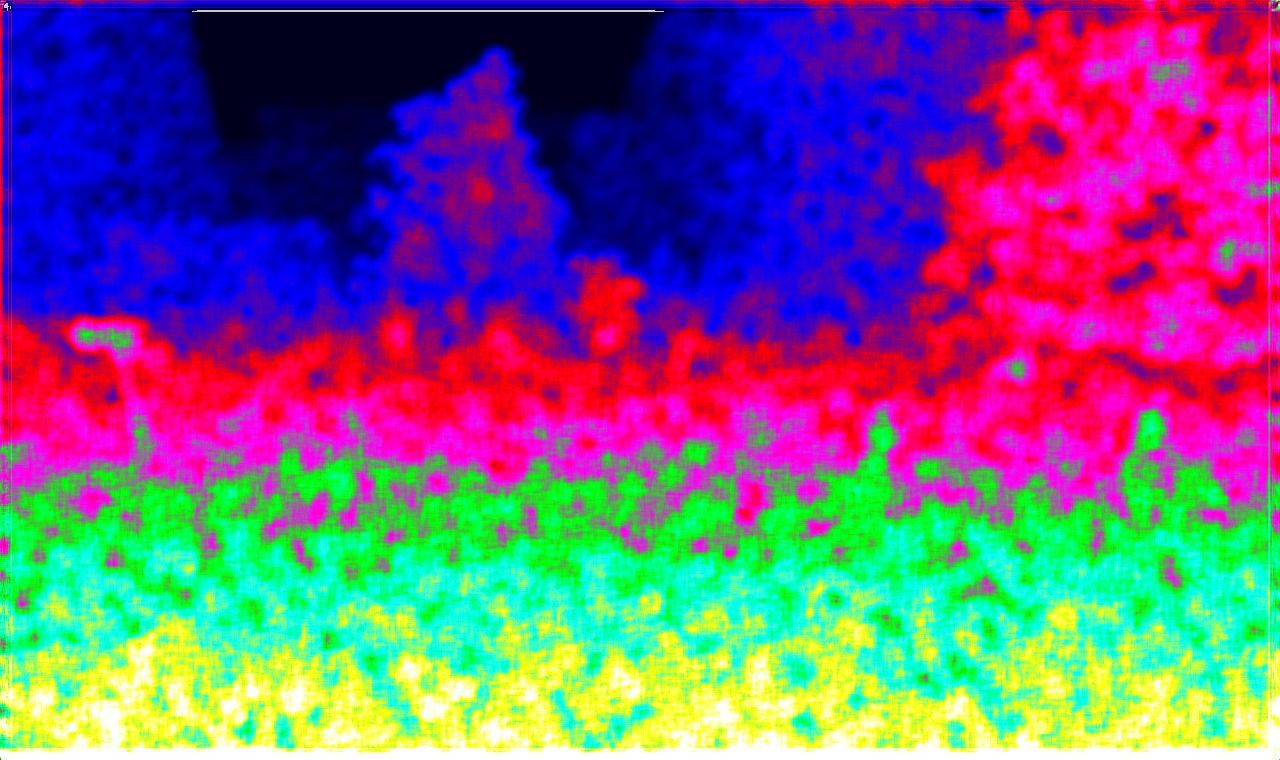} &
			\includegraphics[width=0.195\textwidth]{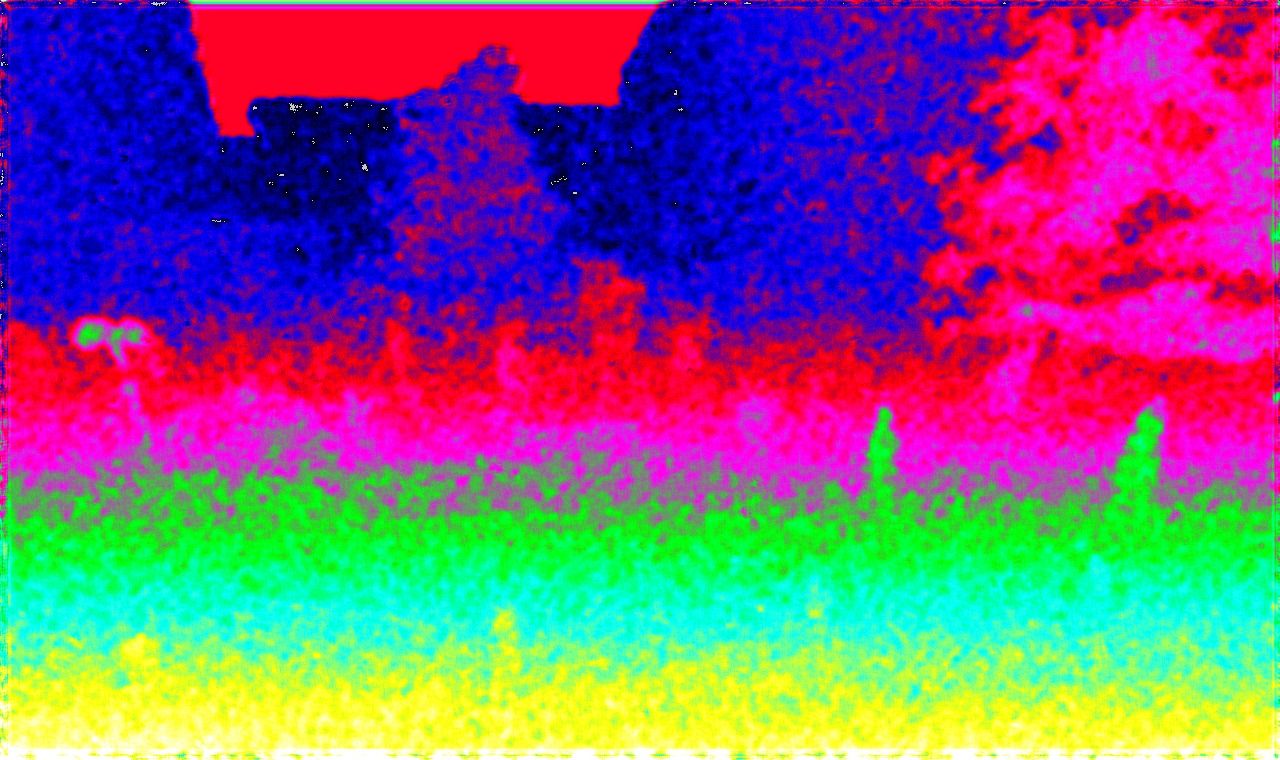} &
			\includegraphics[width=0.195\textwidth]{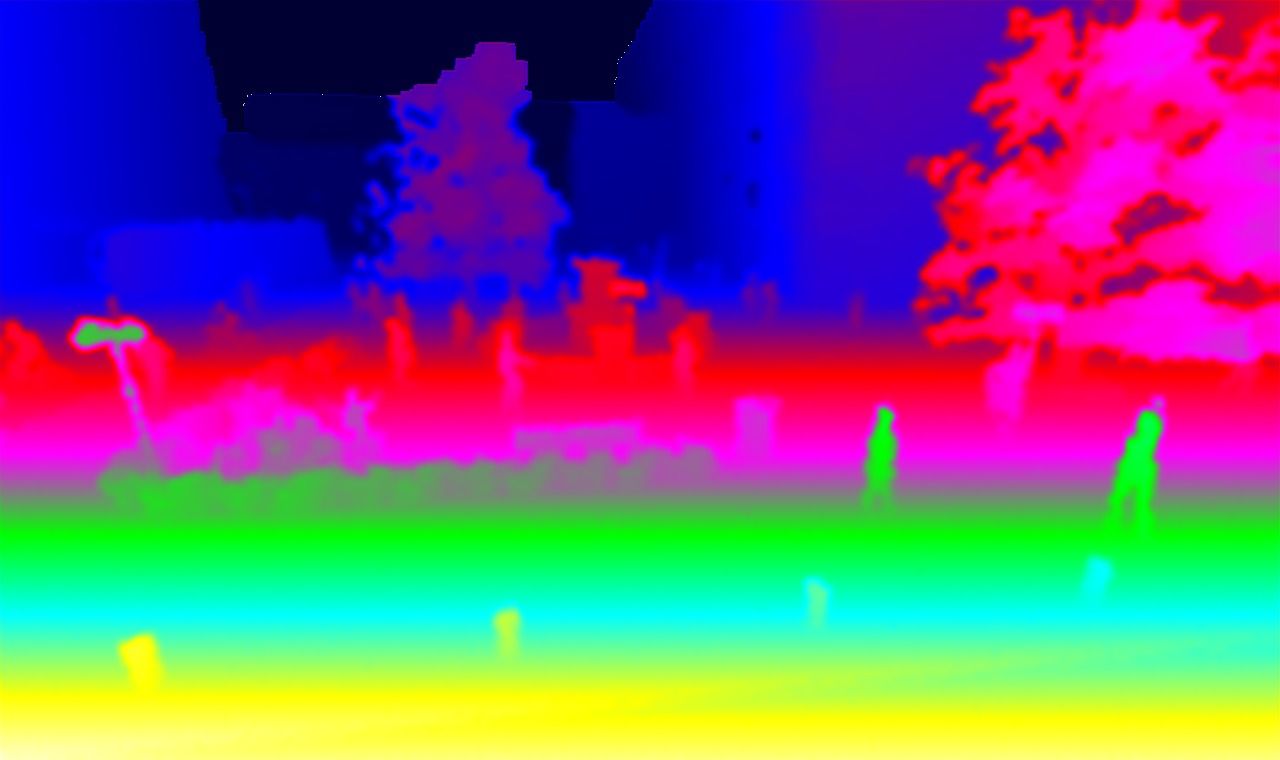} &
			\includegraphics[width=0.195\textwidth]{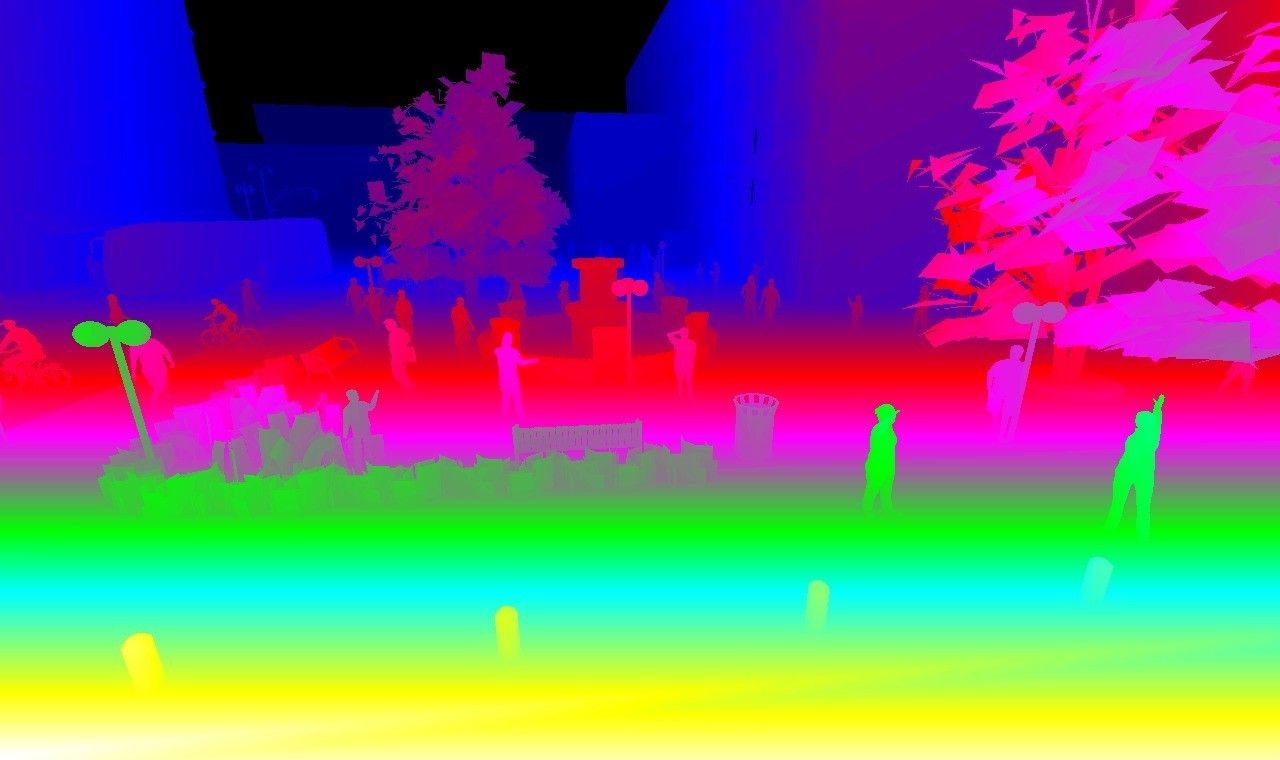}\\
			(a) Input (visually enhanced) &
			(b) ConvNet &
			(c) ConvNet + mask &
			(d) SparseConvNet (ours) &
			(e) Groundtruth
		\end{tabularx}
	\end{center}
	\vspace{-0.3cm}
	\caption{Qualitative comparison of our sparse convolutional network to standard ConvNets on Synthia \cite{Ros2016CVPR}, trained and evaluated at 5\% sparsity.
		(b) Standard ConvNets suffer from large invalid regions in the input leading to noisy results.
		(c) Using a valid mask as input reduces noise slightly.
		(d) In contrast, our approach predicts
		smooth and accurate outputs.}
	\label{fig:sparsity_comparison}
\end{figure*}

\begin{figure}[ht]
	\begin{center}
		\footnotesize
		\centering
		\begin{tabularx}{0.48\textwidth}{YYY}
			\includegraphics[width=0.155\textwidth]{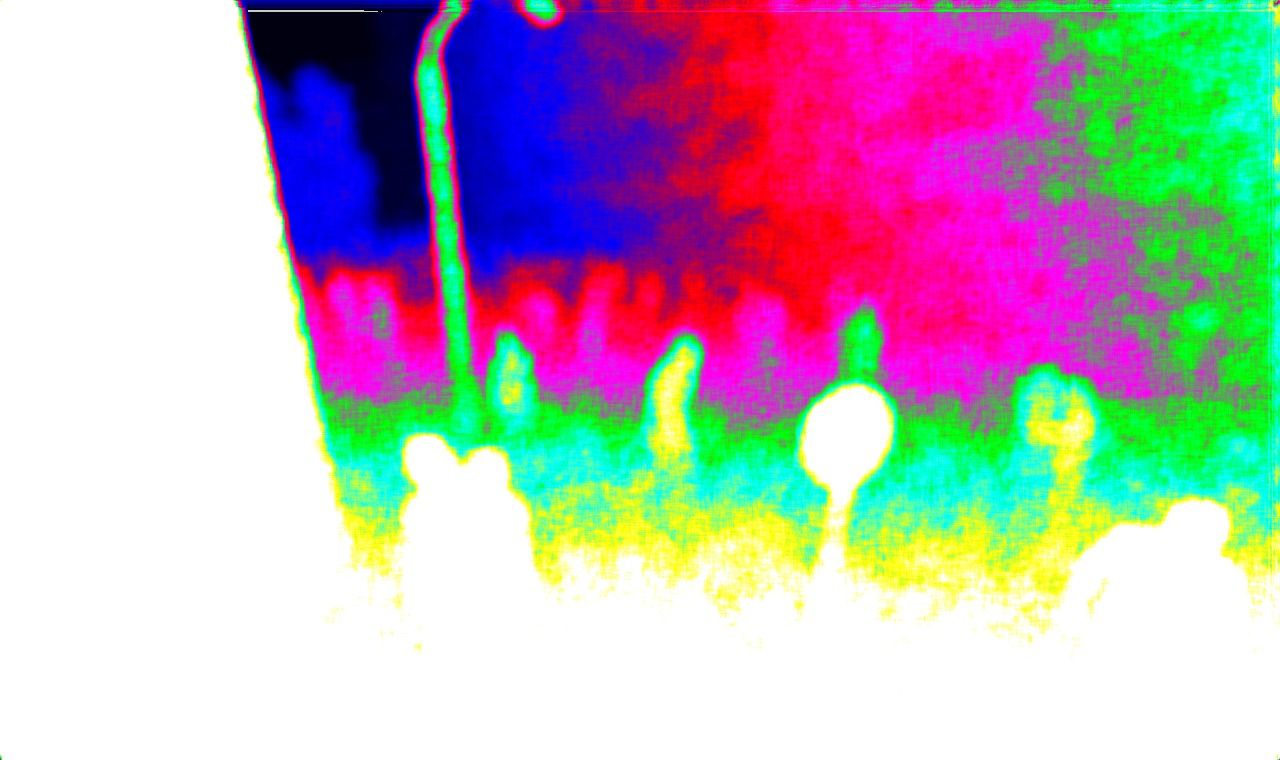} &
			\includegraphics[width=0.155\textwidth]{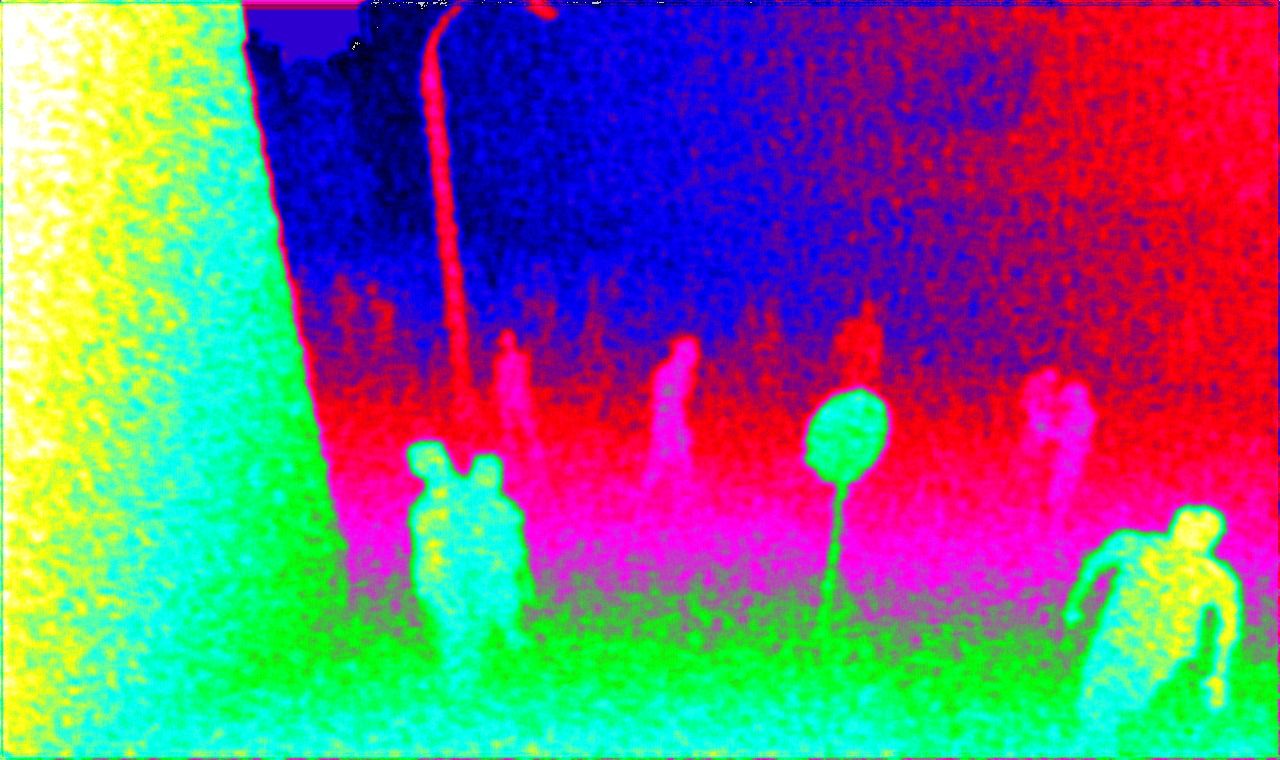} &
			\includegraphics[width=0.155\textwidth]{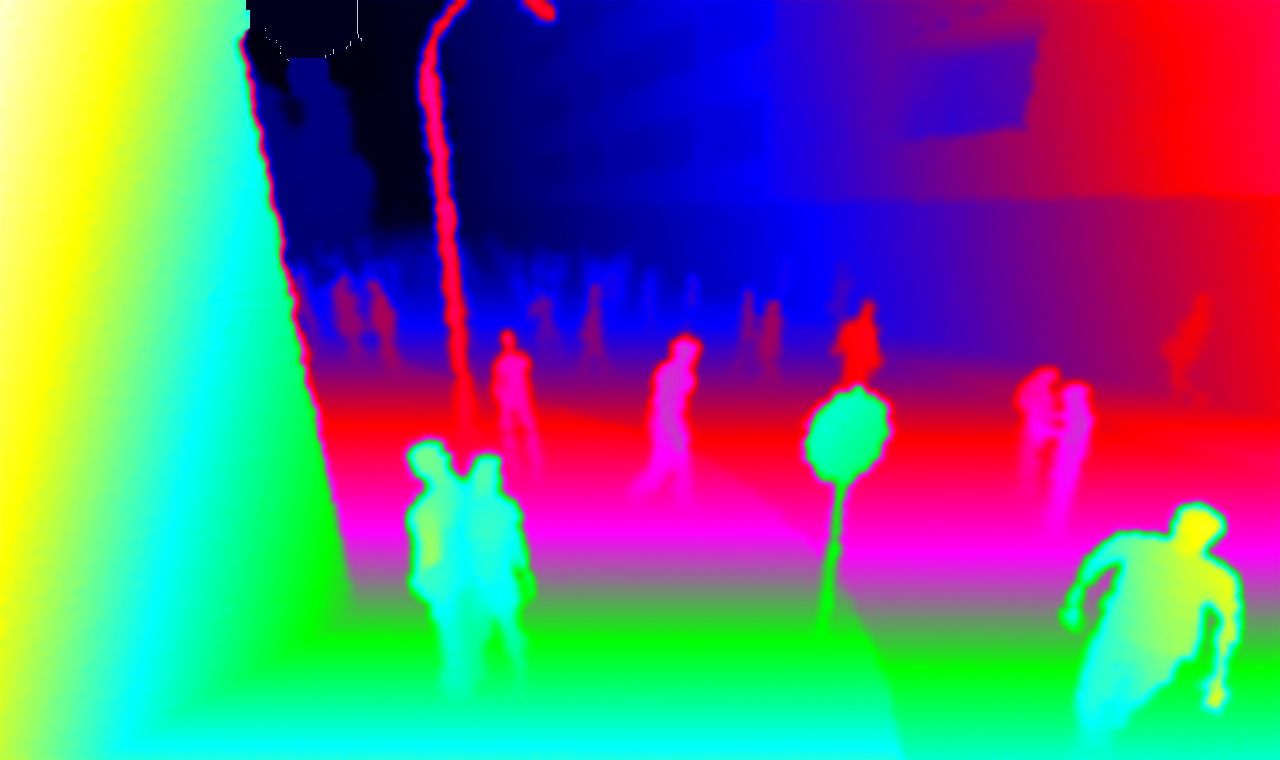}\\
			\includegraphics[width=0.155\textwidth]{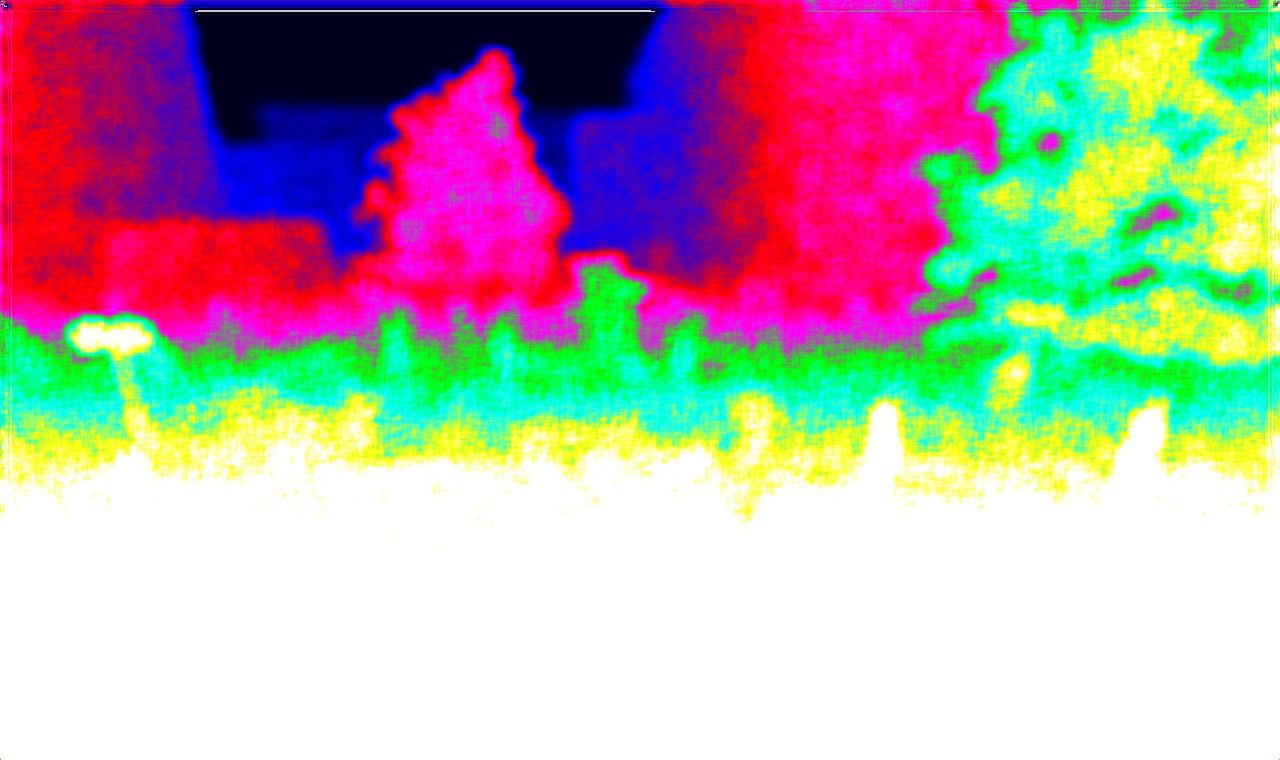} &
			\includegraphics[width=0.155\textwidth]{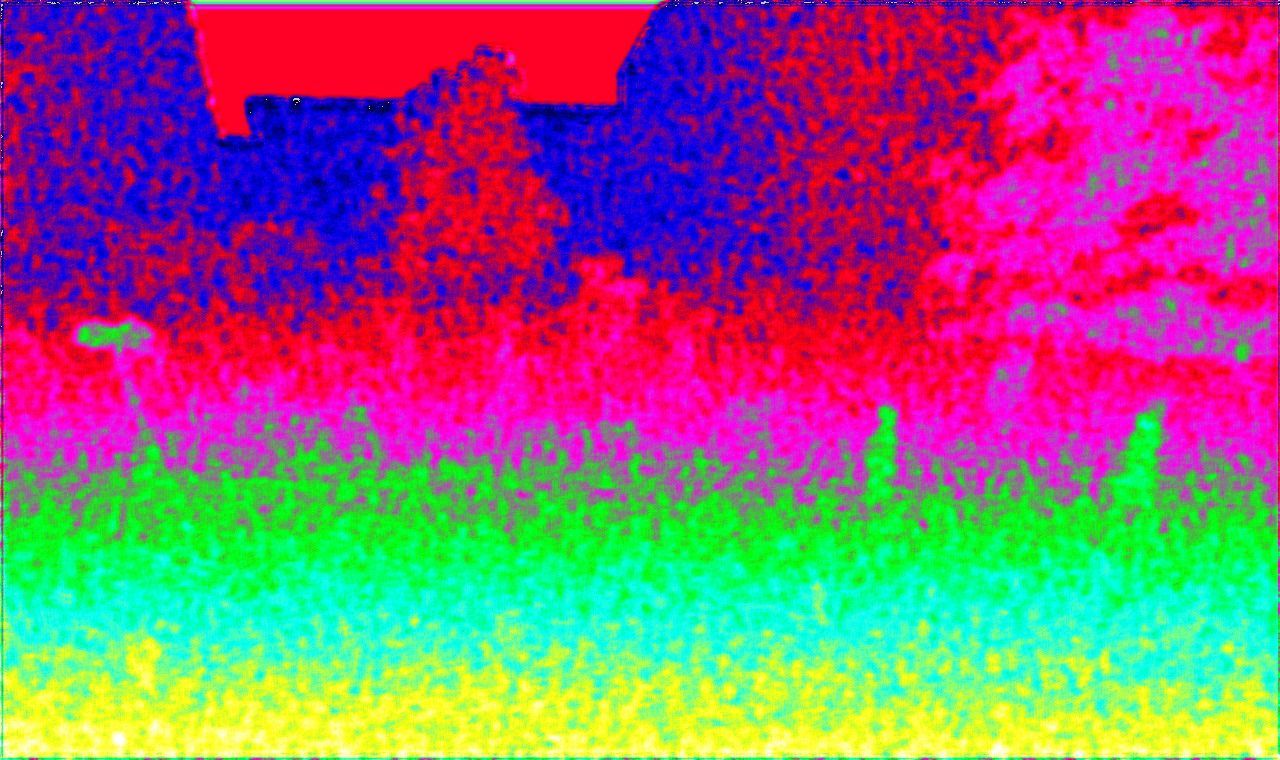} &
			\includegraphics[width=0.155\textwidth]{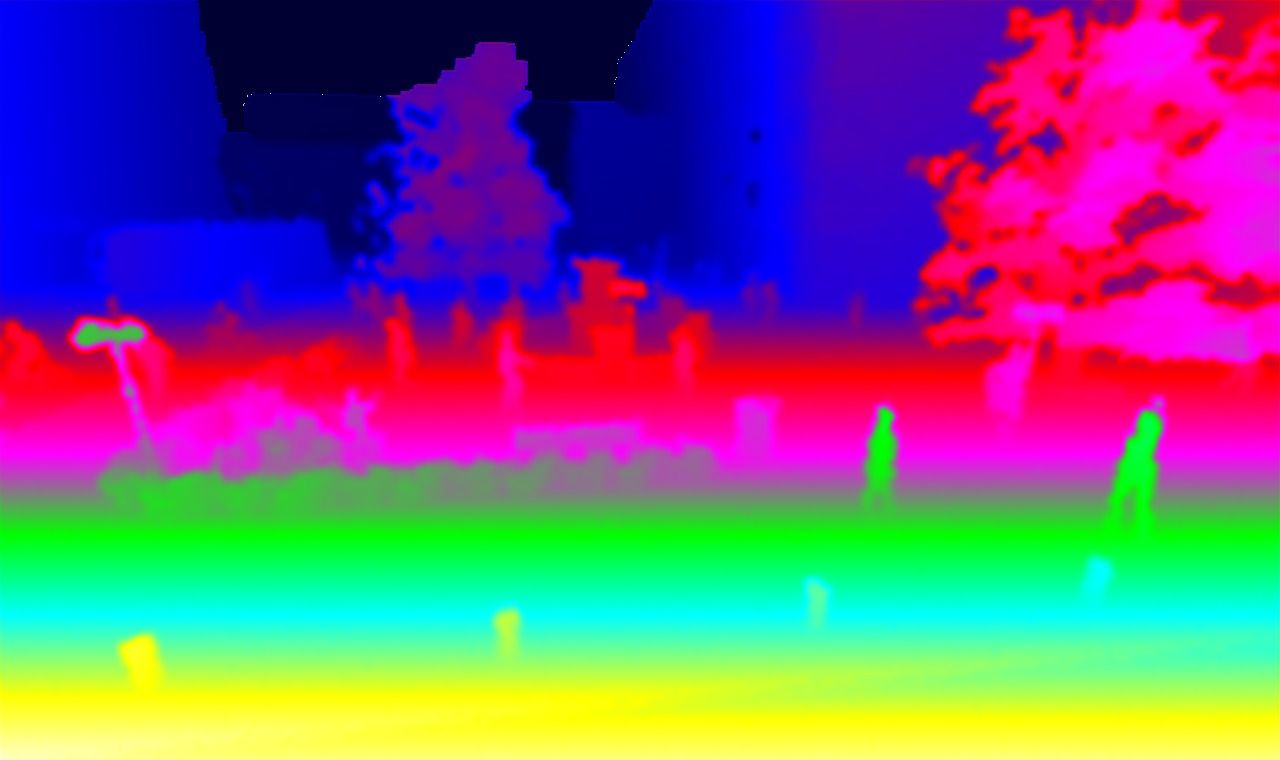}\\
			(a) ConvNet &
			(b) ConvNet + mask &
			(c) SparseConvNet
		\end{tabularx}
	\end{center}
	\vspace{-0.3cm}
	\caption{Network predictions for scenes in Figs. \ref{fig:illustration} and
		\ref{fig:sparsity_comparison}, with all
		networks trained at 5\% sparsity and evaluated at 20\% sparsity. While ConvNets with and without visibility mask
		produce substantially worse results, the results of the proposed sparsity invariant CNN do not degrade.}
	\label{fig:sparsity_comparison_density}
\end{figure}

We investigate the task of depth map completion to evaluate the effect of sparse input data for our \textit{Sparse
Convolution Modules}. For this task, a sparse, irregularly populated depth map from a projected laser scan is
completed to full image resolution without any RGB guidance.

We first evaluate the performance of our method with varying degrees of sparsity in the input.
Towards this goal, we leverage the Synthia dataset of Ros \etal \cite{Ros2016CVPR} which gives us full control over the sparsity level.
To artificially adjust the sparsity of the input, we apply random dropout to the provided dense
depth maps during training.
The probability of a pixel to be dropped is set to different levels ranging from 0\% to 95\%.

We train three different variants of a Fully Convolutional Network (FCN)
with five convolutional layers of kernel size 11, 7, 5, 3, and 3.
Each convolution has a stride of one, 16 output channels, and is followed by a ReLU as nonlinear activation function.
The three variants we consider are: i) plain convolutions with only sparse depth as input, ii) plain convolutions
with sparse depth and concatenated valid pixel map as input, and iii) the proposed
\textit{Sparse Convolution Layers}, \cf \figref{fig:sparse_convolution_network}.
We train separate networks for various levels of sparsity
using the Synthia \textit{Summer} sequences, whereas evaluation is performed on the Synthia \textit{Cityscapes} dataset.
To compare the performance of the different approaches we
first evaluate them on the sparsity level they have been trained on.
To test the generalization ability of the different
models we further apply them to sparsity levels which they have not seen during training.

\figref{fig:sparsity_eval} shows our results. We observe that plain
convolutions perform poorly with very sparse inputs as all pixels (valid and invalid) are considered
in the convolution. This introduces a large degree of randomness during training and testing and results in strong variations in performance.
Convolutions on sparse depth maps with the concatenated valid mask perform slightly better than using only the depth input.
However, in contrast to our \textit{Sparse Convolutions} they perform poorly, especially on very sparse input.

Invariance to the level of sparsity is an important property for depth upsampling methods as it increases
robustness towards random perturbations in the data.
Besides, this property allows to generalize to different depth sensors
such as structured light sensors, PMD cameras or LiDaR scanners.
As evidenced by \figref{fig:sparsity_eval}, all methods perform reasonably well at the performance level they have been trained on (diagonal entries) with the sparse convolution variant performing best.
However, both baselines fail completely in predicting depth estimates on more sparse and, surprisingly,
also on more dense inputs.
In contrast, our proposed \textit{Sparse Convolution Network} performs equally well across all
levels of sparsity no matter which sparsity level has been observed during training.
This highlights the generalization ability of our approach.
\figref{fig:sparsity_comparison} shows a qualitative
comparison of the generated dense depth maps for the two baselines and our approach
using 5\% sparsity during training and testing.
Note that the input in \figref{fig:sparsity_comparison} (a) has been visually enhanced using dilation to improve readability.
It thus appears more dense than the actual input to the networks.
For the same examples, \figref{fig:sparsity_comparison_density} shows the drastic drop in performance when training standard CNNs on 5\% and
evaluating on 20\%, while our approach performs equally well. While ConvNets with input masks lead to noisy results, standard ConvNets even result in a systematic bias as they are unaware of the level of sparsity in the input.

\subsubsection{Synthetic-to-Real Domain Adaptation}

\begin{figure*}
\begin{center}
\begin{subfigure}{0.32\linewidth}
\includegraphics[width=\textwidth,trim= 100 0 140 100, clip]{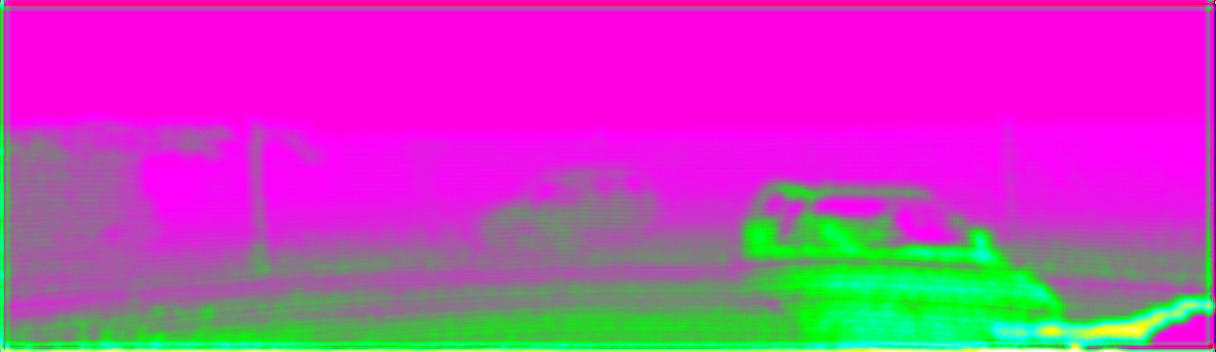}~%
\caption{ConvNet}
\end{subfigure}
\begin{subfigure}{0.32\linewidth}
\includegraphics[width=\textwidth,trim= 100 0 140 100, clip]{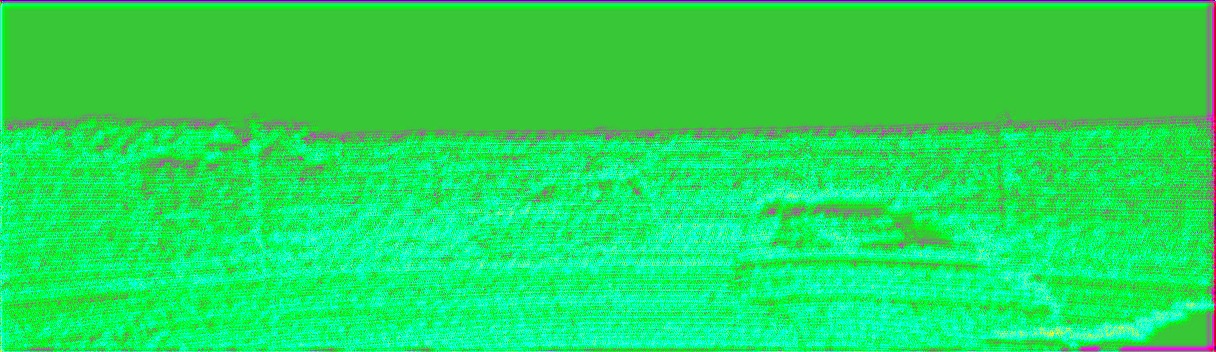}~%
\caption{ConvNet + mask}
\end{subfigure}
\begin{subfigure}{0.32\linewidth}
\includegraphics[width=\textwidth,trim= 100 0 140 100, clip]{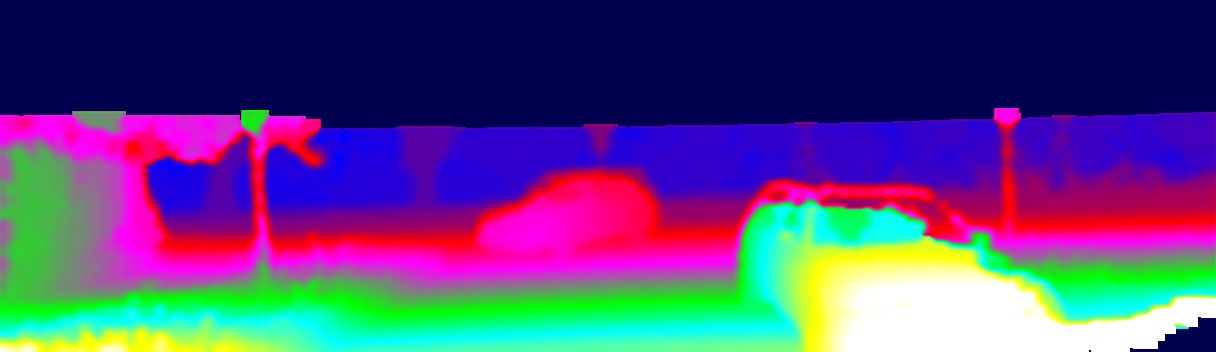}
\caption{SparseConvNet}
\end{subfigure}
\end{center}
\vspace{-0.3cm}
\caption{%
Qualitative comparison of the best network variants from \tabref{tab:domain_ada_eval} trained on synthetic Synthia \cite{Ros2016CVPR} and
evaluated on the proposed real-world KITTI depth dataset. While our SparseConvNet adapts well to the novel domain, standard convolutional neural networks fail completely in recovering sensible depth information.
}
\label{fig:domain_adaptation}
\end{figure*}

To evaluate the domain adaption capabilities of our method, we conduct an experiment
where we train on the Synthia dataset and evaluate on our proposed KITTI validation set.
\tabref{tab:domain_ada_eval} shows the performance of our
network (SparseConv) as well as the two regular CNN baselines using the same number of parameters.
Our experiments demonstrate that sparse convolutions perform as well on KITTI as on Synthia, while the dense baselines are not able to adapt to the new input modality and fail completely.
We show qualitative results of this experiment in \figref{fig:domain_adaptation}.

\setlength{\tabcolsep}{2.5pt}
\ctable[
    caption = { Performance comparison (MAE) of different methods trained on different sparsity
                levels on Synthia and evaluated on our newly proposed KITTI depth dataset.},
    label   = {tab:domain_ada_eval},
    pos     = {tb},
    doinside= \scriptsize,
    width   = 0.45\textwidth,
    mincapwidth = \linewidth
]{Xccccccccccc}{}{
        \FL
        Sparsity at train: & 5\%            & 10\%           & 20\%           & 30\%           & 40\%           & 50\%           & 60\%           & 70\%           \ML
        ConvNet            & 16.03          & 13.48          & 10.97          & 8.437          & 10.02          & 9.73           & 9.57           & 9.90           \NN
        ConvNet+mask       & 16.18          & 16.44          & 16.54          & 16.16          & 15.64          & 15.27          & 14.62          & 14.11          \NN
        SparseConvNet      & \textbf{0.722} & \textbf{0.723} & \textbf{0.732} & \textbf{0.734} & \textbf{0.733} & \textbf{0.731} & \textbf{0.731} & \textbf{0.730} \LL
}

\subsection{Comparison to Guided Upsampling}

As discussed in the related work section, several approaches in the literature leverage a high resolution image to guide the depth map completion task which significantly facilitates the problem.
Dense color information can be very useful to control the interpolation of sparse depth points,
\eg, to distinguish between object boundaries and smooth surfaces.
However, relying on camera information in multi-modal sensor setups, such as used in \eg autonomous cars,
is not always recommended. Bad weather and night scenes can diminish the benefit of image data or even
worsen the result. Therefore, we target an approach which leverages depth as the only input in this paper.

In this section, we show that despite not relying on guidance information, our approach performs on par with
the state-of-the-art in guided upsampling and even outperforms several methods which use image guidance.
\tabref{tab:kitti_eval} (top) shows a comparison of several state-of-the-art methods for guided filtering.
In particular, we evaluated the methods of Barron~\etal \cite{barron2016fast},
Schneider~\etal \cite{schneider2016semantically}, Ferstl~\etal \cite{Ferstl2013}, and Jampani~\etal \cite{Jampani2016CVPR} which all require a non-sparse RGB image as guidance.
For a fair comparison we added the same amount of convolutional
layers as we use in our sparse convolutional network for Jampani~\etal \cite{jampani2016learning}.
For the other baseline methods we optimized the hyper parameters via grid search on the validation split.

In addition, we compare our method to several depth-only algorithms in \tabref{tab:kitti_eval} (bottom).
We first evaluate a simple pooling approach that takes the closest (distance to sensor) valid point to fill in unseen regions within a given window.
Second, we apply the Nadaraya-Watson regressor~\cite{Nadaraya,Watson} using a Gaussian kernel on the sparse depth input. We optimized the hyperparameters of both approaches on the training data.
While the first two baselines do not require large amounts of training data, we also compare our method to high-capacity baselines. In particular, we consider a standard ConvNet with and without visibility mask as additional feature channel.

\setlength{\tabcolsep}{6pt}
\ctable[
    caption = {Performance comparison of different methods on our KITTI depth dataset.
    Our method performs comparable to state-of-the-art methods that incorporate RGB (top), while outperforming all depth-only variants (bottom).},
    label   = {tab:kitti_eval},
    pos     = {tb},
    doinside= \scriptsize,
    width   = 0.4\textwidth,
    mincapwidth = \linewidth
]{Xcccc}{}{
        \FL
        \multirow{2}{*}{Method}                                 & \multicolumn{2}{c}{MAE [m]}       & \multicolumn{2}{c}{RMSE [m]}  \\
                                                                & val           & test          & val           & test          \ML
        Bilateral NN  \cite{Jampani2016CVPR}                    & 1.09          & 1.09          & 4.19          & 5.233         \NN
        SGDU \cite{schneider2016semantically}                   & 0.72          & 0.57          & \uline{2.5}   & \uline{2.02}  \NN
        Fast Bilateral Solver \cite{barron2016fast}             & \uline{0.65}  & \uline{0.52}  & \textbf{1.98} & \textbf{1.75} \NN
        TGVL \cite{Ferstl2013}                                  & \textbf{0.59} & \textbf{0.46} & 4.85          & 4.08          \ML

        Closest Depth Pooling                                   & 0.94          & 0.68          & 2.77          & 2.30          \NN
        Nadaraya Watson\cite{Nadaraya,Watson}                   & \uline{0.74}  & 0.66          & 2.99          & 2.86          \NN
        ConvNet                                                 & 0.78          & \uline{0.62}  & 2.97          & 2.69          \NN
        ConvNet + mask                                          & 0.79          & \uline{0.62}  & \uline{2.24}  & \uline{1.94}  \NN
        SparseConvNet (ours)                                    & \textbf{0.68} & \textbf{0.54} & \textbf{2.01} & \textbf{1.81} \LL
}

It is notable that our approach performs comparable to state-of-the-art guided upsampling techniques despite not using any RGB information.
In particular, it performs second in terms of RMSE on both validation and test split which we attribute to the Euclidean loss used for training.

\subsubsection{Sparsity Evaluation on KITTI}
\label{subsubsec:kitti_sparsity_eval}

In the KITTI dataset, a 64-layer laser scanner with a rotational frequency of 10 Hz was used to acquire ground truth for various tasks such as stereo vision and flow estimation.
If projected to the image, the depth measurements cover approximately 5 \% of the image. For industrial applications such as autonomous driving,
often scanners with only 32 or 16 layers and higher frequencies are used. This results in very sparse depth projections.
To analyze the impact of extremely sparse information, we evaluate the \textit{Sparse Convolutional Network} and several baselines \wrt different levels of sparsity on our newly annotated KITTI subset.
In particular, we train all networks using all laser measurements and evaluate the performance when varying the density of the input using random dropout.
\begin{figure}[]
\centering
\includegraphics[width=0.75\linewidth]{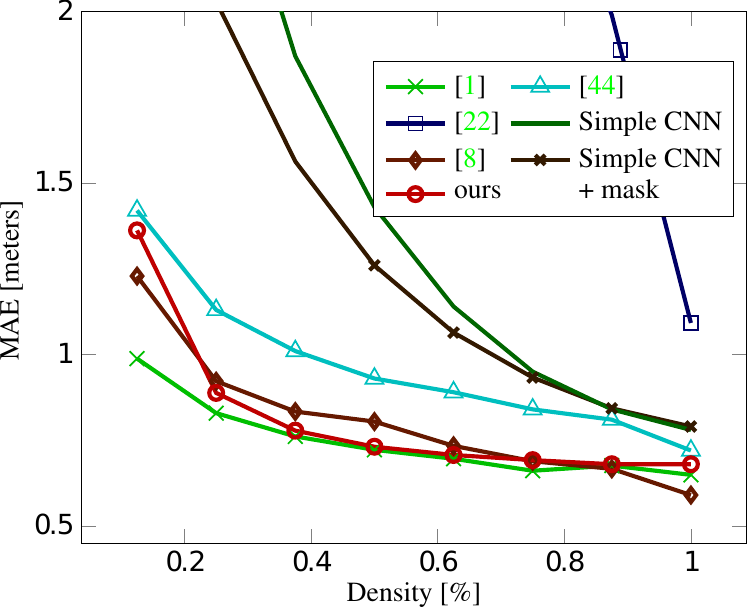}
\caption{Quantitative results in MAE (meters) on our depth annotated KITTI subset for varying levels of input density.
We compare our unguided approach to several baselines \cite{barron2016fast,schneider2016semantically,Ferstl2013,Jampani2016CVPR} which leverage RGB guidance
for upsampling and two standard convolutional neural networks with and without valid mask concatenated to the input.}
\label{fig:maeovernumpointsKitti}
\end{figure}
Our results in \figref{fig:maeovernumpointsKitti} demonstrate the generalization ability of our network for different levels of sparsity.
Regular convolutions as well as several state-of-the-art approaches perform poorly in the presence of sparse inputs.
Note that both Barron~\etal \cite{barron2016fast} and Ferstl~\etal \cite{Ferstl2013} perform slightly better than our method on very sparse data but require a dense high-resolution RGB image for guidance.

\subsection{Semantic Labeling from Sparse Depth}

To demonstrate an output modality different from depth, we also trained
the well-known VGG16 architecture \cite{FCN2015}
for the task of semantic labeling from sparse depth inputs.
We modify VGG16 by replacing the regular convolutions using our sparse convolution modules.
Additionally, we apply the weighted skip connections presented in
\secref{sec:weighted_skip} to generate high-resolution predictions from the small,
spatially downsampled FC7 layer, while incorporating visibility masks of the respective
network stages.

Table \ref{tab:sparse_semantics} shows the mean performance
after training on all Synthia ``Sequence'' frames (left camera to all
directions, summer only) and evaluating on the Synthia ``Cityscapes'' subset\footnote{We map all unknown classes in the validation set to corresponding classes in the training set and ignore all other unknown classes.}.
Again, we observe that the proposed sparse convolution module outperforms the two baselines.
The comparably small numbers can be explained by the different nature of the validation
set which contains more people and also very different viewpoints (bird's eye vs. street-level).

\setlength{\tabcolsep}{3pt}
\ctable[
    caption = { IoU performance of different network variants on the Synthia Cityscapes subset
                 after     training on all Synthia sequences (mean over all
                15 known classes).},
    label   = {tab:sparse_semantics},
    pos     = {tb},
    doinside= \scriptsize,
    width   = 0.5\linewidth,
    mincapwidth = \linewidth
]{Xccc}{}{
        \FL
        Network                             & IoU [\%]       \ML
        VGG - Depth Only                    & 6.4            \NN
        VGG - Depth + Mask                  & 4.9            \NN
        VGG - Sparse Convolutions           & 31.1           \LL
}


\section{Conclusion}

We have proposed a novel sparse convolution module for handling sparse inputs which can replace regular convolution modules and results in improved performance while generalizing well to novel domains or sparsity levels. Furthermore, we provide a newly annotated dataset with 93k depth annotated images for training and evaluating depth prediction and depth upsampling techniques.

In future work, we plan to combine the proposed sparse convolution networks with network compression techniques to handle sparse inputs while at the same time being more efficient.
We further plan to investigate the effect of sparse irregular inputs for 3D CNNs \cite{Riegler2017CVPR}.

{\small
\bibliographystyle{ieee}
\bibliography{bibliography_long,bibliography,bib}
}

\threedvfinalcopy
\def\threedvPaperID{70}
\def\httilde{\mbox{\tt\raisebox{-.5ex}{\symbol{126}}}}
\ifthreedvfinal\pagestyle{empty}\fi

\graphicspath{ {figures/} }


\title{Supplementary Material for\\Sparsity Invariant CNNs}

\author{}

\maketitle

\section*{Convergence Analysis}

\begin{figure*}[tb]
\centering
\includegraphics[width=0.7\linewidth]{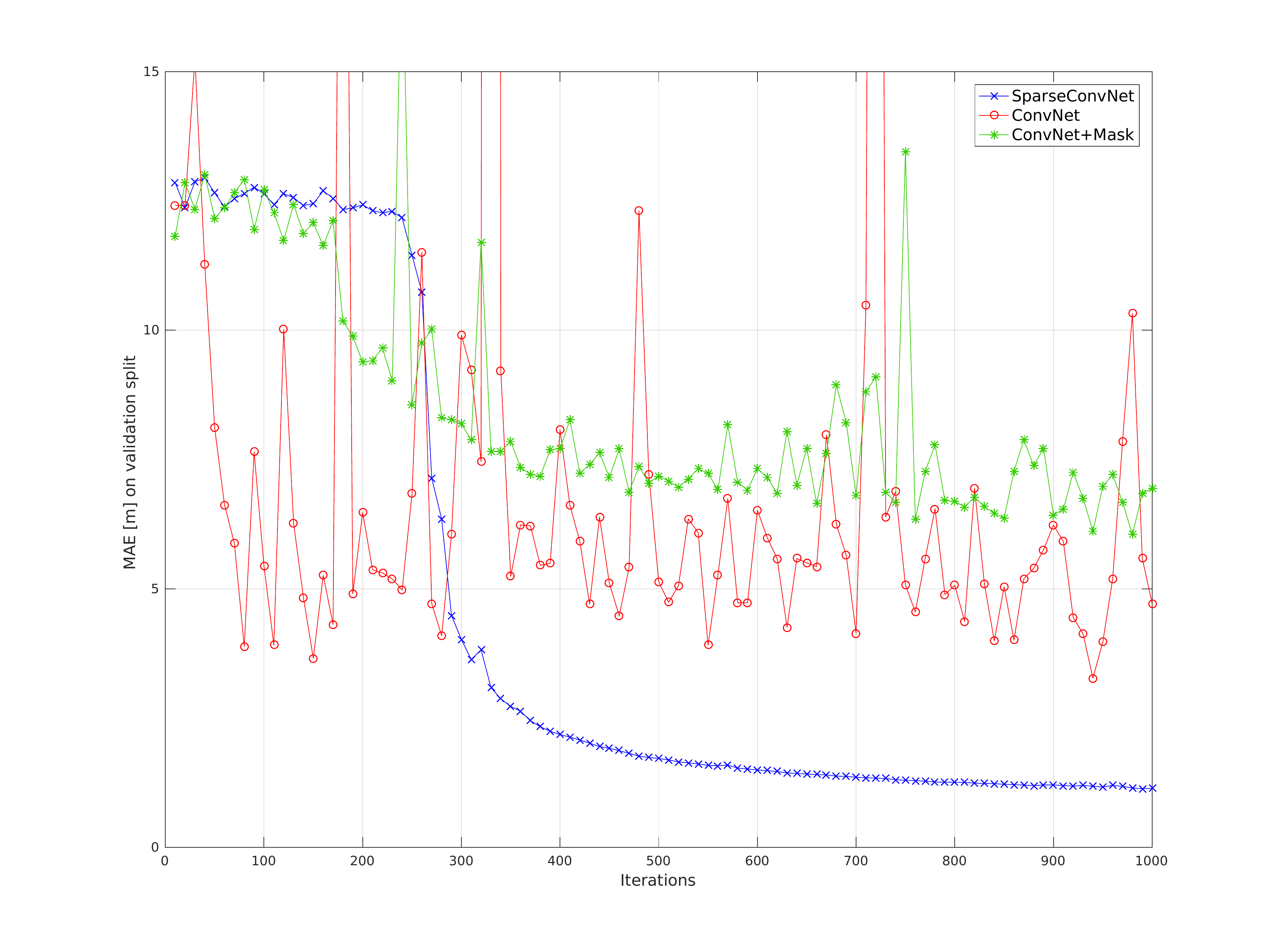}
\captionsetup{width=0.8\textwidth}
\caption{ Convergence of the three considered network baselines from Section 5.1 of the main
          paper for the task of sparse depth upsampling on 5\% dense input depth maps from
          our Synthia train subset. }
\label{fig:convergence}
\end{figure*}

We find that Sparse Convolutions converge much faster than standard
convolutions for most input-output-combinations, especially for those on Synthia
with irregularly sparse depth input, as considered in Section 5.1 of the
main paper. In Figure \ref{fig:convergence}, we show the mean average error in meters
on our validation subset of Synthia over the process of training with identical solver
settings (Adam with momentum terms of $\beta_1 = 0.9$, $\beta_2 = 0.999$ and delta
$1\mathrm{e}{-8}$). We chose for each variant the maximal learning rate which
still causes the network to converge (which turned out to be $1\mathrm{e}{-3}$ for all three variants). We
find that Sparse Convolutions indeed train much faster and much smoother compared
to both ConvNet variants, most likely caused by the explicit ignoring of invalid
regions in the update step. Interestingly, the ConvNet variant with concatenated
visibility mask in the input converges smoother than the variant with only sparse
depth in the input, however, additionally incorporating visibility masks seems to
reduce overall performance for the task of depth upsampling.

\section*{Semantic Segmentation}

\subsection*{Detailed Results on Synthia}

Relating to Section 5.3 of the main paper, we show in Table \ref{tab:app_sparse_semantics}
the class-wise IoU for semantic labeling on 5\% sparse input data and compare the three
proposed VGG-like variants: Convolutions on depth only, convolutions on depth with
concatenated visibility mask, and sparse convolutions using depth and visibility mask. We find that
sparse convolutions learn to predict also less likely classes, while standard convolutions
on such sparse data even struggle to get the most likely classes correct.

\setlength{\tabcolsep}{1pt}
\ctable[
    caption = { Evaluation of the class-level performance for
                pixel-level semantic labeling on our Synthia validation split
                subset (\textit{`Cityscapes`}) after training on all Synthia \textit{`Sequence`} subsets
                using the Intersection over Union (IoU) metric.
                All numbers are in percent and larger is better. },
    label   = {tab:app_sparse_semantics},
    pos     = {tb},
    doinside= \scriptsize,
    width   = 0.65\textwidth,
    mincapwidth = 0.9\linewidth,
    star
]{Xcccccccccccccc}{}{
\FL
& \rotatedlabel{sky}
& \rotatedlabel{building}
& \rotatedlabel{road}
& \rotatedlabel{sidewalk}
& \rotatedlabel{fence}
& \rotatedlabel{vegetation}
& \rotatedlabel{pole}
& \rotatedlabel{car}
& \rotatedlabel{traffic sign}
& \rotatedlabel{pedestrian}
& \rotatedlabel{bicycle}
& \rotatedlabel{lanemarking}
& \rotatedlabel{traffic light}
& \rotatedlabel{\textbf{mean IoU}} \ML
VGG - Depth Only            & 27.1 & 30.3 & 25.6 &  0.0 &  0.0 &  0.0 &   0.0 &  0.0 &   0.0 &   0.0 &  0.0 &  0.0 &   0.0 &  6.4 \NN 
VGG - Depth + Mask          & 20.9 & 27.8 & 14.5 &  0.0 &  0.0 &  0.0 &   0.0 &  0.0 &   0.0 &   0.0 &  0.0 &  0.0 &   0.0 &  4.9 \NN 
VGG - Sparse Convolutions   & \textbf{95.3} & \textbf{59.0} & \textbf{33.0} & \textbf{17.2} &  \textbf{1.0} & \textbf{60.5} &  \textbf{28.7} & \textbf{33.0} &  \textbf{12.5} &  \textbf{35.6} &  \textbf{6.1} &  \textbf{0.5} &  \textbf{22.4} & \textbf{31.1} \NN 
}

\subsection*{Semantic Segmentation on Real Depth Maps}

Many recent datasets provide RGB and aligned depth information along with densely
annotated semantic labels, such as Cityscapes \cite{Cordts2016CVPR} and SUN-RGBD
\cite{Song2015CVPR}. Many state-of-the-art approaches incorporate depth as well as
RGB information in order to achieve highest performance for the task of semantic
segmentation \cite{hazirbasma2016fusenet}. As the provided depth maps are often
not completely dense, we propose to use sparse convolutions on the depth channel
instead of filling depth maps artificially and applying dense convolutions afterwards.

We conduct experiments on SUN-RGBD with only depth maps as input to show the benefit
of using sparse convolutions over traditional convolutions. As seen in Section 5.3
of the main paper, sparse convolutions help to incorporate missing depth information
in the input for very sparse (5\%) depth maps. In Table \ref{tab:sparse_sunrgbd_semantics}
we show the performance of a VGG16 (with half the amount of channels than usual) trained
from scratch for the task of semantic labeling from (sparse) depth maps. We apply skip
connections as used throughout literature \cite{He2016CVPR,Long2015CVPR} up to half the
input resolution. We compare performance on the provided raw sparse depth maps (\textit{raw},
\cf Figure \ref{fig:sunrgbdmissingisgood}) as well
as a dense depth map version obtained from a special inpainting approach using neighboring frames (\textit{filled})
on the SUN-RGBD test dataset, as well as the used convolution type (sparse or standard).
We find that sparse convolutions perform better than standard convolutions, on both
raw and filled depth maps, no matter if a visibility map is concatenated to the input
depth map or not. Like reported in \cite{hazirbasma2016fusenet}, standard convolutions
on the raw depth maps do perform very bad, however, we find that concatenating the
visibility map to the input already doubles the achieved performance. A detailed class-wise
performance analysis can be found in Table \ref{tab:sparse_semantics_details}.
Note that missing information in the input, like missing depth measurements in the SUN-RGBD
dataset, does not always cause less information, which we discuss in the following section.
This phenomenon boosts methods that explicitly learn convolutions on a visibility mask, such
as the two standard convolution networks with concatenated visibility masks. Although we do not explicitly
extract features of the visibility masks we still outperform the other convolution variants.

\setlength{\tabcolsep}{6pt}
\ctable[
    caption = { Performance comparison of different input and convolution
                variants for the task of semantic labeling on (sparse or filled)
                depth maps from the SUN-RGBD dataset \cite{Song2015CVPR}. All
                networks are trained from scratch on the training split using 37
                classes, performance is evaluated on the test split as mean IoU,
                \cf \cite{hazirbasma2016fusenet}. },
    label   = {tab:sparse_sunrgbd_semantics},
    pos     = {tb},
    doinside= \scriptsize,
    width   = 0.8\linewidth,
    mincapwidth = 0.9\linewidth
]{Xllc}{}{
        \FL
        Convolution Type   & Input Depth    & Visibility Mask? & IoU [\%]         \ML
        Standard           & Raw Depth      & No               &          7.697   \NN
        Standard           & Filled Depth   & No               &         10.442   \NN
        Standard           & Raw Depth      & Concatenated     &         18.971   \NN
        Standard           & Filled Depth   & Concatenated     &         18.636   \NN
        Sparse             & Raw Depth      & Yes              & \textbf{19.640}  \LL
}

\subsection*{Discussion: Missing data is not always missing information}
\begin{figure}[tb]
\centering
\begin{center}
\begin{subfigure}{0.32\linewidth}
\includegraphics[width=\textwidth]{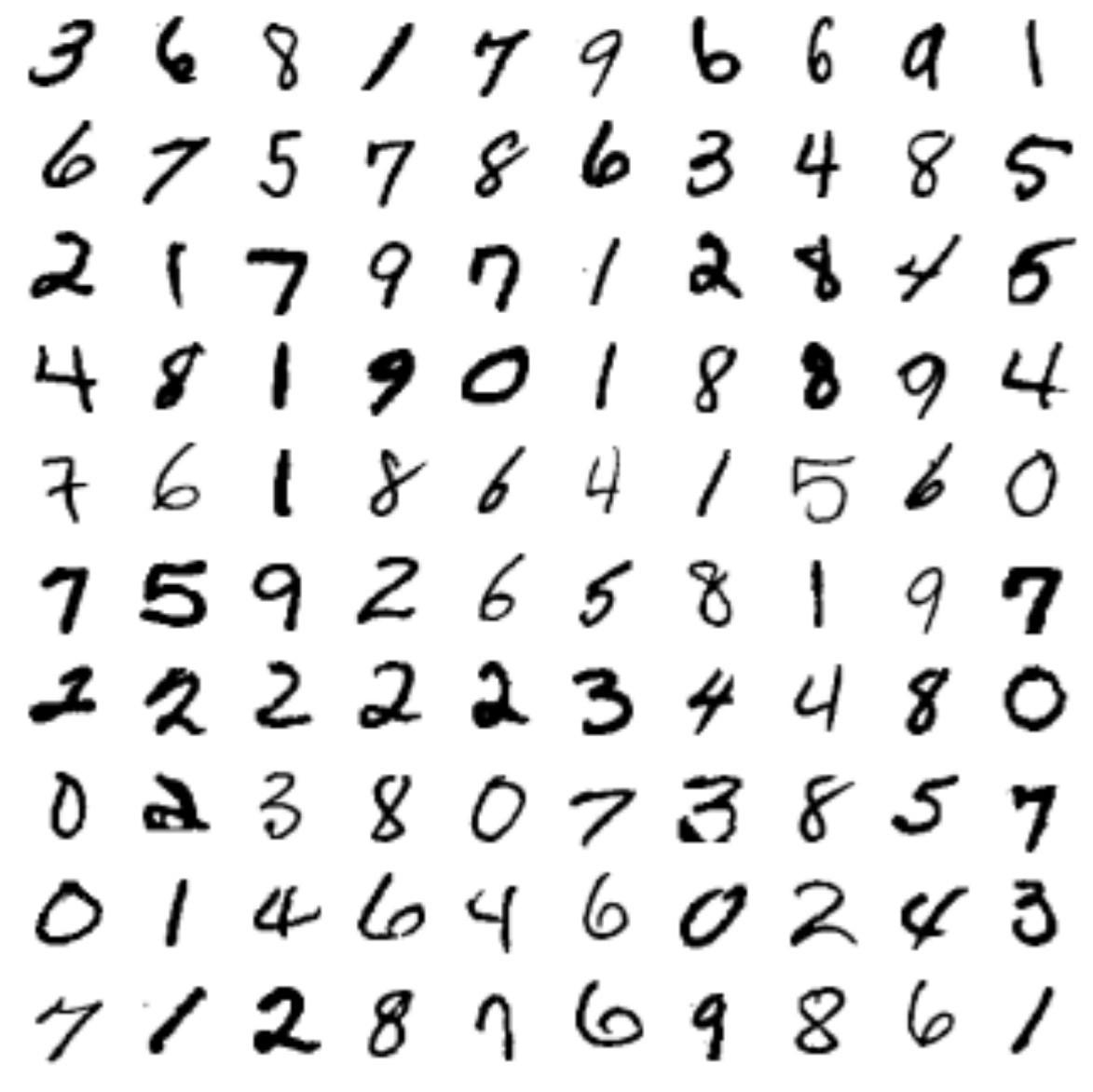}~%
\end{subfigure}
\begin{subfigure}{0.66\linewidth}
\includegraphics[width=\textwidth]{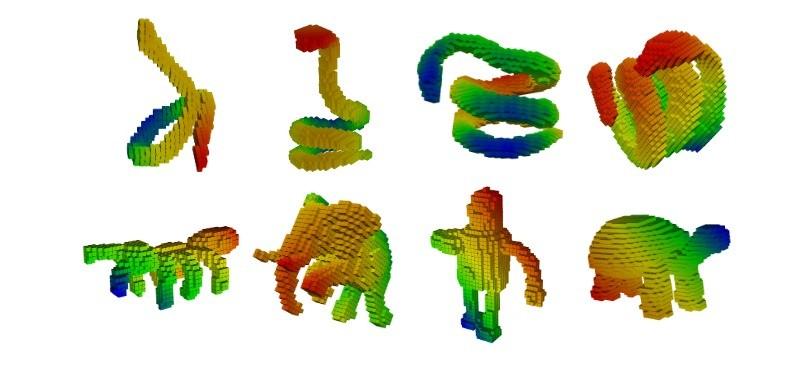}~%
\end{subfigure}
\end{center}
\caption{Missing data sometimes contains useful information as in the example of handwritten digit classification or 3D CAD model classification. Examples are taken from LeCun \etal \cite{lecun1998gradient} and Graham \cite{Graham2015BMVC}.}
\label{fig:mnistgraham}
\end{figure}

In our experiments we recognized that missing data might sometimes be helpful for certain tasks.
Let's consider \eg digit classification \cite{lecun-mnisthandwrittendigit-2010} or shape recognition from 3D CAD models as depicted in Figure \ref{fig:mnistgraham}.
For both cases the relation between invalid (background) and valid pixels/voxels is indispensable information for the classification.
We want to stress that our approach does not tackle such cases. 
Instead it handles cases where unobserved data is irregularly distributed and does not contain additional information.
Therefore, the missing data harms the results of the convolution.

\begin{figure}[tb]
\centering
\begin{center}
\begin{subfigure}{0.45\linewidth}
\includegraphics[width=\textwidth]{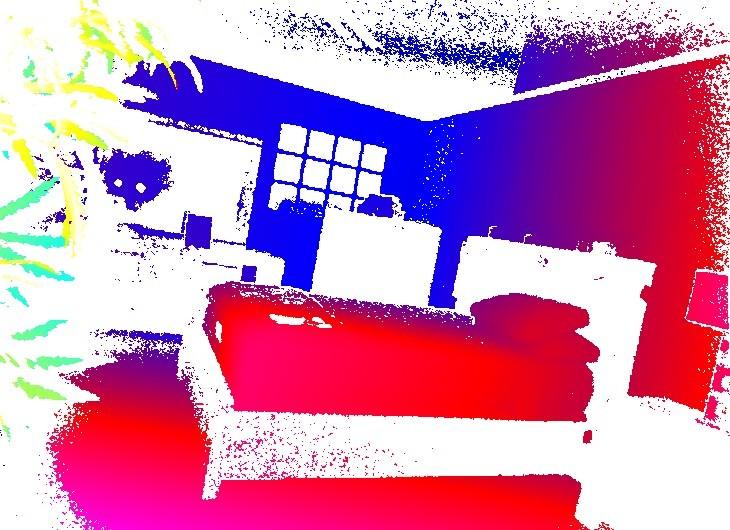}~%
\end{subfigure}
\begin{subfigure}{0.45\linewidth}
\includegraphics[width=\textwidth]{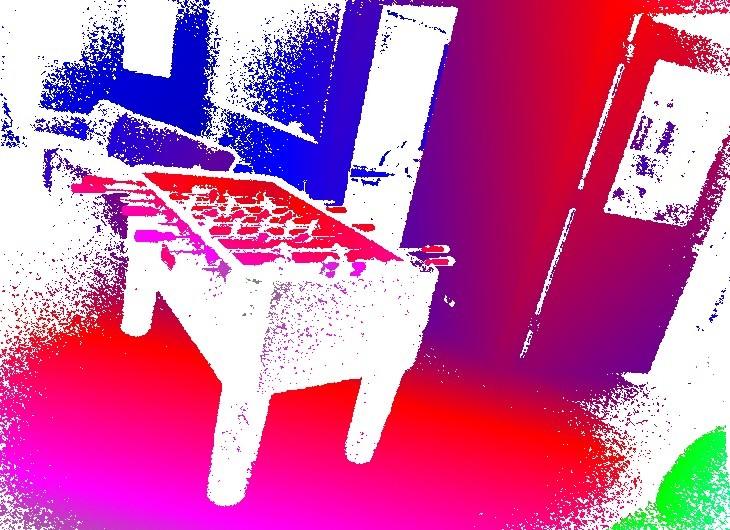}~%
\end{subfigure}
\end{center}
\caption{Active sensors such as ToF cameras might contain missing values because of strongly reflecting surfaces.
However, the missing data clearly outlines the shape of certain objects and therefore gives a hint for semantic segmentation.
This example is taken from the SUN-RGBD dataset \cite{Song2015CVPR}.}
\label{fig:sunrgbdmissingisgood}
\end{figure}

Data from active sensors, such as Time-of-Flight (ToF) cameras used in the SUN-RGBD dataset, is often sparse as shown in Figure \ref{fig:sunrgbdmissingisgood}.
However, the missing data might contain a pattern if \eg only certain materials do reflect the emitted light.
This might be the reason why the results in Table \ref{tab:sparse_sunrgbd_semantics} show a significant improvement for standard convolutions if the visibility mask is concatenated.
Our Sparse Convolution Network does not consider any missing data. Therefore, it might miss information encoded in the visibility mask. Although, the proposed method outperforms the na\"{i}ve approaches, considering the valid mask explicitly will likely further improve the performance of our method.

\setlength{\tabcolsep}{1pt}
\ctable[
    caption = { Evaluation of the class-level performance for
                pixel-level semantic labeling on Synthia Cityscapes
                subset after training on all Synthia Sequence subsets
                using the Intersection over Union (IoU) metric.
                All numbers are in percent and larger is better. Our
                sparse convolutions outperform the other variants on
                18 classes, standard convolutions on filled depth with
                concatenated visibility mask outperform the others on
                11 classes, and on 8 classes standard convolutions on
                raw depth with concatenated mask perform best. },
    label   = {tab:sparse_semantics_details},
    pos     = {tb},
    doinside= \tiny,
    width   = 0.9\textwidth,
    star
]{Xccccccccccccccccccccccccccccccccccccc|c}{}{
\FL
& \rotatedlabel{wall          }
& \rotatedlabel{floor         }
& \rotatedlabel{cabinet       }
& \rotatedlabel{bed           }
& \rotatedlabel{chair         }
& \rotatedlabel{sofa          }
& \rotatedlabel{table         }
& \rotatedlabel{door          }
& \rotatedlabel{window        }
& \rotatedlabel{bookshelf     }
& \rotatedlabel{picture       }
& \rotatedlabel{counter       }
& \rotatedlabel{blinds        }
& \rotatedlabel{desk          }
& \rotatedlabel{shelves       }
& \rotatedlabel{curtain       }
& \rotatedlabel{dresser       }
& \rotatedlabel{pillow        }
& \rotatedlabel{mirror        }
& \rotatedlabel{floor mat     }
& \rotatedlabel{clothes       }
& \rotatedlabel{ceiling       }
& \rotatedlabel{books         }
& \rotatedlabel{fridge        }
& \rotatedlabel{tv            }
& \rotatedlabel{paper         }
& \rotatedlabel{towel         }
& \rotatedlabel{shower curtain}
& \rotatedlabel{box           }
& \rotatedlabel{whiteboard    }
& \rotatedlabel{person        }
& \rotatedlabel{night stand   }
& \rotatedlabel{toilet        }
& \rotatedlabel{sink          }
& \rotatedlabel{lamp          }
& \rotatedlabel{bathtub       }
& \rotatedlabel{bag           }
& \rotatedlabel{\textbf{mean IoU}} \ML
Conv. Raw Depth                       & 49.5 & 72.3 & 0.2 & 9.4 & 26.5 & 5.5 & 29.3 & 0.2 & 17.9 & 2.6 & 0.0 & 6.3 & 0.0 & 0.0 & 0.3 & 9.0 & 0.0 & 7.4 & 0.0 & 0.0 & 0.2 & 45.3 & 0.1 & 0.0 & 0.0 & 0.6 & 0.1 & 0.0 & 0.0 & 0.0 & 0.0 & 0.0 & 0.0 & 1.5 & 0.4 & 0.0 & 0.0 & 7.7  \NN
Conv. Filled Depth                    & 53.1 & 76.1 & 8.7 & 19.6 & 34.5 & 8.5 & 34.5 & 0.3 & 9.1 & 10.4 & 0.5 & 12.3 & 0.0 & 0.0 & 0.0 & 27.6 & 0.0 & 14.0 & 0.1 & 0.0 & 1.3 & 48.9 & 5.1 & 0.0 & 0.0 & 0.1 & 0.8 & 0.0 & 0.0 & 0.0 & 0.0 & 0.0 & 2.1 & 8.4 & 7.6 & 2.9 & 0.1 & 10.4 \NN
Conv. Raw Depth, Mask concat.         & 59.4 & 80.2 & 28.7 & 53.3 & 49.0 & 37.3 & \textbf{42.9} & 2.7 & 21.7 & \textbf{17.7} & 7.3 & \textbf{22.3} & 0.0 & 5.2 & 0.9 & \textbf{35.6} & 11.4 & 21.4 & 14.0 & 0.0 & \textbf{6.3} & 34.8 & 10.0 & \textbf{6.4} & 4.6 & 0.1 & \textbf{8.6} & 0.0 & 4.7 & 12.1 & \textbf{2.7} & 0.1 & 35.0 & 30.2 & 9.2 & 23.3 & 2.7 & 19.0 \NN
Conv. Filled Depth, Mask concat.      & 59.9 & \textbf{81.6} & \textbf{29.2} & 52.5 & \textbf{50.7} & \textbf{38.6} & 42.6 & 0.6 & 15.3 & 16.9 & \textbf{11.1} & 17.0 & 0.1 & 0.5 & 0.2 & 20.5 & \textbf{12.5} & 11.3 & 16.2 & 0.0 & 4.0 & 38.0 & \textbf{18.9} & 4.3 & 5.4 & 0.0 & 5.5 & 0.0 & 3.6 & \textbf{15.7} & 0.0 & \textbf{9.3} & 32.9 & 27.4 & \textbf{17.0} & \textbf{29.9} & 0.6 & 18.6 \NN
Sparse Conv. Raw Depth                & \textbf{60.1} & 80.7 & 26.9 & \textbf{54.2} & 50.3 & 34.7 & 40.5 &  \textbf{9.3} & \textbf{22.0} & 11.0 & 10.0 & 16.6 &  \textbf{4.0} &  \textbf{8.5} &  \textbf{3.0} & 20.7 & 10.7 & \textbf{23.2} & \textbf{17.9} &  \textbf{0.0} &  3.8 & \textbf{44.5} & 10.2 &  6.2 &  \textbf{6.9} &  \textbf{2.5} &  5.2 &  \textbf{4.6} &  \textbf{5.0} & 15.3 &  1.2 &  2.8 & \textbf{42.9} & \textbf{31.6} & 11.2 & 26.4 &  \textbf{3.0} & \textbf{19.6} \LL
}

\section*{Detailed Dataset Evaluation}

Relating to Section 4.1 of the main paper, we manually extract regions in the image containing dynamic objects
in order to compare our dataset's depth map accuracy for foreground and background separately. Various error
metrics for the 142 KITTI images with corresponding raw sequences, where we differentiate between the overall average,
\cf Table \ref{tab:data_eval}, as well as foreground and background pixels, \cf Tables \ref{tab:data_eval_foreground}
and \ref{tab:data_eval_background}.

We find that our generated depth maps have a higher accuracy than all other investigated depth maps. Compared to raw LiDaR,
our generated depth maps are four times denser and contain five times less outliers in average. Even though we lose almost 50\%
of the density of the LiDaR accumulation through our cleaning procedure, we achieve almost 20 times less outliers on
dynamic objects and even a similar boost also on the static environment. This might be explained through the different
noise characteristics in critical regions, \eg where LiDaR typically blurs in lateral direction on depth edges, SGM usually
blurs in longitudinal direction. In comparison to the currently best published stereo algorithm on the KITTI 2015 stereo
benchmark website \cite{Menze2015CVPR}, which achieves 2.48, 3.59, 2.67 KITTI outlier rates for background, foreground and all
pixels (anonymous submission, checked on April 18th, 2017), the quality of our depth maps is in the range of 0.23, 2.99, 0.84.
Therefore, besides boosting depth estimation from single images (as shown in Section \ref{sec:single_image_depth}), we hope to also
boost learned stereo estimation approaches.

\setlength{\tabcolsep}{3.5pt}
\ctable[
    caption = { Evaluation of differently generated depth map variants using the manually
                annotated ground truth disparity maps of 142 corresponding KITTI benchmark
                training images \cite{Menze2015CVPR}. Best values per metric are highlighted.
                Cleaned Accumulation describes the output of our automated dataset generation
                without manual quality assurance, the extension `+ SGM` describes an additional
                cleaning step of our depth maps with SGM depth maps, applied mainly to remove
                outliers on dynamic objects. All metrics are computed in the disparity space. },
    label   = {tab:data_eval},
    pos     = {tb},
    doinside= \scriptsize,
    width   = 0.5\textwidth,
    mincapwidth = \linewidth
]{Xc|cccccc}{}{
        \FL
                                   & \multirow{2}{*}{Density} & \multirow{2}{*}{MAE} & \multirow{2}{*}{RMSE} & KITTI         & \multicolumn{3}{c}{$\delta_i$ inlier rates}      \\
                                   &                          &                      &                       & outliers      & $\delta_1$     & $\delta_2$     & $\delta_3$     \ML
        SGM                        &         82.4\%           &         1.07         &         2.80          &         4.52  &         97.00  &         98.67  &         99.19  \NN
        Raw LiDaR                  &          4.0\%           &         0.35         &         2.62          &         1.62  &         98.64  &         99.00  &         99.27  \NN
        Acc. LiDaR                 &         30.2\%           &         1.66         &         5.80          &         9.07  &         93.16  &         95.88  &         97.41  \NN
        Cleaned Acc.               &         16.1\%           & \textbf{0.35}        & \textbf{0.84}         & \textbf{0.31} & \textbf{99.79} & \textbf{99.92} & \textbf{99.95} \LL
}

\setlength{\tabcolsep}{3.5pt}
\begin{center}
    \begin{minipage}{.49\textwidth}
    \ctable[
        caption = { Evaluation as in Table \ref{tab:data_eval} but only for Foreground pixels. },
        label   = {tab:data_eval_foreground},
        pos     = H,left,
        doinside= \scriptsize,
        width   = 7.5cm
    ]{Xcccccc}{}{
            \FL
            \multirow{2}{*}{Depth Map} & \multirow{2}{*}{MAE} & \multirow{2}{*}{RMSE} & KITTI         & \multicolumn{3}{c}{$\delta_i$ inlier rates}      \\
                                       &                      &                       & outliers      & $\delta_1$     & $\delta_2$     & $\delta_3$     \ML
            SGM                        &         1.23         &          2.98         &         5.91  &         97.6   &         98.2   &         98.5   \NN
            Raw LiDaR                  &         3.72         &         10.02         &        17.36  &         84.29  &         86.11  &         88.56  \NN
            Acc. LiDaR                 &         7.73         &         12.01         &        59.73  &         55.67  &         73.73  &         83.04  \NN
            Cleaned Acc.               & \textbf{0.88}        &  \textbf{2.15}        & \textbf{2.99} & \textbf{98.55} & \textbf{98.96} & \textbf{99.17} \LL
    }
    \end{minipage}
    \begin{minipage}{.49\textwidth}
    \ctable[
        caption = { Evaluation as in Table \ref{tab:data_eval} but only for Background pixels. },
        label   = {tab:data_eval_background},
        pos     = H,left,
        doinside= \scriptsize,
        width   = 7.5cm
    ]{Xcccccc}{}{
            \FL
            \multirow{2}{*}{Depth Map} & \multirow{2}{*}{MAE} & \multirow{2}{*}{RMSE} & KITTI          & \multicolumn{3}{c}{$\delta_i$ inlier rates}      \\
                                       &                      &                       & outliers       & $\delta_1$     & $\delta_2$     & $\delta_3$     \ML
            SGM                        &         1.05         &         2.77          &         4.36   & 96.93          & 98.72          & 99.27          \NN
            Raw LiDaR                  & \textbf{0.22}        &         1.90          &         0.94   & 99.25          & 99.56          & 99.73          \NN
            Acc. LiDaR                 &         1.09         &         4.81          &         4.25   & 96.74          & 97.99          & 98.78          \NN
            Cleaned Acc.               &         0.34         & \textbf{0.77}         & \textbf{0.23}  & \textbf{99.83} & \textbf{99.94} & \textbf{99.97} \LL
    }
    \end{minipage}
\end{center}

\section*{Further Depth Upsampling Results}

\begin{figure*}[tb]
\begin{center}
\setlength{\tabcolsep}{1pt}
\begin{tabular}{ccc}
Sparse Input  & Sparse Conv. Results & Our Dataset \\
\includegraphics[width=0.28\textwidth, trim= 0 0 0 100 , clip]{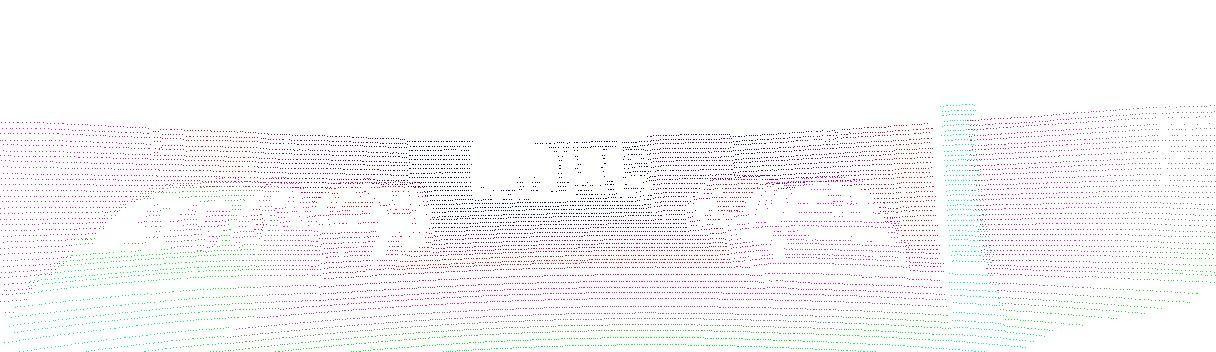}   &
\includegraphics[width=0.28\textwidth, trim= 0 0 0 100 , clip]{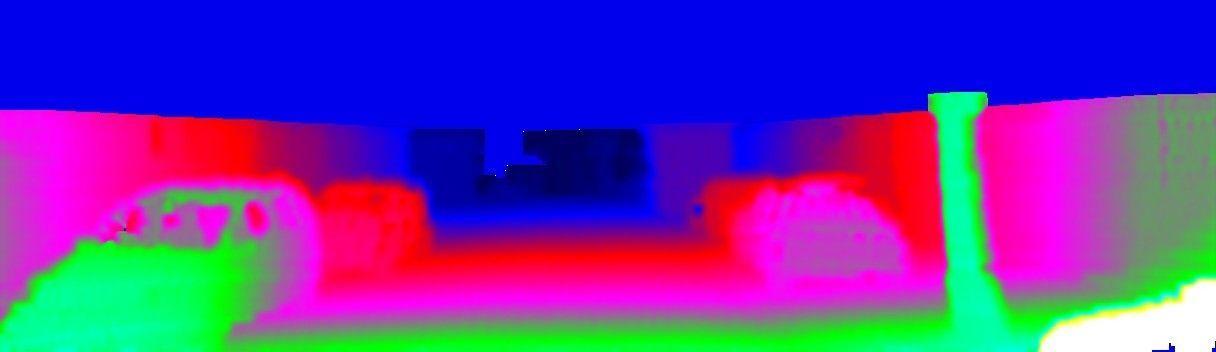} &
\includegraphics[width=0.28\textwidth, trim= 0 0 0 100 , clip]{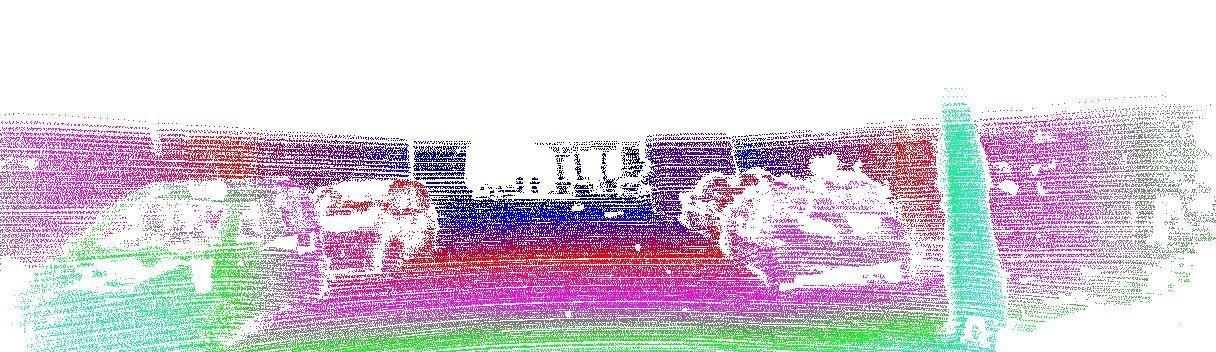}
\\ \vspace{-0.1cm}
\includegraphics[width=0.28\textwidth, trim= 0 0 0 100 , clip]{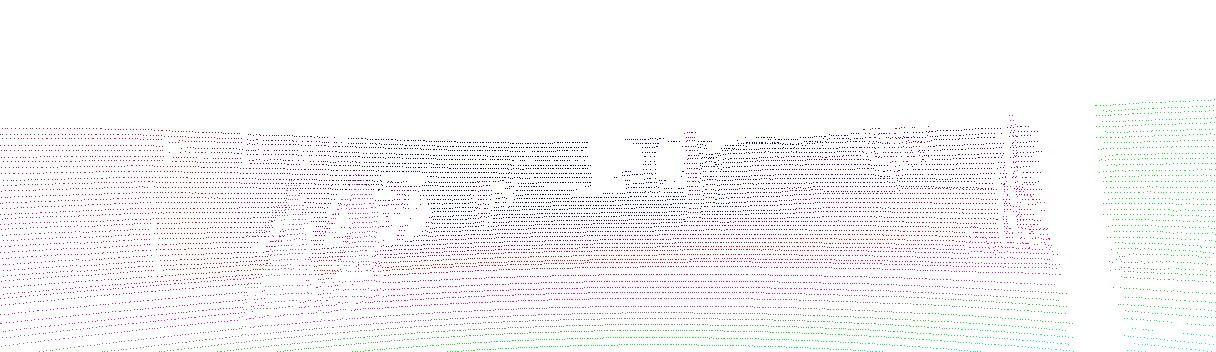}   &
\includegraphics[width=0.28\textwidth, trim= 0 0 0 100 , clip]{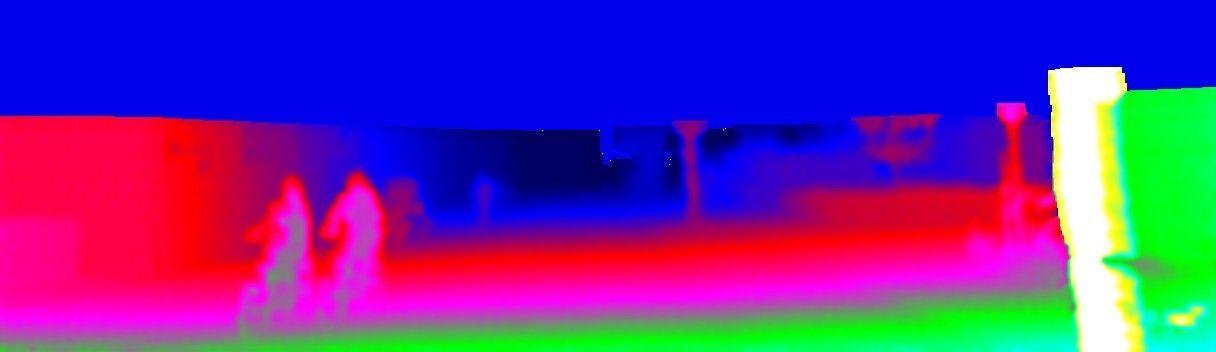} &
\includegraphics[width=0.28\textwidth, trim= 0 0 0 100 , clip]{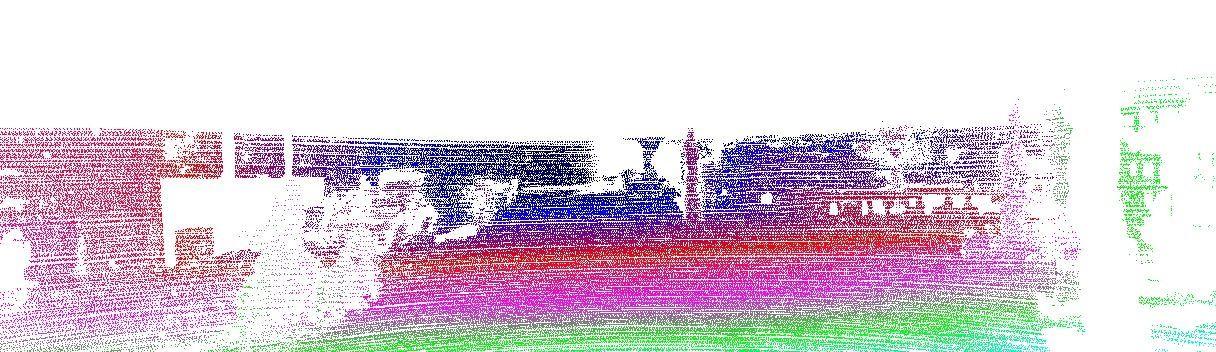}
\\ \vspace{-0.1cm}
\includegraphics[width=0.28\textwidth, trim= 0 0 0 100 , clip]{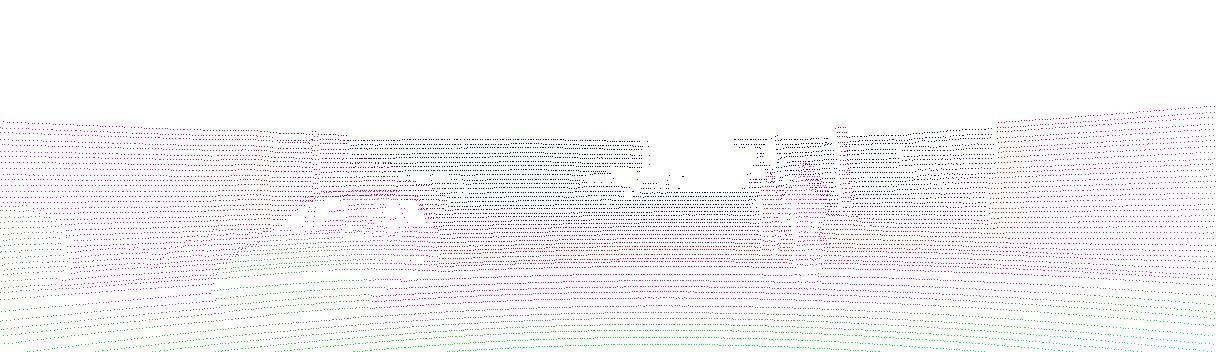}   &
\includegraphics[width=0.28\textwidth, trim= 0 0 0 100 , clip]{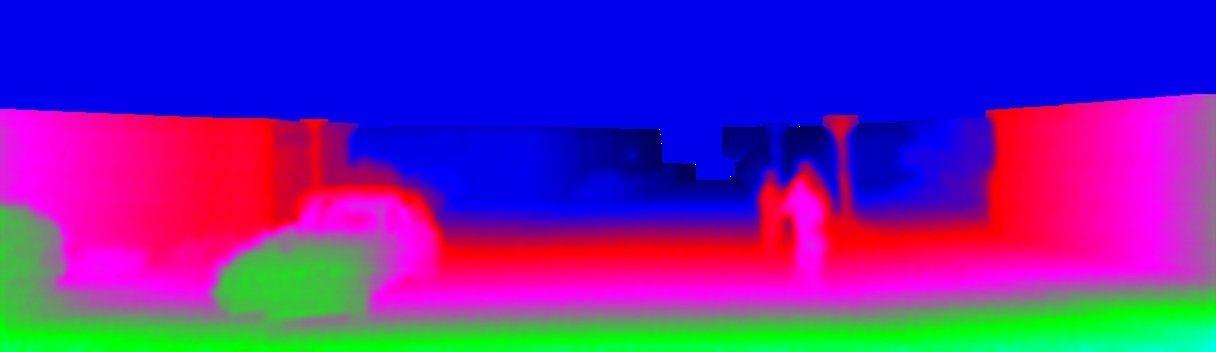} &
\includegraphics[width=0.28\textwidth, trim= 0 0 0 100 , clip]{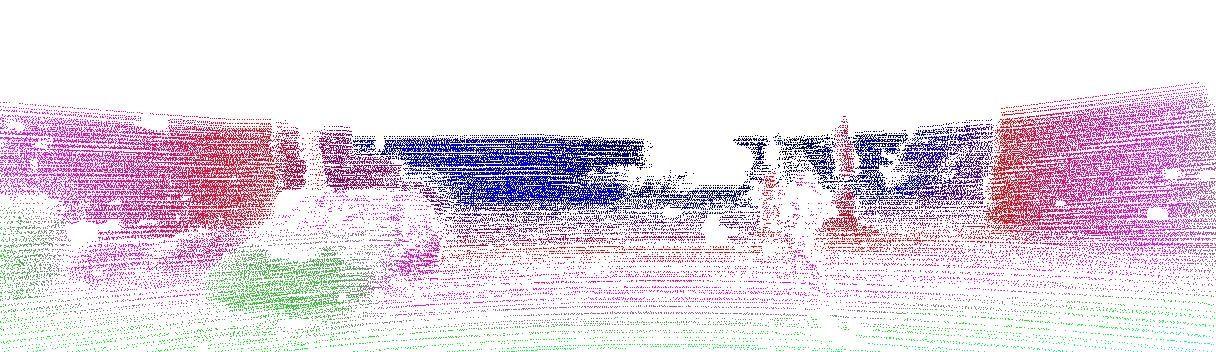}
\\ \vspace{-0.1cm}
\includegraphics[width=0.28\textwidth, trim= 0 0 0 100 , clip]{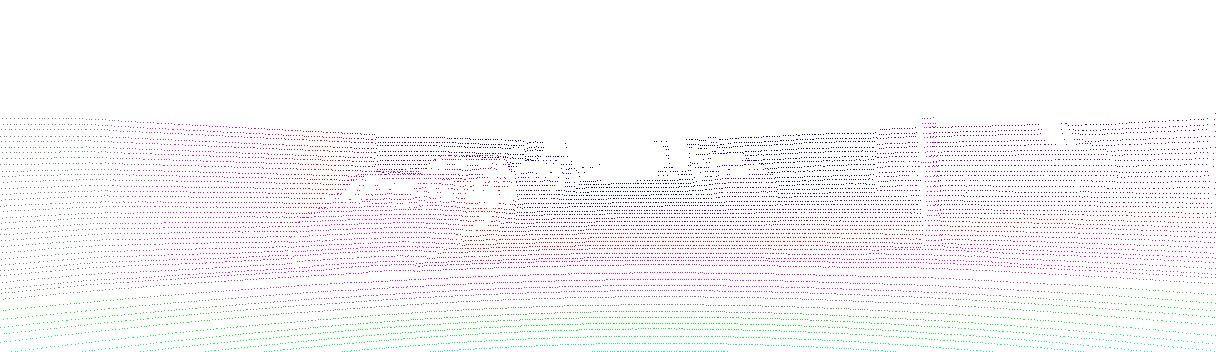}   &
\includegraphics[width=0.28\textwidth, trim= 0 0 0 100 , clip]{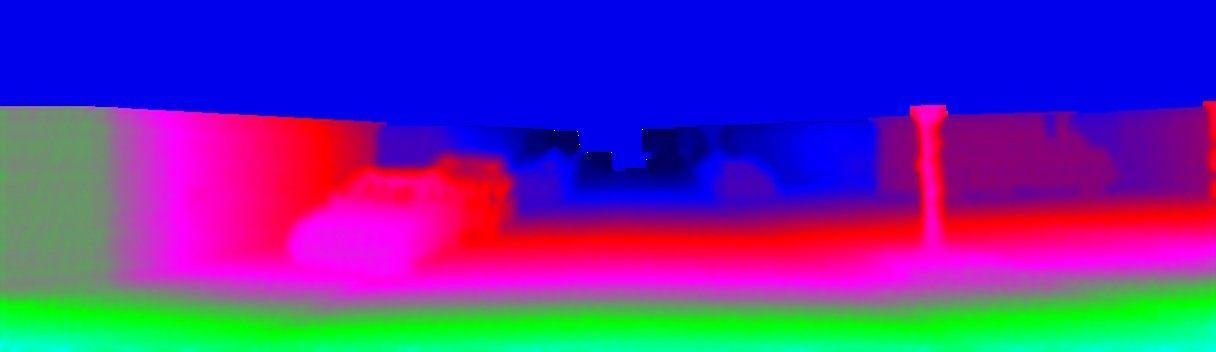} &
\includegraphics[width=0.28\textwidth, trim= 0 0 0 100 , clip]{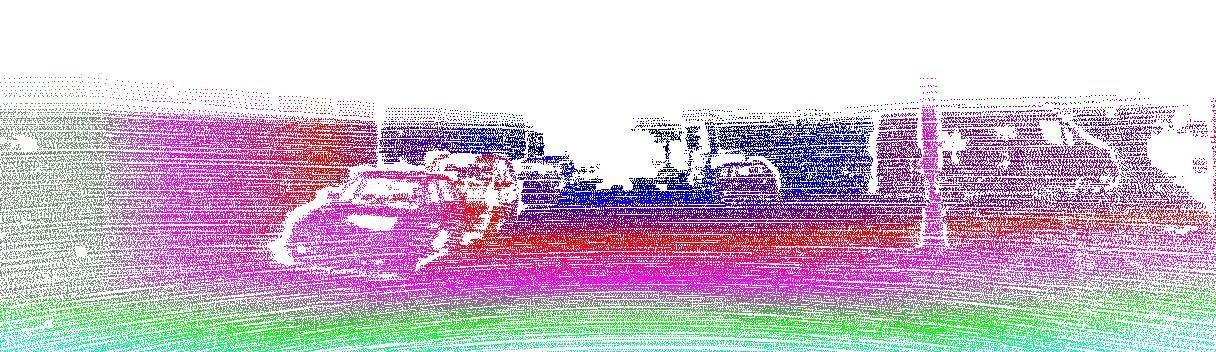}
\\ \vspace{-0.1cm}
\includegraphics[width=0.28\textwidth, trim= 0 0 0 100 , clip]{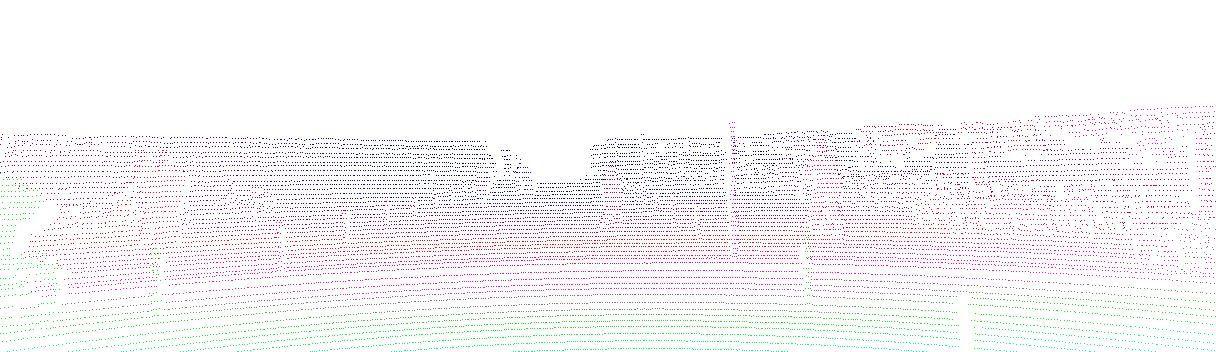}   &
\includegraphics[width=0.28\textwidth, trim= 0 0 0 100 , clip]{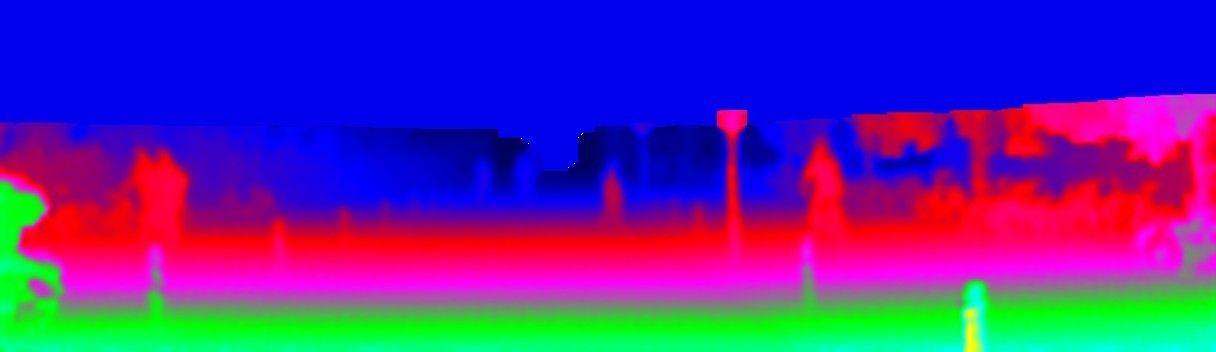} &
\includegraphics[width=0.28\textwidth, trim= 0 0 0 100 , clip]{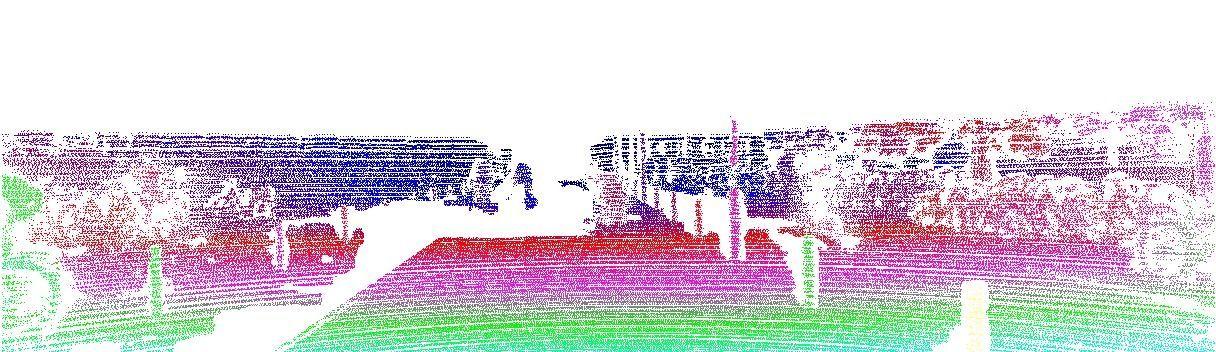}
\\

\end{tabular}
\end{center}
\vspace{-0.3cm}
\caption{Further qualitative results of our depth upsampling approach on the KITTI dataset with corresponding sparse depth input and our generated dense depth map dataset.}
\label{fig:depth_upsampling}
\end{figure*}

We show more results of our depth upsampling approach in Figure \ref{fig:depth_upsampling}. The input data of the Velodyne HDL64 is sparse and randomly distributed when projected to the image.
Our approach can handle fine structures while being smooth on flat surfaces. Sparse convolutions internally incorporate sparsity in the input and apply the learned convolutions only
to those input pixels with valid depth measurements.

\section*{Boosting Single-Image Depth Prediction}
\label{sec:single_image_depth}

As promised in Section 4 of the main paper, we conducted several experiments for a
deep network predicting depth maps from a single RGB image, \eg as done by
\cite{Eigen2014NIPS,Eigen2014ARXIV,liu2016learning} and many more. Due to the lack of
training code and to keep this study independent of current research in loss and
architecture design, we chose the well-known VGG16 architecture \cite{Long2015CVPR} with
weights initialized on the ImageNet dataset \cite{Krizhevsky2012NIPS} and vary only
the used ground truth. For a fair comparison,
we use the same amount of images and the same sequence frames for all experiments but
adapt the depth maps: Our generated dataset (denser than raw LiDaR and even more accurate),
sparse LiDaR scans (as used by most approaches for depth prediction on KITTI scenes),
as well as depth maps from semi-global matching (SGM) \cite{Hirschmueller2008PAMI},
a common real-time stereo estimation approach, \cf Table \ref{tab:depth_gt_variants} (bottom).
We evaluate the effect of training with the standard L1 and L2 losses, but do not
find large performance differences, \cf Table \ref{tab:depth_gt_variants} (top).
Also, we compare the difference between an inverse depth representation, as suggested in the
literature \cite{liu2016learning,ummenhofer2016demon},
as well as an absolute metric representation, \cf Table \ref{tab:depth_gt_variants} (top).
Surprisingly, we find that absolute depth values as ground truth representation outperform
inverse depth values. We use the best setup (absolute depth with L2 loss due to faster
convergence) to evaluate the performance on our test split, where
our dataset outperforms the other most promising depth maps from raw LiDaR, \cf Table
\ref{tab:depth_gt_variants} (bottom).

We find that our generated dataset produces visually more pleasant results and especially much
less outliers in occluded regions, \cf the car on the left for the second and last row of
Figure \ref{fig:depth_from_mono}. Also, our dense depth maps seem to help the networks
to generalize better to unseen areas, such as the upper half of the image. We hope that our
dataset will be used in the future to further boost performance for this challenging task.

\setlength{\tabcolsep}{6pt}
\ctable[
    caption = { Evaluation of different depth ground truth and loss variants (top) used for
                training a VGG16 on single-image depth prediction. L1 and L2 loss achieve
                comparable performance, while absolute depth representation for training
                instead of inverse depth performs significantly better. We compare performance
                on our generated validation and test split, as well as 142 ground truth depth
                maps from KITTI 2015 \cite{Menze2015CVPR} for the best performing setup with
                L2 loss on absolute depth (bottom). },
    label   = {tab:depth_gt_variants},
    pos     = {tb},
    doinside= \scriptsize,
    width   = 0.7\textwidth,
    mincapwidth = 0.9\textwidth
]{Xcccccccc}{}{
        \FL
        \multirow{2}{*}{Depth Maps} & \multirow{2}{*}{Loss} & Inverse     & \multicolumn{3}{c}{MAE}                          & \multicolumn{3}{c}{RMSE}                         \\
                                    &                       & Depth?      & val            & test           & KITTI`15       & val            & test           & KITTI`15       \ML
        Our Dataset                 & L2                    & yes         & 2.980          &                &                & 6.748          &                &                \NN
        Our Dataset                 & L1                    & yes         & 2.146          &                &                & 4.743          &                &                \NN
        Our Dataset                 & L2                    & \textbf{no} & 2.094          &                &                & \textbf{3.634} &                &                \NN
        Our Dataset                 & L1                    & \textbf{no} & \textbf{2.069} &                &                & 3.670          &                &                \ML

        \textbf{Our Dataset}        & L2                    & no          & \textbf{2.094} & \textbf{1.913} & \textbf{1.655} & \textbf{3.634} & \textbf{3.266} & \textbf{3.275} \NN
        Raw LiDaR Scans             & L2                    & no          & 2.184          &         1.940  &         1.790  & 3.942          &         3.297  &         3.610  \NN
        SGM                         & L2                    & no          & 3.278          &         2.973  &         3.652  & 5.826          &         4.811  &         8.927  \LL
}

\begin{figure*}[tb]
\begin{center}
\setlength{\tabcolsep}{1pt}
\begin{tabular}{cc|cc}
\multicolumn{2}{c}{\bf{Raw Lidar}} & \multicolumn{2}{c}{\bf{Our Dataset}} \\
Groundtruth & Output  & Groundtruth & Output \\
\includegraphics[width=0.23\textwidth, trim= 0 0 240 0 , clip]{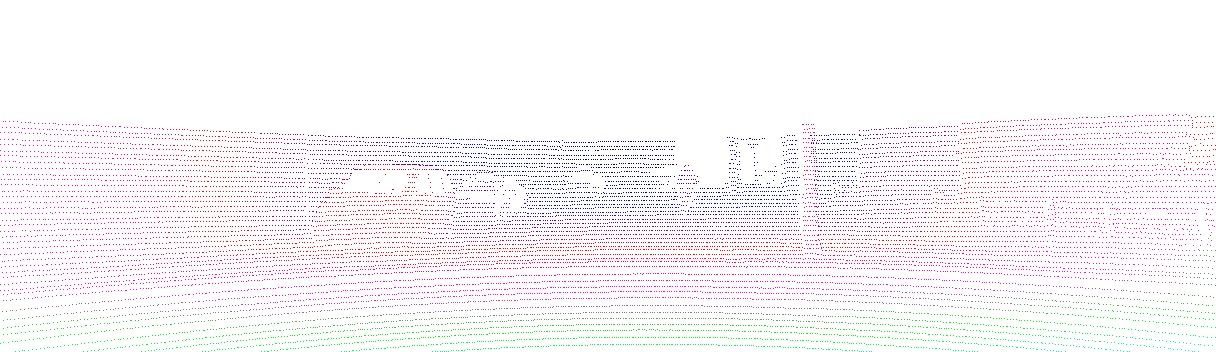}   &
\includegraphics[width=0.23\textwidth, trim= 0 0 240 0 , clip]{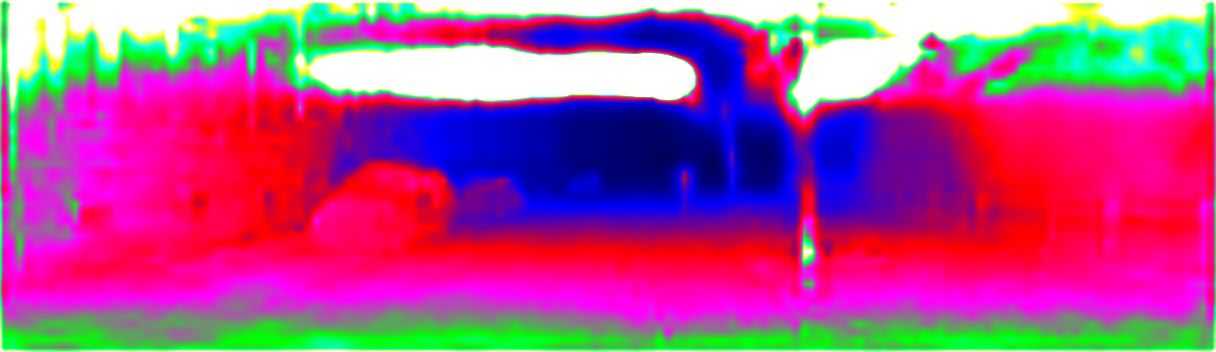} &
\includegraphics[width=0.23\textwidth, trim= 0 0 240 0 , clip]{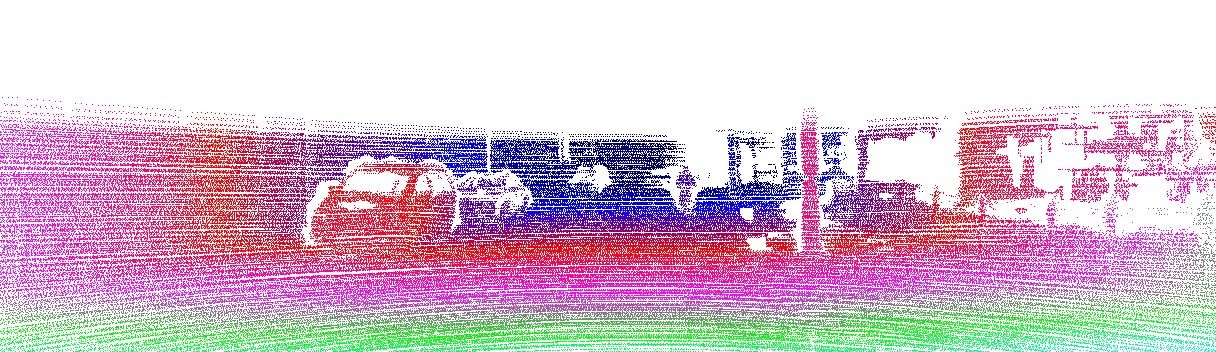}    &
\includegraphics[width=0.23\textwidth, trim= 0 0 240 0 , clip]{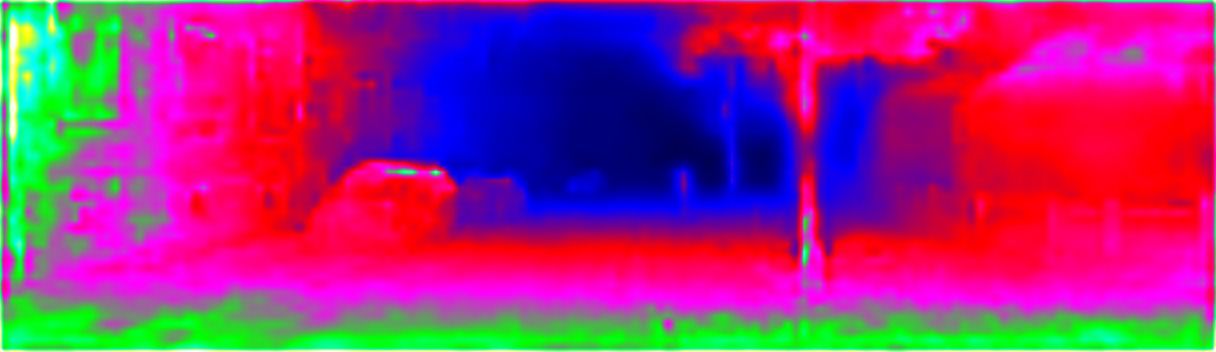}
\\ \vspace{-0.1cm}
\includegraphics[width=0.23\textwidth, trim= 0 0 240 0 , clip]{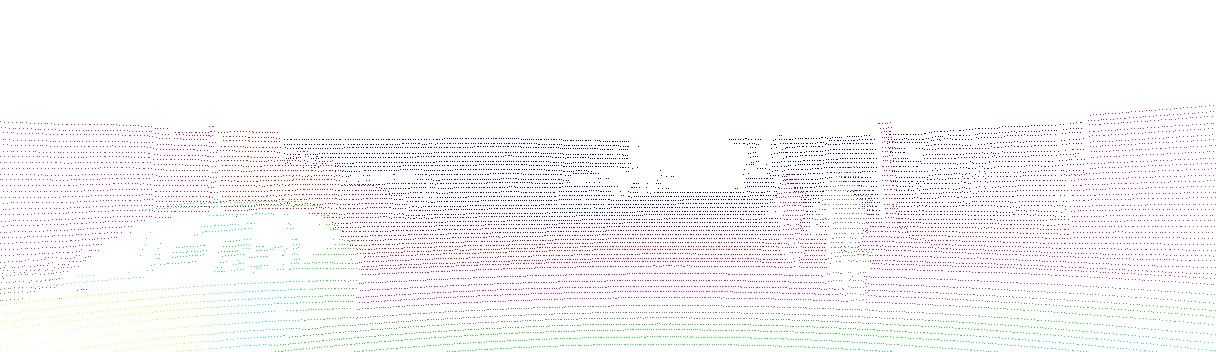}   &
\includegraphics[width=0.23\textwidth, trim= 0 0 240 0 , clip]{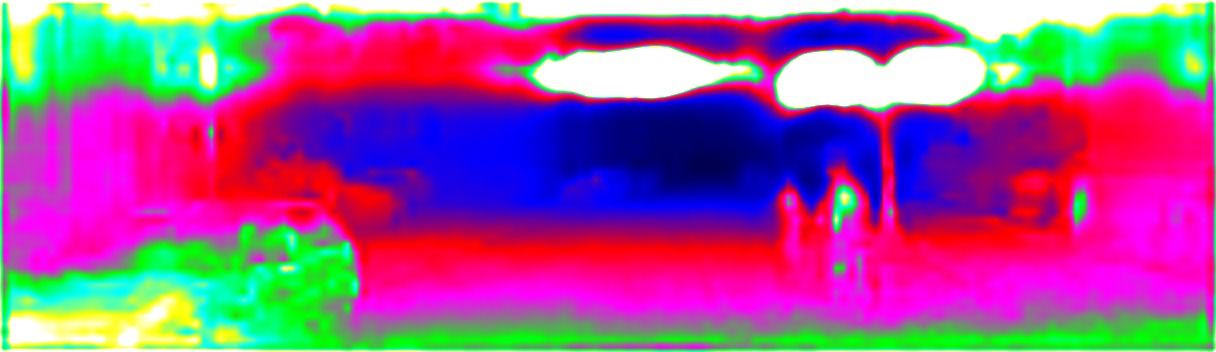} &
\includegraphics[width=0.23\textwidth, trim= 0 0 240 0 , clip]{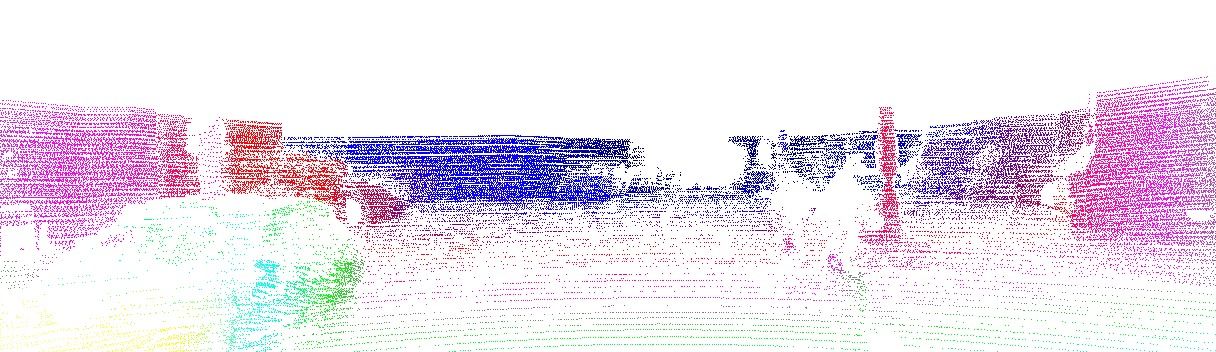}    &
\includegraphics[width=0.23\textwidth, trim= 0 0 240 0 , clip]{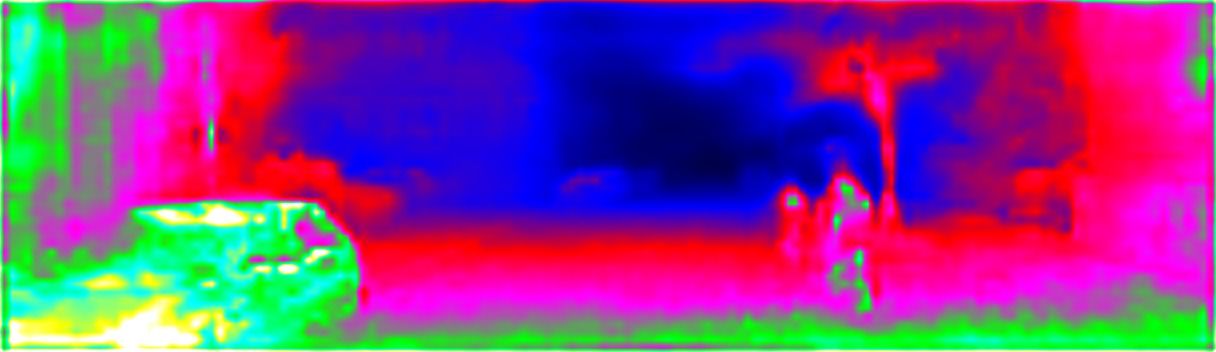}
\\ \vspace{-0.1cm}
\includegraphics[width=0.23\textwidth, trim= 0 0 240 0 , clip]{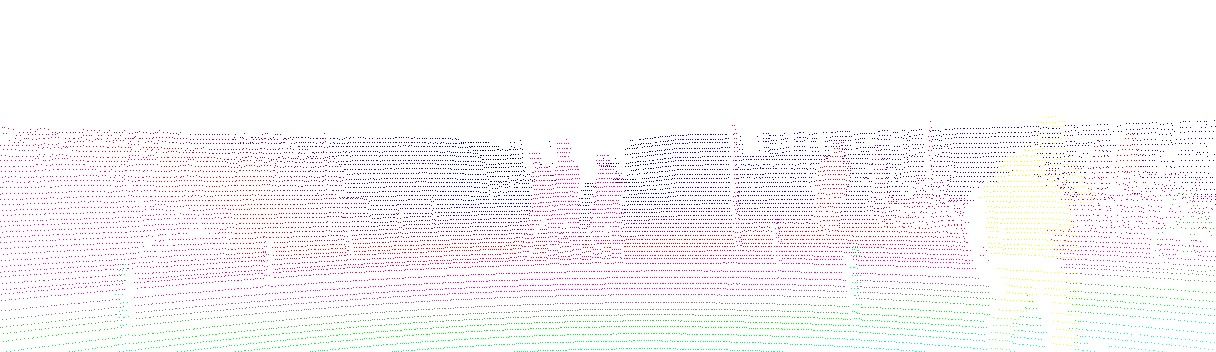}   &
\includegraphics[width=0.23\textwidth, trim= 0 0 240 0 , clip]{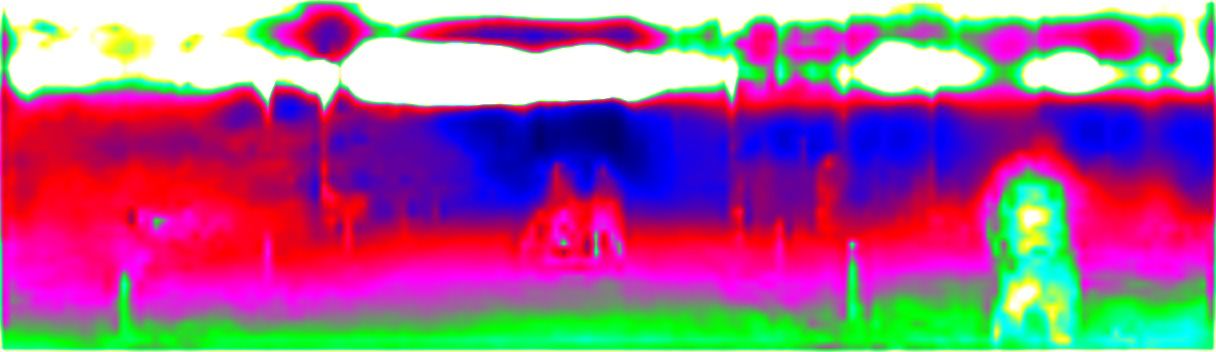} &
\includegraphics[width=0.23\textwidth, trim= 0 0 240 0 , clip]{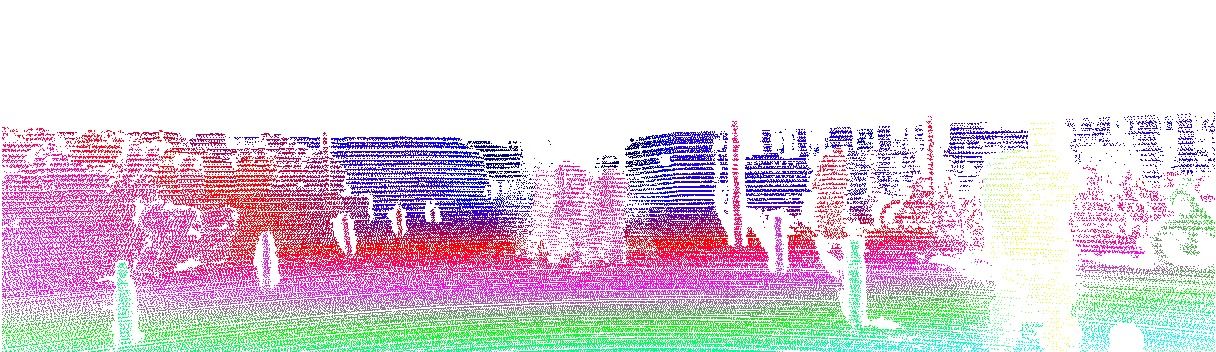}    &
\includegraphics[width=0.23\textwidth, trim= 0 0 240 0 , clip]{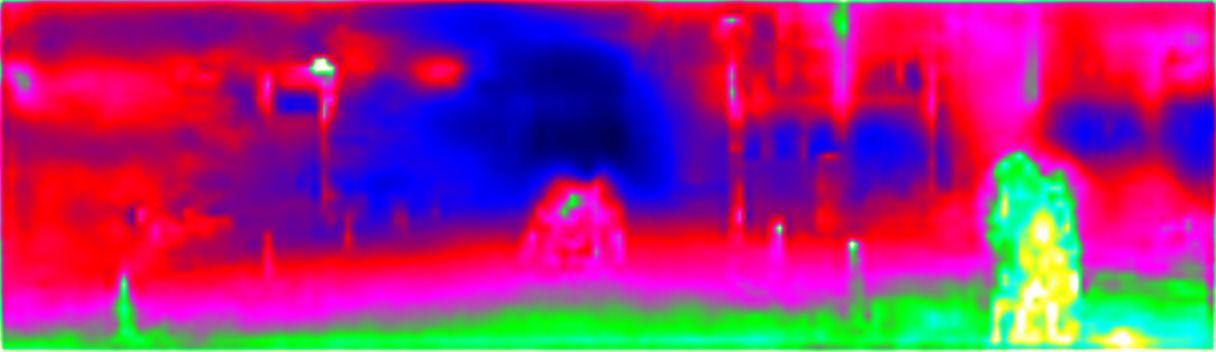}
\\ \vspace{-0.1cm}
\includegraphics[width=0.23\textwidth, trim= 0 0 240 0 , clip]{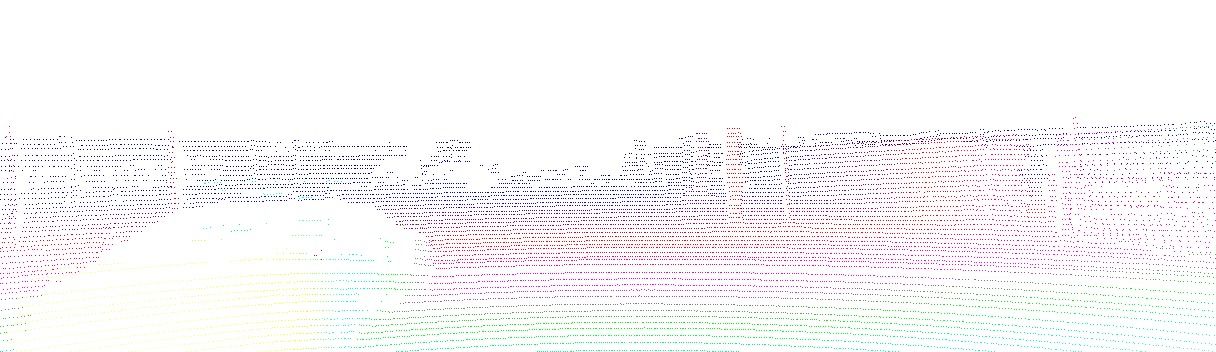}   &
\includegraphics[width=0.23\textwidth, trim= 0 0 240 0 , clip]{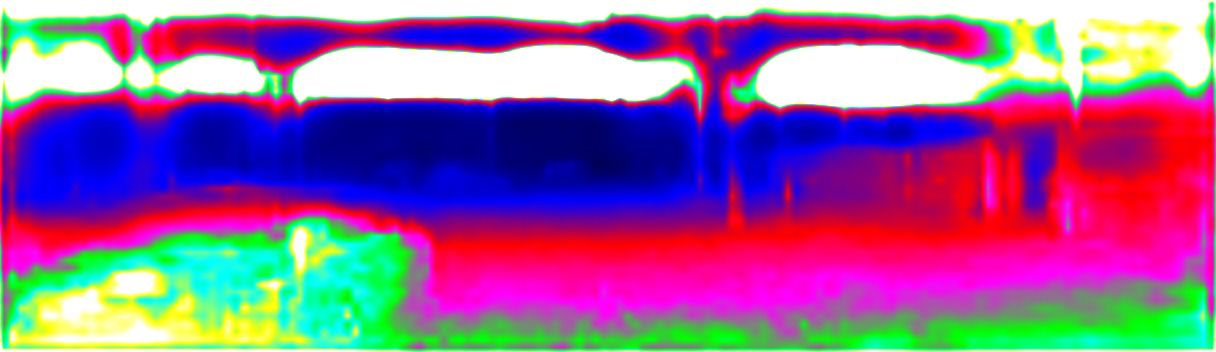} &
\includegraphics[width=0.23\textwidth, trim= 0 0 240 0 , clip]{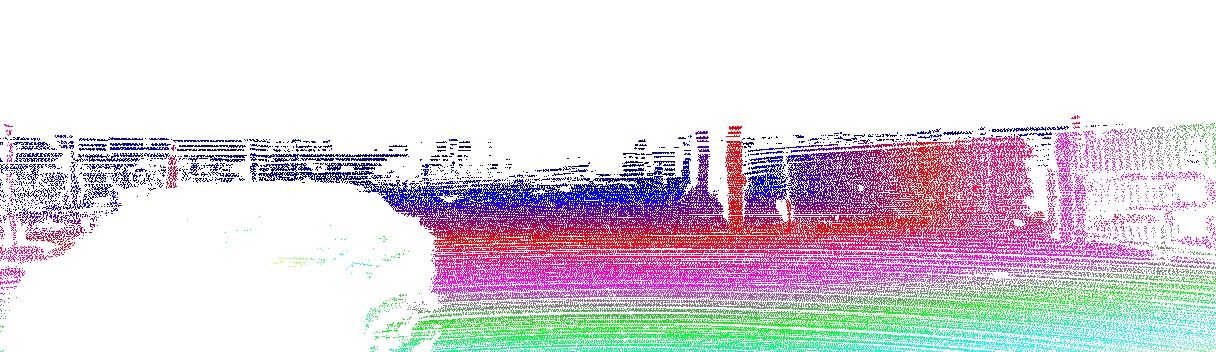}    &
\includegraphics[width=0.23\textwidth, trim= 0 0 240 0 , clip]{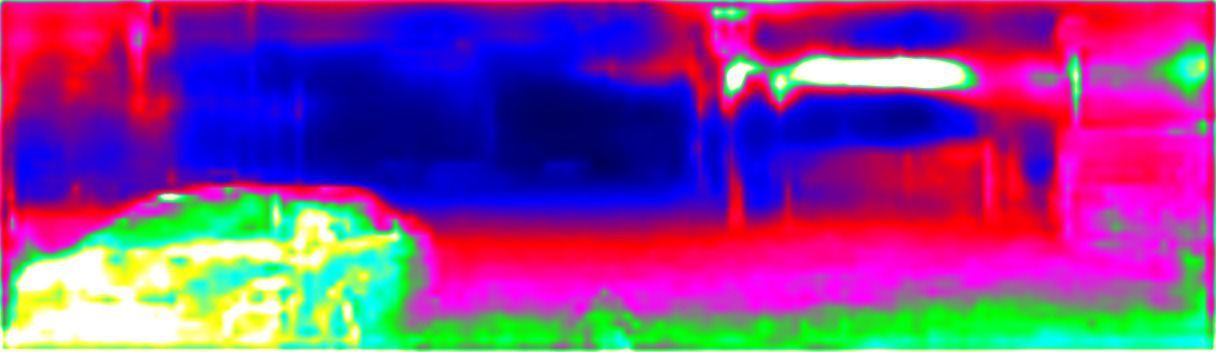}
\\
\end{tabular}
\end{center}
\vspace{-0.3cm}
\caption{%
\textbf{Raw Lidar vs our dataset as training data for depth from mono:} Qualitative examples of the depth-from-mono CNN
trained on our generated dense and outlier-cleaned dataset in contrast to the sparse raw LiDaR data. It becomes apparent that
denser training data leads to improved results \eg in the upper half of the image and at object boundaries (where most
LiDaR outliers occur).
}
\label{fig:depth_from_mono}
\end{figure*}

\end{document}